%% file: manuscript.tex
\documentclass{article}

\PassOptionsToPackage{numbers, compress}{natbib}
\usepackage[final]{neurips_2024}

\usepackage[pagebackref=true,unicode=true,colorlinks=true,citecolor=blue,linkcolor=red]{hyperref}
\usepackage[utf8]{inputenc} % allow utf-8 input
\usepackage[T1]{fontenc}    % use 8-bit T1 fonts
\usepackage{hyperref}       % hyperlinks
\usepackage{url}            % simple URL typesetting
\usepackage{booktabs}       % professional-quality tables
\usepackage{amsfonts}       % blackboard math symbols
\usepackage{nicefrac}       % compact symbols for 1/2, etc.
\usepackage{microtype}      % microtypography
\usepackage{xcolor}         % colors
\usepackage{graphicx}
\usepackage{adjustbox}
\usepackage{pifont}
\usepackage{amsmath}
\usepackage{graphicx}
\usepackage{pgfplots}
\usepackage{comment}
\usepackage{amsmath,amssymb} %
\usepackage{color}
\usepackage{soul}
\usepackage{svg}
\usepackage{wrapfig}
\usepackage{array}
\usepackage{xspace}
\usepackage{subcaption}
\usepackage{wrapfig}
\usepackage{mathrsfs}
\input{preamble}

\newcommand{\ie}{{\emph{i.e.}},\xspace}
\newcommand{\eg}{{\emph{e.g.}},\xspace}

\title{Context and Geometry Aware Voxel Transformer for Semantic Scene Completion}

\author{
	Zhu Yu$^{1}$ \hspace{1em}
	Runmin Zhang$^{1}$ \hspace{1em}
	Jiacheng Ying$^{1}$ \hspace{1em}
	Junchen Yu$^{1}$
	\\ \bf
	Xiaohai Hu$^{3}$  \hspace{1em}
	Lun Luo$^{4}$  \hspace{1em}
	Si-Yuan Cao$^{2,1*}$ \hspace{1em}
	Hui-Liang Shen$^{1}$\thanks{Corresponding author} \hspace{1em}
	\\[2pt]
	$^1$Zhejiang University\hspace{2em}
	$^2$Ningbo Innovation Center, Zhejiang University
	\\
	$^3$University of Washington\hspace{2em}
	$^4$HAOMO.AI Technology Co., Ltd.
	\\
	\href{https://github.com/pkqbajng/CGFormer}{https://github.com/pkqbajng/CGFormer}\hspace{1em}
}

\begin{document}

\maketitle

\begin{abstract}
	\input{src/abstract}
\end{abstract}

\input{src/introduction}
\input{src/related}
\input{src/method}
\input{src/experiments}

\input{src/conclusion}
\input{src/acknowledgments}
\bibliography{references}
\input{src/appendix}
\bibliographystyle{ieee_fullname}

\end{document}

%% file: preamble.tex
\usepackage{multirow}
\usepackage{graphicx}
\usepackage{wrapfig}
\usepackage{float}

\definecolor{White}{rgb}{1.,0.,1.}
\definecolor{first}{rgb}{.8,.0,.0}
\definecolor{second}{rgb}{.0,.6,.0}
\definecolor{third}{rgb}{.0,.0,.8}

\definecolor{ceiling}{RGB}{214,  38, 40}
\definecolor{floor}{RGB}{43, 160, 4}
\definecolor{wall}{RGB}{158, 216, 229}
\definecolor{window}{RGB}{114, 158, 206}
\definecolor{chair}{RGB}{204, 204, 91}
\definecolor{bed}{RGB}{255, 186, 119}
\definecolor{sofa}{RGB}{147, 102, 188}
\definecolor{table}{RGB}{30, 119, 181}
\definecolor{tvs}{RGB}{160, 188, 33}
\definecolor{furniture}{RGB}{255, 127, 12}
\definecolor{objects}{RGB}{196, 175, 214}

\definecolor{car}{rgb}{0.39215686, 0.58823529, 0.96078431}
\definecolor{bicycle}{rgb}{0.39215686, 0.90196078, 0.96078431}
\definecolor{motorcycle}{rgb}{0.11764706, 0.23529412, 0.58823529}
\definecolor{truck}{rgb}{0.31372549, 0.11764706, 0.70588235}
\definecolor{othervehicle}{rgb}{0.39215686, 0.31372549, 0.98039216}
\definecolor{person}{rgb}{1.        , 0.11764706, 0.11764706}
\definecolor{bicyclist}{rgb}{1.        , 0.15686275, 0.78431373}
\definecolor{motorcyclist}{rgb}{0.58823529, 0.11764706, 0.35294118}
\definecolor{road}{rgb}{1.        , 0.        , 1.        }
\definecolor{parking}{rgb}{1.        , 0.58823529, 1.        }
\definecolor{sidewalk}{rgb}{0.29411765, 0.        , 0.29411765}
\definecolor{otherground}{rgb}{0.68627451, 0.        , 0.29411765}
\definecolor{building}{rgb}{1.        , 0.78431373, 0.        }
\definecolor{fence}{rgb}{1.        , 0.47058824, 0.19607843}
\definecolor{vegetation}{rgb}{0.        , 0.68627451, 0.        }
\definecolor{trunk}{rgb}{0.52941176, 0.23529412, 0.        }
\definecolor{terrain}{rgb}{0.58823529, 0.94117647, 0.31372549}
\definecolor{pole}{rgb}{1.        , 0.94117647, 0.58823529}
\definecolor{trafficsign}{rgb}{1.        , 0.        , 0.        }
\definecolor{otherstructure}{rgb}{0.98039215, 0.58823529, 0.}
\definecolor{otherobject}{rgb}{0.19607843, 1.        , 1.        }

\makeatletter
\newcommand{\car@semkitfreq}{3.92}
\newcommand{\bicycle@semkitfreq}{0.03}
\newcommand{\motorcycle@semkitfreq}{0.03}
\newcommand{\truck@semkitfreq}{0.16}
\newcommand{\othervehicle@semkitfreq}{0.20}
\newcommand{\person@semkitfreq}{0.07}
\newcommand{\bicyclist@semkitfreq}{0.07}
\newcommand{\motorcyclist@semkitfreq}{0.05}
\newcommand{\road@semkitfreq}{15.30}
\newcommand{\parking@semkitfreq}{1.12}
\newcommand{\sidewalk@semkitfreq}{11.13}
\newcommand{\otherground@semkitfreq}{0.56}
\newcommand{\building@semkitfreq}{14.1}
\newcommand{\fence@semkitfreq}{3.90}
\newcommand{\vegetation@semkitfreq}{39.3}
\newcommand{\trunk@semkitfreq}{0.51}
\newcommand{\terrain@semkitfreq}{9.17}
\newcommand{\pole@semkitfreq}{0.29}
\newcommand{\trafficsign@semkitfreq}{0.08}
\newcommand{\semkitfreq}[1]{{\csname #1@semkitfreq\endcsname}}

\newcommand{\car@sscbkitfreq}{2.85}
\newcommand{\bicycle@sscbkitfreq}{0.01}
\newcommand{\motorcycle@sscbkitfreq}{0.01}
\newcommand{\truck@sscbkitfreq}{0.16}
\newcommand{\othervehicle@sscbkitfreq}{5.75}
\newcommand{\person@sscbkitfreq}{0.02}
\newcommand{\road@sscbkitfreq}{14.98}
\newcommand{\parking@sscbkitfreq}{2.31}
\newcommand{\sidewalk@sscbkitfreq}{6.43}
\newcommand{\otherground@sscbkitfreq}{2.05}
\newcommand{\building@sscbkitfreq}{15.67}
\newcommand{\fence@sscbkitfreq}{0.96}
\newcommand{\vegetation@sscbkitfreq}{41.99}
\newcommand{\terrain@sscbkitfreq}{7.10}
\newcommand{\pole@sscbkitfreq}{0.22}
\newcommand{\trafficsign@sscbkitfreq}{0.06}
\newcommand{\otherstructure@sscbkitfreq}{4.33}
\newcommand{\otherobject@sscbkitfreq}{0.28}
\newcommand{\sscbkitfreq}[1]{{\csname #1@sscbkitfreq\endcsname}}

%% file: src/abstract.tex
Vision-based Semantic Scene Completion (SSC) has gained much attention due to its widespread applications in various 3D perception tasks. Existing sparse-to-dense approaches typically employ shared context-independent queries across various input images, which fails to capture distinctions among them as the focal regions of different inputs vary and may result in undirected feature aggregation of cross-attention. Additionally, the absence of depth information may lead to points projected onto the image plane sharing the same 2D position or similar sampling points in the feature map, resulting in depth ambiguity. In this paper, we present a novel context and geometry aware voxel transformer. It utilizes a context aware query generator to initialize context-dependent queries tailored to individual input images, effectively capturing their unique characteristics and aggregating information within the region of interest. Furthermore, it extend deformable cross-attention from 2D to 3D pixel space, enabling the differentiation of points with similar image coordinates based on their depth coordinates. Building upon this module, we introduce a neural network named CGFormer to achieve semantic scene completion. Simultaneously, CGFormer leverages multiple 3D representations (i.e., voxel and TPV) to boost the semantic and geometric representation abilities of the transformed 3D volume from both local and global perspectives. Experimental results demonstrate that CGFormer achieves state-of-the-art performance on the SemanticKITTI and SSCBench-KITTI-360 benchmarks, attaining a mIoU of 16.87 and 20.05, as well as an IoU of 45.99 and 48.07, respectively. Remarkably, CGFormer even outperforms approaches employing temporal images as inputs or much larger image backbone networks.

%% file: src/introduction.tex
\section{Introduction}
\label{sec:intro}
Semantic Scene Completion (SSC) aims to jointly infer the complete scene geometry and semantics. It serves as a crucial step in a wide range of 3D perception tasks such as autonomous driving~\cite{UniAD, bevformer}, robotic navigation~\cite{occ3d, DenseCaption}, mapping and planning~\cite{SGN}. SSCNet~\cite{SSCNet} initially formulates the semantic scene completion task. Subsequently, many LiDAR-based approaches~\cite{S3CNet, JS3CNet, LMSCNet} have been proposed, but these approaches usually suffer from high-cost sensors. 

Recently, there has been a shift towards vision-based SSC solutions. MonoScene~\cite{MonoScene} lifts the input 2D images to 3D volumes by densely assigning the same 2D features to both visible and occluded regions, leading to many ambiguities. With advancements in bird's-eye-view (BEV) perception~\cite{bevformer, BEVFormerv2}, transformer-based approaches~\cite{TPVFormer, surroundOcc, VoxFormer} achieve feature lifting by projecting 3D queries from 3D space to image plane and aggregating 3D features through deformable attention mechanisms~\cite{DeformableDetr}. Among these, VoxFormer~\cite{VoxFormer} introduces a sparse-to-dense architecture, which first aggregates 3D information for the visible voxels using the depth-based queries and then completes the 3D information for non-visible regions by leveraging the reconstructed visible areas as starting points. Building upon VoxFormer~\cite{VoxFormer}, the following approaches further improve performance through self-distillation training strategy~\cite{HASSC}, extracting instance features from images~\cite{Symphonize}, or incorporating an image-conditioned cross-attention module~\cite{MonoOcc}.

\begin{figure*}[t]
	\centering\includegraphics[width=0.7\linewidth]{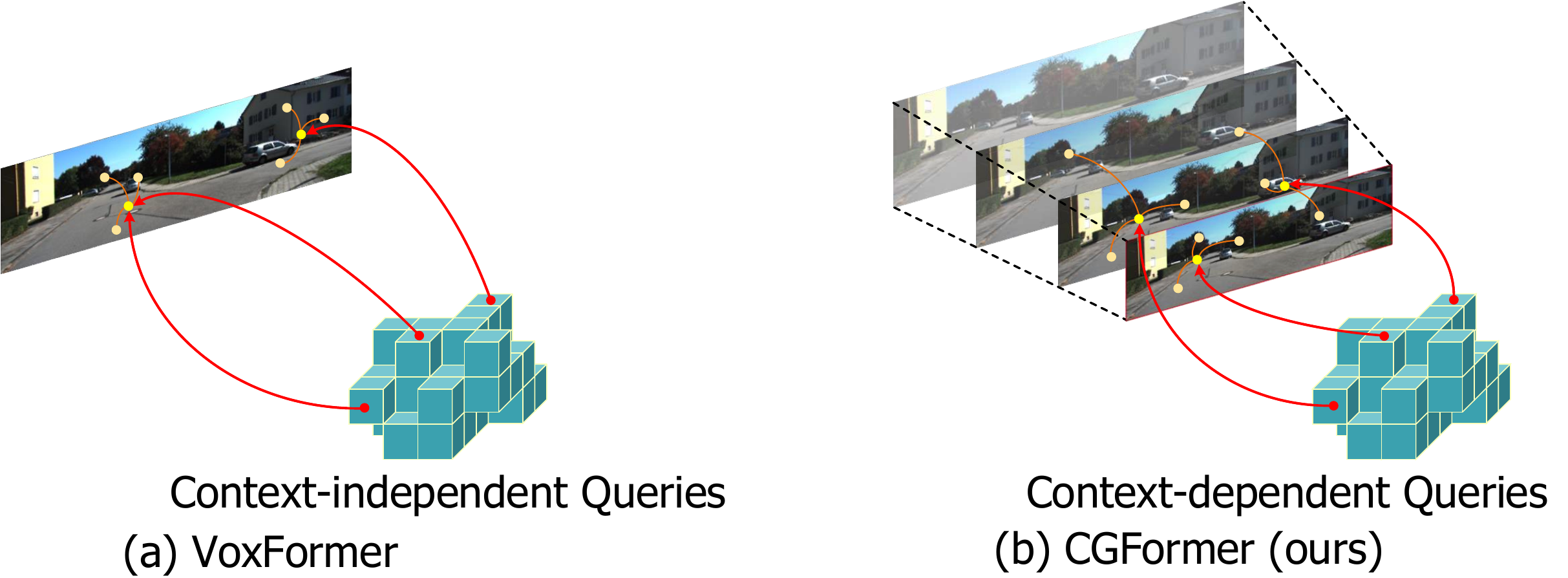}
	\caption{Comparison of feature aggregation. (a) VoxFormer~\cite{VoxFormer} employs a set of shared context-independent queries for different input images, which fails to capture distinctions among them and may lead to undirected feature aggregation. Besides, due to the ignorance of depth information, multiple 3D points may be projected to the same 2D point, causing depth ambiguity. (b) Our CGFormer initializes the voxel queries based on individual input images, effectively capturing their unique features and aggregating information within the region of interest. Furthermore, the deformable cross-attention is extended from 2D to 3D pixel space, enabling the points with similar image coordinates to be distinguished based on their depth coordinates.}
	\label{fig:head}
	\vspace{-4mm}
\end{figure*}

Despite significant progress, existing sparse-to-dense approaches typically employ shared voxel queries (referred to as context-independent queries) across different input images. These queries are defined as a set of learnable parameters that are independent of the input context. Once the training process is completed, these parameters remain constant for all input images during inference, failing to capture distinctions among various input images, as the focal regions of different images vary. Additionally, these queries also encounter the issue of undirected feature aggregation in cross-attention, where the sampling points may fall in irrelevant regions. Besides, when projecting the 3D points onto the image plane, many points may end with the same 2D position with similar sampling points in the 2D feature map, causing a crucial depth ambiguity problem. An illustrative diagram is presented in Fig.~\ref{fig:head}\textcolor{red}{(a)}.

In this paper, we propose a context and geometry aware voxel transformer (CGVT) to lift the 2D features. We observe that context-dependent query tend to aggregate information from the points within the region of interest. Fig.~\ref{fig:sampling_offsets} presents an example of the sampling points of the context-dependent queries. Thus, before the aggregation of cross-attention, we first utilizes a context aware query generator to predict context-aware queries from the input individual images. Additionally, we extend deformable cross-attention from 2D to 3D pixel space, which allows points ending with similar image coordinates to be distinguished by their depth coordinates, as illustrated in Fig.~\ref{fig:head}\textcolor{red}{(b)}. Furthermore, we propose a simple yet efficient depth refinement block to enhance the accuracy of estimated depth probability. This involves incorporating a more precise estimated depth map from a pretrained stereo depth estimation network~\cite{MobileStereoNet}, avoiding the heavy computational burden as observed in StereoScene~\cite{StereoScene}.

Based on the aforementioned module, we devise a neural network, named as CGFormer. To enhance the obtained 3D volume from the view transformation we integrate multiple representation, \ie voxel and tri-perspective view (TPV). The TPV representation offers a global perspective that encompasses more high-level semantic information, while the voxel representation focuses more on the fine-grained structures. Drawing from the above analyses, we propose a 3D local and global encoder (LGE) that dynamically fuses the results of the voxel-based and TPV-based branches, to further enhance the 3D features from both local and global perspectives.

Our contributions can be summarized as follows:
\begin{itemize}
\item We propose a context and geometry aware voxel transformer (CGVT) to improve the performance of semantic scene completion. This module initializes the queries based on the context of individual input images and extends the deformable cross-attention from 2D to 3D pixel space, thereby improving the performance of feature lifting.
\item We introduce a simple yet effective depth refinement block to enhance the accuracy of estimated depth probability with only introducing minimal computational burden.
\item We devise a 3D local and global encoder (LGE) to strengthen the semantic and geometric discriminability of the 3D volume. This encoder employs various 3D representations (voxel and TPV) to encode the 3D features, capturing information from both local and global perspectives.
\item Benefiting from the aforementioned modules, our CGFormer attains state-of-the-art results with a mIoU of $16.63$ and an IoU of $44.41$ on SemanticKITTI, as well as a mIoU of $20.05$ and an IoU of $48.07$ on SSCBench-KITTI-360. Notably, our method even surpasses methods employing temporal images as inputs or using much larger image backbone networks.
\end{itemize}

%% file: src/related.tex
\section{Related Work}
\label{sec:related}

\subsection{Vision-based 3D Perception}
Vision-based 3D perception~\cite{bevformer, BEVFusion, BEVFusionMIT, BEVStereo, BEVDet, DETR3D, PointOcc, DFA3D, OpenOccupancy, LODE, SSCGroupConv, YanOcc, TPVD, RigNet, PointDC} has received extensive attention due to its ease of deployment, cost-effectiveness, and the preservation of intricate visual attributes, emerging as a crucial component in the autonomous driving. Current research efforts focus on constructing unified 3D representations (\textit{e.g.,} BEV, TPV, voxel) from input images. LiftSplat~\cite{lss} lifts image features by performing outer product between the 2D image features and their estimated depth probability to generate a frustum-shaped pseudo point cloud of contextual features. The pseudo point cloud features are then splatted to predefined 3D anchors through a voxel-pooling operation. Building upon this, BEVDet~\cite{BEVDet} extends LiftSplat to 3D object detection, while BEVDepth~\cite{bevdepth} further enhances performance by introducing ground truth supervision for the estimated depth probability. With the advancements in attention mechanisms~\cite{DETR, DeformableDetr, DINO, rtDETR, DETR3D}, BEVFormer~\cite{bevformer, BEVFormerv2} transforms image features into BEV features using point deformable cross-attention~\cite{DeformableDetr}. Additionally, many other methods, such as OFT~\cite{OFT}, Petr~\cite{PETR, PETRv2}, Inverse Perspective Mapping (IPM)~\cite{IPM}, have also been presented to transform 2D image features into a 3D representation.

\subsection{Semantic Scene Completion}
SSCNet~\cite{SSCNet} initially introduces the concept of semantic scene completion, aiming to infer the semantic voxels. Following methods~\cite{LMSCNet, AICNet, JS3CNet, S3CNet} commonly utilize explicit depth maps or LIDAR point clouds as inputs. MonoScene~\cite{MonoScene} is the pioneering method for directly predicting the semantic occupancy from the input RGB image, which presents a FLoSP module for 2D-3D feature projection. StereoScene\cite{StereoScene} introduces explicit epipolar constraints to mitigate depth ambiguity, albeit with heavy computational burden for correlation. TPVFormer~\cite{TPVFormer} introduces a tri-perspective view (TPV) representation to describe the 3D scene, as an alternative to the BEV representation. The elements fall into the field of view aggregates information from the image features using deformable cross-attention~\cite{bevformer, DeformableDetr}. Beginning from depth-based~\cite{VoxFormer} sparse proposal queries, VoxFormer~\cite{VoxFormer} construct the 3D representation in a coarse-to-fine manner. The 3D information from the visible queries are diffused to the overall 3D volume using deformable self-attention, akin to the masked autoencoder (MAE)\cite{MAE}. HASSC~\cite{HASSC} introduces a self-distillation training strategy to improve the performance of VoxFormer~\cite{VoxFormer}, while MonoOcc~\cite{MonoOcc} further enhance the 3D volume with an image-conditioned cross-attention module. Symphonize~\cite{Symphonize} extracts high level instance features from the image feature map, serving as the key and value of the cross-attention. 

%% file: src/method.tex
\section{CGFormer}
\label{sec:method}
\subsection{Overview}
\begin{figure*}[t]
	\centering\includegraphics[width=0.95\linewidth]{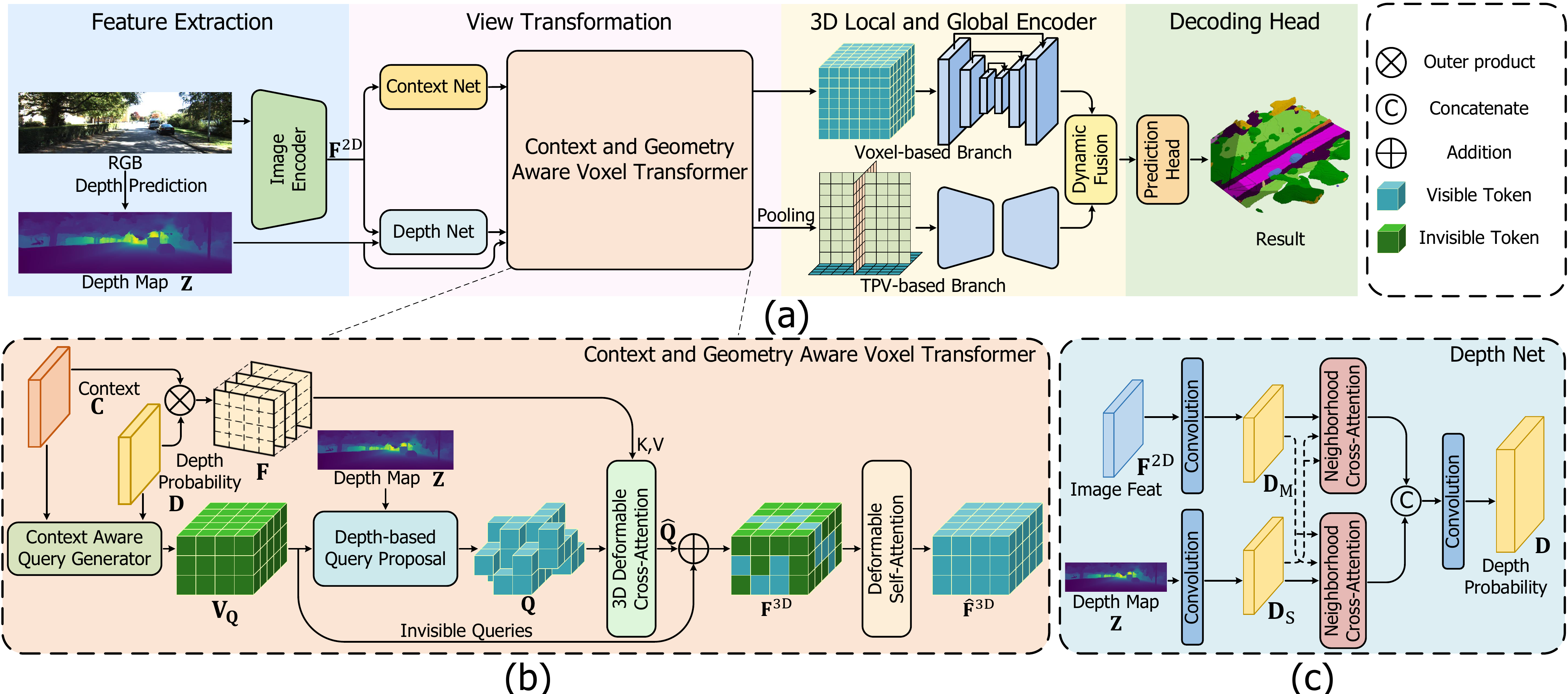}
	\caption{Schematics and detailed architectures of CGFormer. (a) The framework of the proposed CGFormer for camera-based semantic scene completion. The pipeline consists of the image encoder for extracting 2D features, the context and geometry aware voxel (CGVT) transformer for lifting the 2D features to 3D volumes, the 3D local and global encoder (LGE) for enhancing the 3D volumes and a decoding head to predict the semantic occupancy. (b) Detailed structure of the context and geometry aware voxel transformer. (c) Details of the Depth Net.}
	\label{fig:architecture}
	\vspace{-4mm}
\end{figure*}

As shown in Fig.~\ref{fig:architecture}, the overall framework of CGFormer is composed of four parts: feature extraction to extract 2D image features, view transformation (Section~\ref{sec:view_transformation}) to lift the 2D features to 3D volumes, 3D encoder (Section~\ref{sec:local_and_global_encoder}) to further enhance the semantic and geometric discriminability of the 3D features, and a decoding head to infer the final result. 

\noindent\textbf{Image Encoder.} The image encoder consists of a backbone network for extracting multi-scale features and a feature pyramid network (FPN) to fuse them. We adopt $\mathbf{F}^{\text{2D}}\in \mathbb{R}^{H \times W\times C}$ to represent the extracted 2D image feature, where $C$ is the channel number, and $(H, W)$ refers to the resolution.

\noindent\textbf{Depth Estimator.} In alignment with VoxFormer~\cite{VoxFormer}, we use an off-the-shelf stereo depth estimation model~\cite{MobileStereoNet} to predict the depth $\mathbf{Z}(u,v)$ for each image pixel $(u,v)$. The resulting estimated depth map is then employed to define the visible voxels located on the surface, serving as query proposals. Additionally, it is also used to refine the depth probability for lifting the 2D features.

\subsection{View Transformation}
\label{sec:view_transformation}

A detailed diagram of our context and geometry aware voxel transformer is presented in Fig.~\ref{fig:architecture}\textcolor{red}{(b)}. The process begins with a context aware query generator, which takes the context feature map to generate the context-dependent queries. Subsequently, the visible query proposals are located by the depth map from the pretrained depth estimation network. These proposals then attend to the image features to aggregate semantic and geometry information based on the 3D deformable cross-attention. Finally, the aggregated 3D information is further diffused from the updated proposals to the overall 3D volume via deformable self-attention.

\noindent\textbf{Context-dependent Query Generation.} Previous coarse-to-fine approaches~\cite{VoxFormer, HASSC, H2GFormer, MonoOcc} typically employ shared context-independent queries for all the inputs. However, these approaches may overlook the differences among different images and aggregate information from irrelevant areas. In contrast, CGFormer first generates context-dependent queries from the image features using a context aware query generator. To elaborate, the extracted $\mathbf{F}^{2D}$ is fed to the context net and the depth net~\cite{bevdepth} to generate the context feature $\mathbf{C}\in\mathbb{R}^{H \times W\times C}$ and depth probability $\mathbf{D}\in\mathbb{R}^{H\times W\times D}$, respectively. Then the query generator $f$ takes $\mathbf{C}$ and $\mathbf{D}$ as inputs to generate voxel queries $\mathbf{V}_{\textbf{Q}}\in\mathbb{R}^{X\times Y\times Z\times C}$
\begin{equation}
	\mathbf{V}_{\textbf{Q}}=f(\mathbf{C},\mathbf{D}),
\end{equation}
where $(X, Y, Z)$ denotes the spatial resolution of the 3D volume. Benefiting from the initialization, the sampling points of the context-dependent queries usually locate within the region of interest. An example of the deformable sampling points is displayed in Fig.~\ref{fig:sampling_offsets}.

\input{fig/SamplingOffsets}

\noindent\textbf{Depth-based Query Proposal.} Following VoxFormer~\cite{VoxFormer}, we determine the visible voxels through the conversion of camera coordinates to world coordinates using the pre-estimated depth map, except that we do not employ an additional occupancy network~\cite{LMSCNet}. A pixel $(u,v)$ will be transformed to a 3D point $(x,y,z)$, utilizing the camera intrinsic and extrinsic matrices ($K\in\mathbb{R}^{4\times 4}, E\in\mathbb{R}^{4\times 4}$). The synthetic point cloud is further converted into a binary voxel grid map $\mathbf{M}$, where each voxel is masked as $1$ if occupied by at least one point and $0$ otherwise. The query proposals are selected based on the binary $\mathbf{M}$ as
\begin{equation}
	\mathbf{Q} = \mathbf{V}_{\textbf{Q}}[\mathbf{M}].
\end{equation}

\noindent\textbf{3D Deformable Cross-Attention.} 
The query proposals then attend to the image features to aggregate the visual features of the 3D scene. We first expand the dimension of $\mathbf{C}$ from 2D to 3D pixel space by conducting the outer product between $\mathbf{D}$ and $\mathbf{C}$, formulated as $\mathbf{F}=\mathbf{C}\otimes\mathbf{D}$. Here, the operator $\otimes$ refers to the outer product conducted at the last dimension. Taking $\mathbf{F}$ as key and value, for a 3D query $\mathbf{Q}_{p}$ located at position $p$, the 3D deformable cross-attention mechanism can be formulated as
\begin{equation}
	\hat{\mathbf{Q}}_{p} = \mathrm{3D\text{-}DCA}(\mathbf{Q}_{p}, \mathbf{F}, p) = \sum_{n=1}^{N}A_{n}W\phi(\mathbf{F},\mathcal{P}(p)+\Delta{p}),
\end{equation}
where $n$ indexes the sampled point from a total of $N$ points, ${\mathcal{P}(p)}$ denotes camera projection function to obtain the reference points in the 3D pixel space, $A_{n}\in [0,1]$ is the learnable attention weight, and $W$ denotes the projection weight. $\Delta{p}\in\mathbb{R}^{3}$ is the predicted offset to the reference point $p$, and $\phi(\mathbf{F},\mathcal{P}_{d}(p)+\Delta{p_d})$ indicates the trilinear interpolation used to sample features in the expanded 3D feature map $\mathbf{F}$. $\hat{\mathbf{Q}}$ denotes the updated query proposals. For simplicity, we only show the formulation of single-head attention.

\noindent\textbf{Deformable Self-Attention.} After several layers of deformable cross-attention, we merge $\hat{\mathbf{Q}}$ and the invisible elements of $\mathbf{V}_{\textbf{Q}}$, indicated as $\mathbf{F}^{\text{3D}}$. The information of the updated query proposals is diffused to the overall 3D volume via deformable self-attention and $\hat{\mathbf{F}}^{\text{3D}}$ indicates the updated 3D volume. This process of a query located at position $p$ can be formulated as
\begin{equation}
	\hat{\mathbf{F}}^{\text{3D}}_{p} = \mathrm{DSA}(\mathbf{F}^{\text{3D}}_{p}, \mathbf{F}^{\text{3D}}, p) = \sum_{n=1}^{N}A_{n}W\phi(\mathbf{F}^{\text{3D}}, p+\Delta{p}).
\end{equation}

\noindent\textbf{Depth Net.} The depth probability serves as the inputs of both the context aware query generator and 3D deformable cross-attention, whose accuracy greatly influences the performance. The results of single image depth estimation are usually unsatisfactory. Incorporating epipolar constraint can help estimate more accurate depth map. However, computing correlation to achieve this may introduce much computation burden during the estimation of the semantic voxels. Instead, we introduce a simple yet effective depth refinement strategy. The detailed architecture of the depth net is shown in Fig.~\ref{fig:architecture}\textcolor{red}{(c)}. We first estimate monocular depth feature $\mathbf{D}_{\text{M}}$ from $\mathbf{F}^{\text{2D}}$. The depth map generated from the stereo method~\cite{MobileStereoNet} is encoded as stereo feature $\mathbf{D}_{\text{S}}$ by several convolutions. These two features are further processed by cross-attention blocks for information interaction. To save memory, the window size of cross-attention is restricted within a range of the nearest neighboring pixels~\cite{NeighborAttn}, which can be formulated as
\begin{equation}
	\begin{split}
	\hat{\mathbf{D}}_{\text{M}} & = \psi(\mathbf{D}_{\text{M}}, \mathbf{D}_{\text{S}}, \mathbf{D}_{\text{S}}),                               \\
	\hat{\mathbf{D}}_{\text{S}} & = \psi(\mathbf{D}_{\text{S}}, \mathbf{D}_{\text{M}}, \mathbf{D}_{\text{M}}),
	\label{eq:neighbor_attntion}
	\end{split}
\end{equation}
where $\hat{\mathbf{D}}_{\text{M}}$ and $\hat{\mathbf{D}}_{\text{S}}$ denote the enhanced $\mathbf{D}_{\text{M}}$ and $\mathbf{D}_{\text{S}}$, and $\psi$ denotes the neighborhood attention~\cite{NeighborAttn}. We set the window size to $5\times 5$. Finally, these two volumes are concatenated and fed to estimate the final depth probability. This strategy can boost the accuracy of the depth probability a lot with only minimal computation cost.

\subsection{3D Local and Global Encoder}
\label{sec:local_and_global_encoder}
The TPV representation is derived by compressing the voxel representation along its three axes. TPV focuses more on the global high-level semantic information, whereas the voxel emphasizes more on the details of intricate structures. We combine these two representations to enhance representation capability of the 3D features. As shown in Fig.~\ref{fig:architecture}\textcolor{red}{(a)}, our 3D local and global encoder consists of the voxel-based and TPV-based branches. With the input 3D volume, the two branches first aggregate features in parallel. The dual-branch outputs are then fused dynamically and fed to the final decoding block.

\noindent\textbf{Voxel-based Branch.} The voxel-based branch aims to enhance the fine-grained structure features of the 3D volume. We adopt a lightweight ResNet~\cite{ResNet} with 3D convolutions to extract multi-scale voxel features from $\hat{\mathbf{F}}^{\text{3D}}$ and merge them into $\mathbf{F}^{\text{3D}}_{\text{voxel}}\in\mathbb{R}^{X\times Y\times Z\times C}$ using a 3D feature pyramid network.

\noindent\textbf{TPV-based Branch.} To obtain three-plane features of the TPV representation, we apply spatial pooling to the voxel features along three axes. Directly performing MaxPooling may cause the loss of the fine-grained details. Considering the balance between performance and efficiency, we apply the group spatial to channel operation~\cite{S2C, BEVFusion, PointOcc} to transform the voxel representation to TPV representation. Specifically, we split the voxel features along the pooling axis into $K$ groups and apply the MaxPooling within each group. Then we concatenate each group features along the channel dimension and use convolutions to reduce the channel dimension into $C$. Denote the 2D view features as $\mathbf{F}^{\text{3D}}_{XY}\in\mathbb{R}^{X\times Y\times C}, \mathbf{F}^{\text{3D}}_{XZ}\in\mathbb{R}^{X\times Z\times C}, \mathbf{F}^{\text{3D}}_{YZ}\in\mathbb{R}^{Y\times Z\times C}$, the above process can be formulated as
\begin{equation}
	\begin{split}
		\mathbf{F}^{\text{3D}}_{XY}&=\mathrm{Conv}_{XY}(\mathrm{Concat}(\{\mathrm{Pooling}_{\{Z,i\}}(\hat{\mathbf{F}}^{\text{3D}})\}^{K}_{i=1})),\\
		\mathbf{F}^{\text{3D}}_{YZ}&=\mathrm{Conv}_{YZ}(\mathrm{Concat}(\{\mathrm{Pooling}_{\{X,i\}}(\hat{\mathbf{F}}^{\text{3D}})\}^{K}_{i=1})), \\
		\mathbf{F}^{\text{3D}}_{XZ}&=\mathrm{Conv}_{XZ}(\mathrm{Concat}(\{\mathrm{Pooling}_{\{Y,i\}}(\hat{\mathbf{F}}^{\text{3D}})\}^{K}_{i=1}))
	\end{split}
\end{equation}
After obtaining the TPV features, we employ a 2D image backbone to process each plane to produce corresponding multi-scale features, which are then fused by a 2D feature pyramid network. For simplicity, we leverage $\mathbf{F}^{\text{3D}}_{XY}, \mathbf{F}^{\text{3D}}_{XZ}, \mathbf{F}^{\text{3D}}_{YZ}$ to represent the outputs of these networks, as well.

\noindent\textbf{Dynamic Fusion.} The dual-path outputs are fused dynamically. We generate aggregation weights $\mathbf{W}\in\mathbb{R}^{X\times Y\times Z\times 4}$ from $\mathbf{F}^{\text{3D}}_{\text{voxel}}$ using a convolution with a softmax on the last dimension. The final 3D feature volume $\mathbf{F}^{\text{3D}}_{\text{final}}$ is computed as
\begin{equation}
	\mathbf{F}^{\text{3D}}_{\text{final}} = \sum_{i}^{4} \mathbf{W}_{i}\odot\mathbf{F}^{\text{3D}}_{i},
\end{equation}
where $\mathbf{W}_{i}\in\mathbb{R}^{X\times Y\times Z\times 1}$ is a slice of the $\mathbf{W}$ and $\mathbf{F}^{\text{3D}}_{i}$ is an element of the set $[\mathbf{F}^{\text{3D}}_{\text{voxel}}, \mathbf{F}^{\text{3D}}_{XY}, \mathbf{F}^{\text{3D}}_{XZ}, \mathbf{F}^{\text{3D}}_{YZ}]$. The operation $\odot$ refers to the element multiplication and the dimension broadcasting of the features from TPV-based branches is omitted in the equation for conciseness. $\mathbf{F}^{\text{3D}}_{\text{final}}$ is finally fed to the decoding block to infer the complete scene geometry and semantics. Please refer to the supplementary material for more detailed schematics of the 3D local and global encoder.
\subsection{Training Loss}
To train the proposed CGFormer, cross-entropy loss weighted by class frequencies $\mathcal{L}_{ce}$ is leveraged to optimize the network. Following MonoScene~\cite{MonoScene}, we also utilize affinity loss $\mathcal{L}_{scal}^{geo}$ and $\mathcal{L}_{scal}^{sem}$ to optimize the scene-wise and class-wise metrics (geometric IoU and semantic mIoU). We also employ explicit depth loss $\mathcal{L}_{d}$~\cite{bevformer} to supervise the estimated depth probability of the depth net. Hence, the overall loss function can be formulated as
\begin{equation}
	\mathcal{L} = \lambda_{d}\mathcal{L}_{d} + \mathcal{L}_{ce} + \mathcal{L}_{scal}^{geo} + \mathcal{L}_{scal}^{sem},
\end{equation}
where we set the weight of depth loss $\lambda_{d}$ to $0.001$.

%% file: fig/SamplingOffsets.tex
\begin{figure*}
	\centering
	\newcolumntype{P}[1]{>{\centering\arraybackslash}m{#1}}
	\setlength{\tabcolsep}{0.001\textwidth}
	\footnotesize
	\begin{tabular}{P{0.3\textwidth} P{0.3\textwidth} P{0.3\textwidth} } 		
		\includegraphics[width=1.\linewidth]{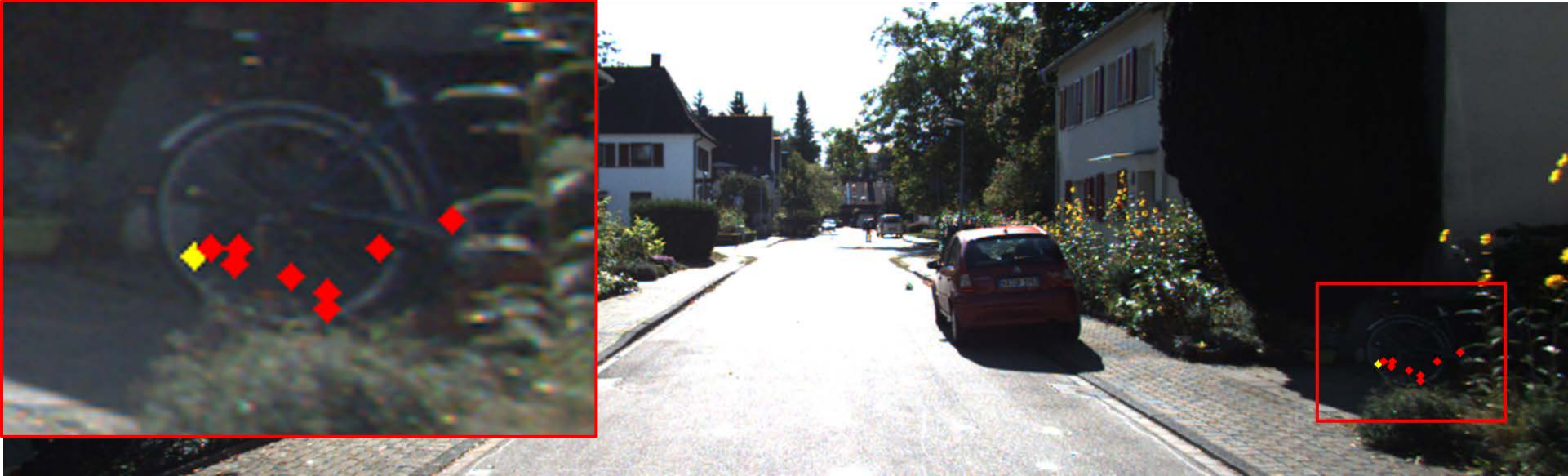} &
		\includegraphics[width=.8\linewidth]{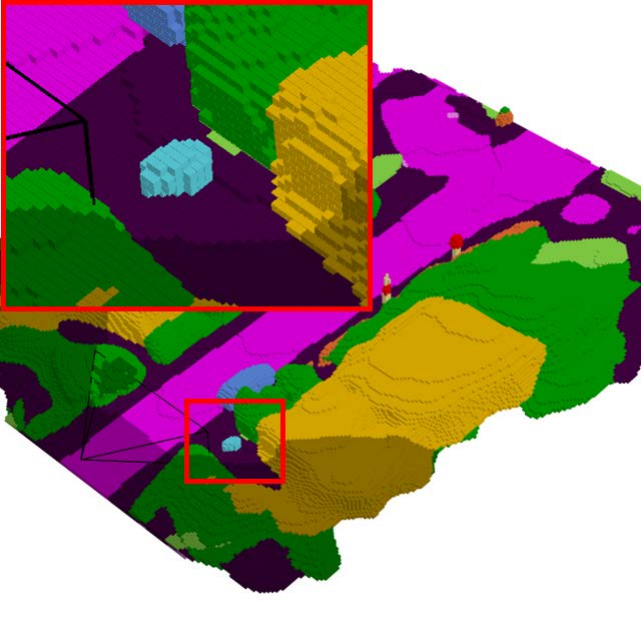} &
		\includegraphics[width=.8\linewidth]{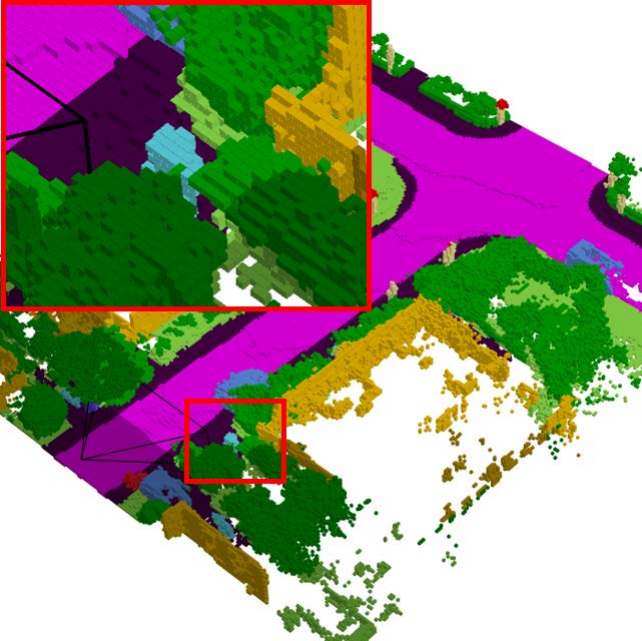}
		\\
		\includegraphics[width=1.05\linewidth]{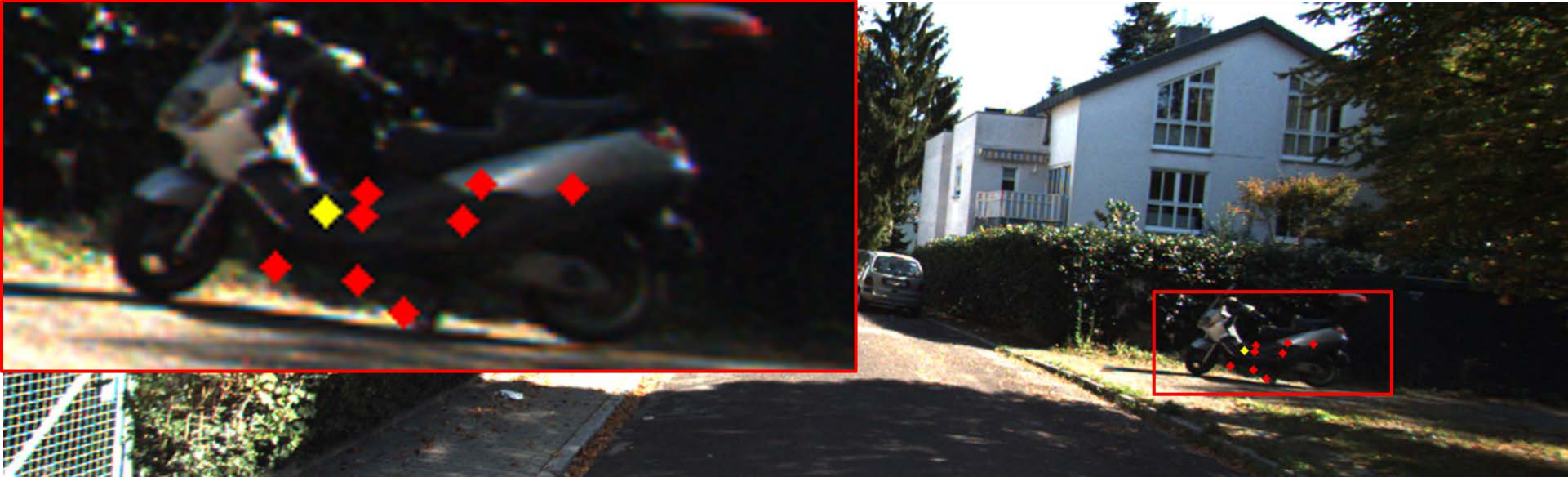} &
		\includegraphics[width=.8\linewidth]{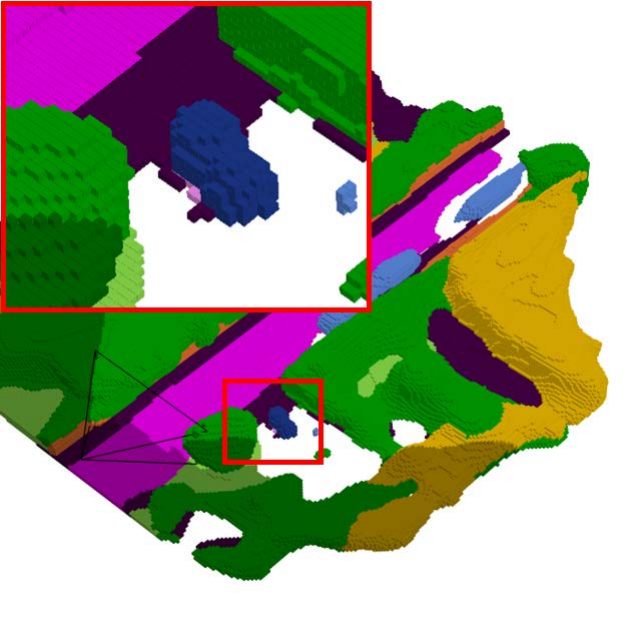} &
		\includegraphics[width=.8\linewidth]{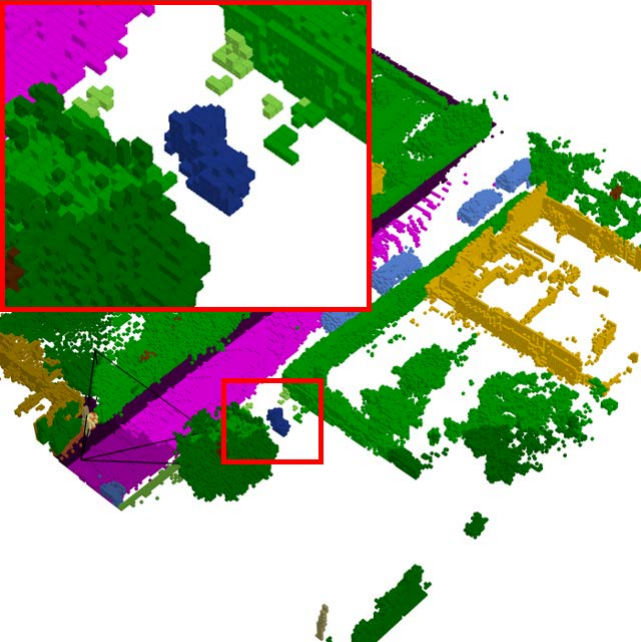}
		\\
		(a) RGB & (b) CGFormer (ours) & (c) Ground Truth \\
	\end{tabular}
	\caption{Visualization of the sampling locations for different small objects. The yellow dot represents the query point, while the red dots indicate the locations of the deformable sampling points. The sampling points of the context-dependent query (a) tend to be distributed within the regions of interest. Beneficial from this, CGFormer achieve better performance than previous methods.}
	\label{fig:sampling_offsets}
	\vspace{-4mm}
\end{figure*} 

%% file: src/experiments.tex
\section{Experiments}
\subsection{Quantitative Results}
\input{tab/tab_sem_test}
\input{tab/tab_kitti_360_test}
We conducted experiments to evaluate the performance of our CGFormer with the latest state-of-the-art methods on the SemanticKITTI~\cite{SemanticKITTI} and SSCBench-KITTI-360~\cite{SSCBench}. Refer to the supplement material for information of datasets, metrics and detailed implementation details.

We list the results on SemanticKITTI hidden test set in Table~\ref{tab:sem_kitti_test}. CGFormer achieves a mIoU of $16.63$ and an IoU of $44.41$. Notably, CGFormer outperforms all competing methods in terms of mIoU, demonstrating its superior performance. Regarding IoU, CGFormer surpasses all other methods except DepthSSC~\cite{DepthSSC}. While ranking 2nd in terms of IoU, CGFormer exhibits only a slight marginal difference of $0.17$ in comparison to DepthSSC~\cite{DepthSSC}. However, CGFormer significantly outperforms DepthSSC~\cite{DepthSSC} by a substantial margin of $3.52$ in terms of mIoU. Additionally, compared to VoxFormer-T~\cite{VoxFormer}, HASSC-T~\cite{HASSC}, H2GFormer-t~\cite{H2GFormer}, and MonoOcc-L~\cite{MonoScene}, although they utilize temporal stereo image pairs as inputs or larger image backbone networks, our CGFormer exceeds them a lot in terms of both metrics. For specific instances, CGFormer achieves the best or second-best performance on most of the classes, such as road, sidewalk, parking, and building.

To demonstrate the versatility of our model, we also conduct experiments on the SSCBench-KITTI-360~\cite{SSCNet} dataset and list the results in Table~\ref{tab:kitti_360_test}. It is observed that CGFormer surpasses all the published camera-based methods by a margin of $3.95$ IoU and $1.67$ mIoU. Notably, CGFormer even outperforms the two LiDAR-based methods in terms of mIoU. The above analyses further verifies the effectiveness and excellent performance of CGFormer.

\subsection{Ablation Study}
We conduct ablation analyses on the components of CGFormer on the SemanticKITTI validation set. 

\input{tab/tab_ablation_arch}
\input{tab/tab_ablation_addition}

\noindent\textbf{Ablation on the context and geometry aware voxel transformer (CGVT).} Table~\ref{ablation:arch} presents a detailed analysis of various architectural components within CGFormer. The baseline can be considered as a simplified version of VoxFormer~\cite{VoxFormer}, without utilizing the extra occupancy network~\cite{LMSCNet} to generate coarse occupancy masks. It begins with an image encoder to extract image features, a depth-based query proposal layer to define the visible queries, the deformable cross-attention layer to aggregate features for visible areas, the deformable self-attention layer to diffuse features from the visible regions to the invisible ones. Extending the cross-attention to 3D deformable cross-attention (a) brings a notable improvement of 1.63 mIoU and 2.15 IoU. The performance is further enhanced by giving the voxel queries a good initialization by introducing the context ware query generator (b).

\noindent\textbf{Ablation on the local and global encoder (LGE).} After obtaining the 3D features, CGFormer incorporates multiple representation to refine the 3D volume form both local and global perspectives. Through experimentation, we validate that the voxel-based and TPV-based branches collectively contribute to performance improvement with a suitable fusion strategy. Specifically, we compate the results of solely utilizing the local voxel-based branch (c), simply adding the results of dual branches (d), and dynamically fusing the dual-branch outputs (h), as detailed in Table~\ref{ablation:arch}. The accuracy is notably elevated to an IoU of $45.99$ and mIoU of $16.87$ by dynamically fusing the dual-branch outputs. Additionally, we conduct an ablation study on the three TPV planes utilized in the TPV-based branch (e,f,g). The results demonstrate that any individual plane improves performance compared to the model with only the local branch. Combining the information from all three planes into the TPV representation achieves superior performance, underscoring the complementary nature of the three TPV planes in effectively representing complex 3D scenes.

\noindent\textbf{Ablation on the context aware query generator.} We present the ablation analysis of the context-aware query generator in Table~\ref{ablation:CAQG}. We remove the CAQG and increase the number of attention layers, where the results of the previous layers can be viewed as a initialization of the queries for the later layers. This configuration (6 cross-attention layers and 4 self-attention layers) significantly improves IoU but only marginally lifts mIoU, and it requires much more training memory. Employing FLoSP~\cite{MonoScene} with the occupancy-aware depth module~\cite{OccDepth} can improve performance without much additional computational burden. By replacing it with the voxel pooling~\cite{lss}, the model achieves the best performance. Thus, we employ voxel pooling as our context aware query generator.

\noindent\textbf{Ablation on the depth refinement block.} Table~\ref{ablation:DR} presents analyses of the impact of each module within the depth net. By removing the stereo feature $\mathbf{D}_{S}$ and employing the same structure as BEVDepth~\cite{bevdepth}, we observe a performance drop of $1.42$ IoU and $1.79$ mIoU. When incorporating the stereo feature $\mathbf{D}_{S}$ but fusing it with a simple concatenate operation without using the neighborhood attention, the model achieves a mIoU of $45.72$ and an IoU of $16.26$. These results emphasize that deactivating any component of the depth net leads to a decrease of the accuracy of the full network. Additionally, we replace the depth refinement module with the dense correlation module from StereoScene~\cite{StereoScene}. Compared to this, our depth refinement module achieves comparable results while using significantly fewer parameters and less training memory.

\subsection{Qualitative Results} 
\input{fig/QualitativeResults}

Fig.~\ref{fig:visualization} presents visualizations of predicted results on the SemanticKITTI validation set obtained from MonoScene\cite{MonoScene}, VoxFormer~\cite{VoxFormer}, OccFormer~\cite{OccFormer}, and our proposed CGFormer. Notably, CGFormer outperforms other methods by effectively capturing the semantic scene and inferring previously invisible regions. The predictions generated by CGFormer exhibit distinctly clearer geometric structures and improved semantic discrimination, particularly for classes like cars and the overall scene layout. This enhancement is attributed to the precision achieved through the context and geometry aware voxel transformer applied in our proposed approach. In contrast, the instances generated by the comparison methods appear to be influenced by depth ambiguity, resulting in vague shapes.

%% file: tab/tab_sem_test.tex
\begin{table}
	\newcommand{\clsname}[2]{
		\rotatebox{90}{
			\hspace{-6pt}
			\textcolor{#2}{$\blacksquare$}
			\hspace{-6pt}
			\renewcommand\arraystretch{0.6}
			\begin{tabular}{l}
				#1                                      \\
				\hspace{-4pt} ~\tiny(\semkitfreq{#2}\%) \\
			\end{tabular}
	}}
	\centering\huge
	\caption{Quatitative results on SemanticKITTI~\cite{SemanticKITTI} test set. $^\ast$ represents the reproduced results in ~\cite{TPVFormer, OccFormer}. The best and the second best results are in \textbf{bold} and \underline{underlined}, respectively.}
	\resizebox{\linewidth}{!}
	{
		\begin{tabular}{c|cc|ccccccccccccccccccc}
			\toprule		
			Method 								 & 
			IoU 								 & 
			mIoU  								 &  
			\clsname{road}{road}                 & 
			\clsname{sidewalk}{sidewalk}         &        
			\clsname{parking}{parking}           & 
			\clsname{other-grnd.}{otherground}   & 
			\clsname{building}{building}         & 
			\clsname{car}{car} 					 & 
			\clsname{truck}{truck}               &
			\clsname{bicycle}{bicycle}           &
			\clsname{motorcycle}{motorcycle}     &
			\clsname{other-veh.}{othervehicle}   &
			\clsname{vegetation}{vegetation}     &
			\clsname{trunk}{trunk}               &
			\clsname{terrain}{terrain}           &
			\clsname{person}{person}             &
			\clsname{bicyclist}{bicyclist}       &
			\clsname{motorcyclist}{motorcyclist} &
			\clsname{fence}{fence}               &
			\clsname{pole}{pole}                 &
			\clsname{traf.-sign}{trafficsign}
			\\
			\midrule
			MonoScene$^\ast$~\cite{MonoScene} & 34.16 & 11.08 & 54.70 & 27.10 & 24.80 & 5.70
			& 14.40 & 18.80 & 3.30 & 0.50 & 0.70 & 4.40  & 14.90 & 2.40  & 19.50 & 1.00  & 1.40
			& 0.40  & 11.10 & 3.30 & 2.10         \\
			TPVFormer~\cite{TPVFormer}        &34.25 & 11.26 & 55.10 & 27.20 & 27.40 & 6.50
			& 14.80 & 19.20 & 3.70 & 1.00 & 0.50 & 2.30  & 13.90 & 2.60  & 20.40 & 1.10  & 2.40
			& 0.30  & 11.00 & 2.90 & 1.50 		  \\
			SurroundOcc~\cite{surroundOcc}    & 34.72 & 11.86 & 56.90 & 28.30 & 30.20 & 6.80 
			& 15.20 & 20.60 & 1.40 & 1.60 & 1.20 & 4.40  & 14.90 & 3.40  & 19.30 & 1.40  & 2.00
			& 0.10  & 11.30 & 3.90 & 2.40         \\
			OccFormer~\cite{OccFormer}        & 34.53 & 12.32 & 55.90 & 30.30 & \underline{31.50} & 6.50          
			& 15.70 & 21.60 & 1.20 & 1.50 & 1.70 & 3.20  & 16.80 & 3.90  & 21.30 & 2.20  & 1.10
			& 0.20  & 11.90 & 3.80 & 3.70         \\
			IAMSSC~\cite{IAMSSC}  & 43.74 & 12.37 & 54.00 & 25.50 & 24.70 & 6.90 & 19.20 & 21.30 & 3.80 & 1.10 & 0.60 & 3.90 & 22.70 & 5.80 & 19.40 & 1.50 & 2.90 & 0.50 & 11.90 & 5.30 & 4.10 \\
			VoxFormer-S~\cite{VoxFormer}      & 42.95 & 12.20 & 53.90 & 25.30 & 21.10 & 5.60
			& 19.80 & 20.80 & 3.50 & 1.00 & 0.70 & 3.70  & 22.40 & 7.50  & 21.30 & 1.40  & 2.60 
			& 0.20  & 11.10 & 5.10 & 4.90         \\
			VoxFormer-T~\cite{VoxFormer}       & 43.21 & 13.41 & 54.10 & 26.90 & 25.10 & 7.30 & 23.50 & 21.70 & 3.60 & 1.90 & 1.60 & 4.10 & 24.40 & 8.10 & 24.20 & 1.60 & 1.10 & 0.00 & 13.10 & 6.60 & 5.70 \\
			DepthSSC~\cite{DepthSSC}          & \textbf{44.58} & 13.11 & 55.64 & 27.25 & 25.72 & 5.78
			& 20.46 & 21.94 & 3.74 & 1.35 & 0.98 & 4.17  & 23.37 & 7.64  & 21.56 & 1.34  & 2.79
			& 0.28  & 12.94 & 5.87 & 6.23         \\
			Symphonize~\cite{Symphonize}	  & 42.19 & 15.04 & 58.40 & 29.30 & 26.90 & \underline{11.70}
			& \underline{24.70} & 23.60 & 3.20 & 3.60 & \textbf{2.60} & 5.60  & 24.20 & 10.00 & 23.10 & \textbf{3.20}  & 1.90  
			& \textbf{2.00}  & 16.10 & \underline{7.70} & 8.00         \\
			HASSC-S~\cite{HASSC} & 43.40 & 13.34 & 54.60 & 27.70 & 23.80 & 6.20 & 21.10 & 22.80 & 4.70 & 1.60 & 1.00 & 3.90 & 23.80 & 8.50 & 23.30 & 1.60 & \underline{4.00} & 0.30 & 13.10 & 5.80 & 5.50 \\
			HASSC-T~\cite{HASSC}  & 42.87 & 14.38 & 55.30 & 29.60 & 25.90 & 11.30 & 23.10 & 23.00 & 2.90 & 1.90 & 1.50 & 4.90 & 24.80 & 9.80 & 26.50 & 1.40 & 3.00 & 0.00 & 14.30 & 7.00 & 7.10 \\
			StereoScene~\cite{StereoScene}  & 43.34 & 15.36 & \underline{61.90} & \underline{31.20} & 30.70 & 10.70 & 24.20 & 22.80 & 2.80 &  3.40 & \underline{2.40} & \underline{6.10} & 23.80 & 8.40 & \underline{27.00} & \underline{2.90} & 2.20 & 0.50 & 16.50 & 7.00 & 7.20 \\
			H2GFormer-S~\cite{H2GFormer} & 44.20 & 13.72 & 56.40 & 28.60 & 26.50 & 4.90 & 22.80 & 23.40 & 4.80 & 0.80 & 0.90 & 4.10 & 24.60 & 9.10 & 23.80 & 1.20 & 2.50 & 0.10 & 13.30 & 6.40 & 6.30 \\
			H2GFormer-T~\cite{H2GFormer}  & 43.52 & 14.60 & 57.90 & 30.40 & 30.00 & 6.90 & 24.00 & 23.70 & \underline{5.20} & 0.60 & 1.20 & 5.00 & \textbf{25.20} & \underline{10.70} & 25.80 & 1.10 & 0.10 & 0.00 & 14.60 & 7.50 & \textbf{9.30} \\
			MonoOcc-S~\cite{MonoOcc}  & - & 13.80 & 55.20 & 27.80 & 25.10 & 9.70 & 21.40 & 23.20 & \underline{5.20} & 2.20 & 1.50 & 5.40 & 24.00 & 8.70 & 23.00 & 1.70 & 2.00 & 0.20 & 13.40 & 5.80 & 6.40 \\
			MonoOcc-L~\cite{MonoOcc} & - & \underline{15.63} & 59.10 & 30.90 & 27.10 & 9.80 & 22.90 & \underline{23.90} & \textbf{7.20} & \textbf{4.50} & \underline{2.40} & \textbf{7.70} & \underline{25.00} & 9.80 & 26.10 & 2.80 & \textbf{4.70} & \underline{0.60} & \underline{16.90} & 7.30 & \underline{8.40} \\
			\hline
			CGFormer (ours) & \underline{44.41} & \textbf{16.63} & \textbf{64.30} & \textbf{34.20} & \textbf{34.10} & \textbf{12.10} & \textbf{25.80} & \textbf{26.10} & 4.30 & \underline{3.70} & 1.30 & 2.70 & 24.50 & \textbf{11.20} & \textbf{29.30} & 1.70 & 3.60 & 0.40 & \textbf{18.70} & \textbf{8.70} & \textbf{9.30} \\
			\bottomrule
		\end{tabular}
	}
	\setlength{\abovecaptionskip}{0cm}
	\setlength{\belowcaptionskip}{0cm}
	\label{tab:sem_kitti_test}
	\vspace{-4mm}
\end{table}

%% file: tab/tab_kitti_360_test.tex
\begin{table}
	\newcommand{\clsname}[2]{
		\rotatebox{90}{
			\hspace{-6pt}
			\textcolor{#2}{$\blacksquare$}
			\hspace{-6pt}
			\renewcommand\arraystretch{0.6}
			\begin{tabular}{l}
				#1                                       \\
				\hspace{-4pt} ~\tiny(\sscbkitfreq{#2}\%) \\
			\end{tabular}
	}}
	\centering\huge
	\caption{Quantitative results on SSCBench-KITTI360 test set. The results for counterparts are provided in \cite{SSCBench}. The best and the second best results for all camera-based methods are in \textbf{bold} and \underline{underlined}, respectively. The best results from the LiDAR-based methods are in \textcolor{red}{red}.}
	\resizebox{\linewidth}{!}
	{
		\begin{tabular}{c|cc|cccccccccccccccccc}
			\toprule
			Method                            			&
			IoU                                 		&
			mIoU                                		&
			\clsname{car}{car}                      	&
			\clsname{bicycle}{bicycle}              	&
			\clsname{motorcycle}{motorcycle}        	&
		    \clsname{truck}{truck}                   	&
			\clsname{other-veh.}{othervehicle}      	&
			\clsname{person}{person}                	&
			\clsname{road}{road}                    	&
			\clsname{parking}{parking}              	&
			\clsname{sidewalk}{sidewalk}            	&
			\clsname{other-grnd.}{otherground}      	&
			\clsname{building}{building}            	&
			\clsname{fence}{fence}                  	&
			\clsname{vegetation}{vegetation}        	&
			\clsname{terrain}{terrain}              	&
			\clsname{pole}{pole}                    	&
			\clsname{traf.-sign}{trafficsign}       	&
			\clsname{other-struct.}{otherstructure} 	&
			\clsname{other-obj.}{otherobject}
			\\
			\midrule
			\multicolumn{21}{l}{\textit{LiDAR-based methods}}                                                                                                                                                                                                                                                                                                                                               \\
			\hline
			SSCNet~\cite{SSCNet}        & \textcolor{red}{53.58} & 16.95 & \textcolor{red}{31.95} & 0.00 & 0.17 & 10.29 & 0.00 & 0.07 & \textcolor{red}{65.70}
			& \textcolor{red}{17.33} & \textcolor{red}{41.24} & 3.22 & \textcolor{red}{44.41} & 6.77 & \textcolor{red}{43.72} & \textcolor{red}{28.87} & 0.78  & 0.75 & 8.69 & 0.67 \\
			LMSCNet~\cite{LMSCNet}      & 47.35 & 13.65 & 20.91 & 0.00 & 0.00 & 0.26  & 0.58 & 0.00 & 62.95        
			& 13.51 & 33.51 & 0.20 & 43.67 & 0.33 & 40.01 & 26.80 & 0.00  & 0.00 & 3.63 & 0.00 \\
			\specialrule{0.7pt}{0pt}{0pt}
			\multicolumn{21}{l}{\textit{Camera-based methods}}                                                                                                                                                                                                                                                                                                                                              \\
			\hline
			MonoScene~\cite{MonoScene}  & 37.87 & 12.31 & 19.34 & 0.43 & 0.58 & 8.02  & 2.03 & 0.86 & 48.35        
			& 11.38 & 28.13 & 3.32 & 32.89 & 3.53 & 26.15 & 16.75 & 6.92 & 5.67 & 4.20 & 3.09  \\
			TPVFormer~\cite{TPVFormer}  & 40.22 & 13.64 & 21.56 & 1.09 & 1.37 & 8.06  & 2.57 & 2.38 & 52.99        
			& 11.99 & 31.07 & 3.78 & 34.83 & 4.80 & 30.08 & 17.52 & 7.46 & 5.86 & 5.48 & 2.70  \\
			OccFormer~\cite{OccFormer}  & 40.27 & 13.81 & 22.58 & 0.66 & 0.26 & 9.89  & 3.82 & 2.77 & 54.30       
			& 13.44 & 31.53 & 3.55 & 36.42 & 4.80 & 31.00 & 19.51 & 7.77 & 8.51 & 6.95 & 4.60  \\
			VoxFormer~\cite{VoxFormer}  & 38.76 & 11.91 & 17.84 & 1.16 & 0.89 & 4.56  & 2.06 & 1.63 & 47.01
			& 9.67  & 27.21 & 2.89 & 31.18 & 4.97 & 28.99 & 14.69 & 6.51 & 6.92 & 3.79 & 2.43  \\
			IAMSSC~\cite{IAMSSC}   & 41.80 & 12.97 & 18.53 & \underline{2.45} & 1.76 & 5.12 & 3.92 & 3.09 & 47.55 & 10.56 & 28.35 & 4.12 & 31.53 & 6.28 & 29.17 & 15.24 & 8.29 & 7.01 & 6.35 & 4.19 \\
			DepthSSC~\cite{DepthSSC}    & 40.85 & 14.28 & 21.90 & 2.36 & \underline{4.30} & 11.51 & 4.56 & 2.92 & 50.88
			& 12.89 & 30.27 & 2.49 & \underline{37.33} & 5.22 & 29.61 & \underline{21.59} & 5.97 & 7.71 & 5.24 & 3.51  \\
			Symphonies~\cite{Symphonize} & \underline{44.12} & \underline{18.58} & \textbf{30.02} & 1.85 & \textbf{5.90} & \textbf{25.07} & \textbf{12.06} & \textbf{8.20} & \underline{54.94}
			& \underline{13.83} & \underline{32.76} & \textbf{6.93} & 35.11 & \textbf{8.58} & \underline{38.33} & 11.52 & \underline{14.01} & \underline{9.57} & \textbf{14.44} & \textbf{11.28} \\
			\hline
			CGFormer (ours) & \textbf{48.07} & \textbf{20.05} & \underline{29.85} & \textbf{3.42} & 3.96 & \underline{17.59} & \underline{6.79} & \underline{6.63} & \textbf{63.85} & \textbf{17.15} & \textbf{40.72} & \underline{5.53} & \textbf{42.73} & \underline{8.22} & \textbf{38.80} & \textbf{24.94} & \textbf{16.24} & \textbf{17.45} & \underline{10.18} & \underline{6.77} \\
			\bottomrule
		\end{tabular}
	}
	\setlength{\abovecaptionskip}{0cm}
	\setlength{\belowcaptionskip}{0cm}
	\label{tab:kitti_360_test}
	\vspace{-4mm}
\end{table}

%% file: tab/tab_ablation_arch.tex
\begin{table*}[t]
	\centering
	%\Large
	\renewcommand\arraystretch{1}
	\caption{Ablation study of the architectural components on SemanticKITTI~\cite{SemanticKITTI} validation set. CGVT: context and geometry aware voxel transformer. LGE: local and global encoder. 3D-DCA: 3D deformable cross attention. CAQG: context aware query generator. LB: local voxel-based branch. $\mathcal{T}_{XY}$, $\mathcal{T}_{YZ}$, $\mathcal{T}_{XZ}$: planes of the TPV-based branch. DF: dynamic fusion. There are 32M predefined parameters.}
	\resizebox{\textwidth}{!}{
		\begin{tabular}{l|cc|ccccc|cc|cc}
			\toprule
			\multirow{2}{*}{Method}  &  \multicolumn{2}{c|}{CGVT} & \multicolumn{5}{c|}{LGE}  & \multirow{2}{*}{IoU$\uparrow$} & \multirow{2}{*}{mIoU$\uparrow$} & \multirow{2}{*}{Params (M)}  & \multirow{2}{*}{Memory (M)} \\ \cline{2-8}
			& 3D-DCA & CAQG & LB & $\mathcal{T}_{XY}$ & $\mathcal{T}_{YZ}$ & $\mathcal{T}_{XZ}$ & DF & & & & \\
			\hline
			Baseline &            &            &            &            &            & 		   & 		    & 37.99  & 12.71 & 76.57  & 13222  \\ 
			(a)      & \checkmark &            &            &            &            &            &            & 40.14  & 14.34 & 86.17  & 15150  \\
			(b)      & \checkmark & \checkmark &            &            &            &            &            & 42.86  & 15.60 & 86.19  & 15488  \\ \hline
			(c)      & \checkmark & \checkmark & \checkmark &            &            &            &            & 44.84  & 16.41 & 93.78  & 17843  \\
			(d)      & \checkmark & \checkmark & \checkmark & \checkmark & \checkmark & \checkmark &            & 44.63  & 16.54 & 122.42 & 19188  \\
			(e)      & \checkmark & \checkmark & \checkmark & \checkmark &            &            & \checkmark & 45.46  & 16.38 & 122.12 & 19024  \\
			(f)      & \checkmark & \checkmark & \checkmark &            & \checkmark &            & \checkmark & 45.53  & 16.74 & 122.12 & 18912  \\
			(g)      & \checkmark & \checkmark & \checkmark &            &            & \checkmark & \checkmark & 45.71  & 16.49 & 122.12 & 18912  \\ \hline
			(h)      & \checkmark & \checkmark & \checkmark & \checkmark & \checkmark & \checkmark & \checkmark & \textbf{45.99} & \textbf{16.87} & 122.42 & 19330 \\ 
			\bottomrule
		\end{tabular}
	}
	\label{ablation:arch}
	\vspace{-4mm}
\end{table*}

%% file: tab/tab_ablation_addition.tex
\begin{table*}[t]
	\centering
	\small
	\begin{minipage}[t]{0.47\linewidth}
		\centering
		\caption{Ablation on the choices of context aware query generator.}
		\resizebox{1\textwidth}{!}{
		\begin{tabular}{c|cc|cc}
			\toprule
			Method & IoU & mIoU & Params (M) & Memory (M) \\ \hline
			w/o CAQG & 40.14 & 14.34 & 86.17 & 15150 \\
			More attention layers & 41.83 & 14.43 & 86.97 & 21596 \\
			FLoSP~\cite{MonoScene} & 41.54 & 14.66 & 86.19 & 15907 \\
			Voxel Pooling & \textbf{42.86} & \textbf{15.60} & 86.19 & 15488 \\
			\bottomrule
		\end{tabular}}
		\label{ablation:CAQG}
	\end{minipage}
	\hfill
	\begin{minipage}[t]{0.47\linewidth}
		\centering
		\caption{Ablation on the depth refinement block of the depth net.}
		\resizebox{1\textwidth}{!}{
			\begin{tabular}{c|cc|cc}
				\toprule
				Model & IoU & mIoU & Params (M) & Memory (M)  \\ \hline
				w/o stereo feature $\mathbf{D}_{S}$ 	& 44.57 & 15.08 & 116.44 & 18547 \\
				w/o neighborhood attn. 					& 45.72 & 16.26 & 122.32 & 19260 \\
				w / StereoScene~\cite{StereoScene} 		& 45.55 & 16.49 & 156.50 & 22347 \\ \hline
				full model 								& 45.99 & 16.87 & 122.42 & 19330 \\
				\bottomrule
		\end{tabular}}
		\label{ablation:DR}
	\end{minipage}
	\vspace{-4mm}
\end{table*}

%% file: fig/QualitativeResults.tex
\begin{figure*}
	\centering\tiny
	\newcolumntype{P}[1]{>{\centering\arraybackslash}m{#1}}
	\renewcommand{\arraystretch}{0.8}
	\footnotesize
	\begin{tabular}{P{0.16\textwidth} P{0.18\textwidth} P{0.18\textwidth} P{0.19\textwidth} P{0.17\textwidth}}
		\includegraphics[width=.9\linewidth]{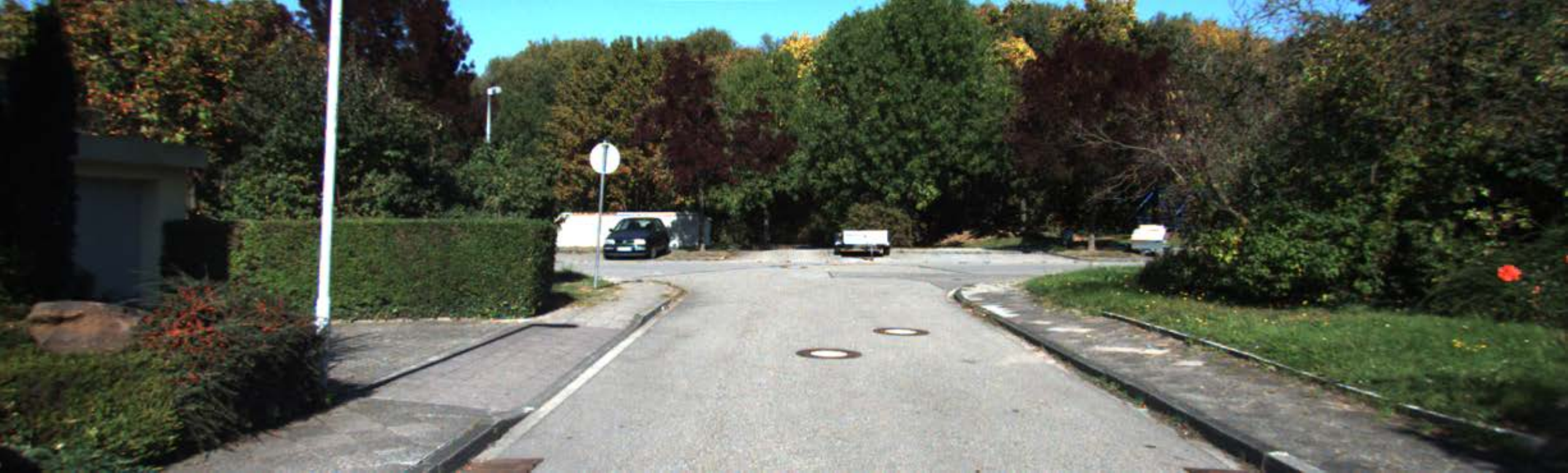} &		
		\includegraphics[width=.9\linewidth]{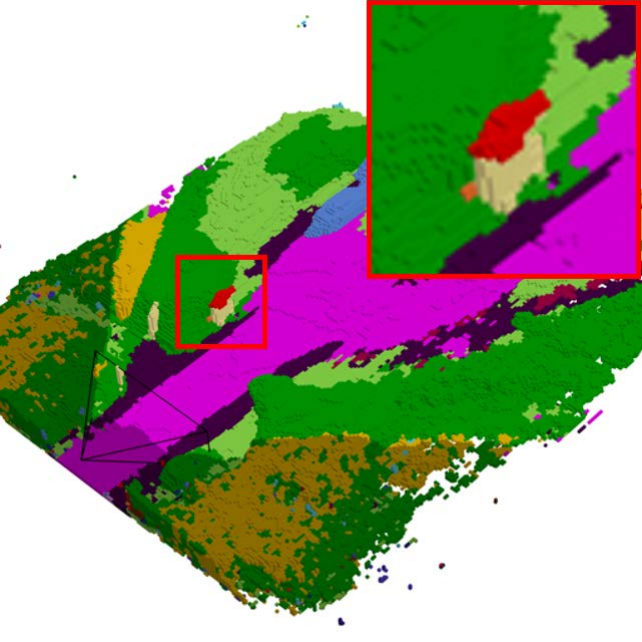} & 
		\includegraphics[width=.9\linewidth]{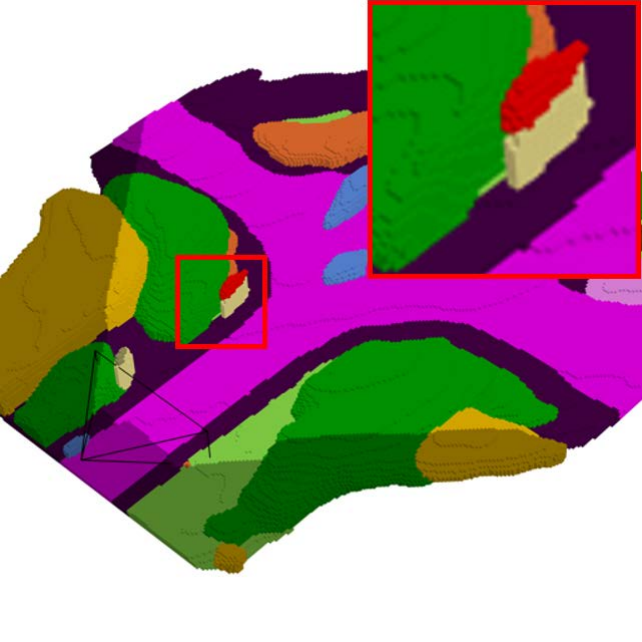} & 
		\includegraphics[width=.9\linewidth]{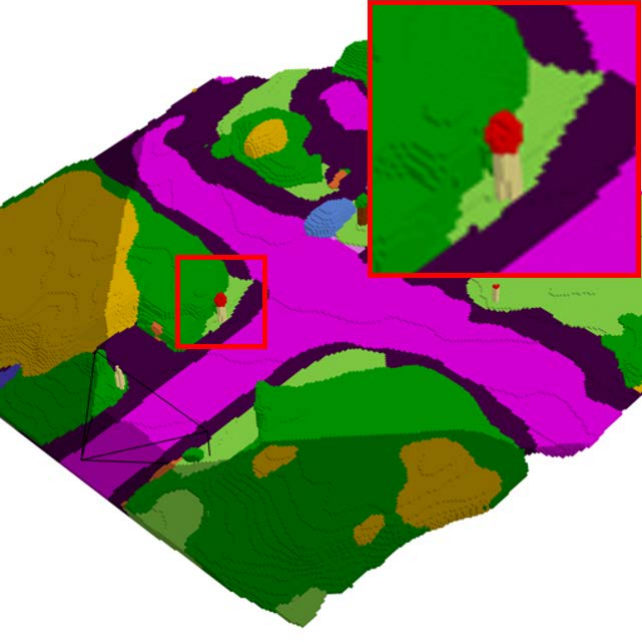} &
		\includegraphics[width=.9\linewidth]{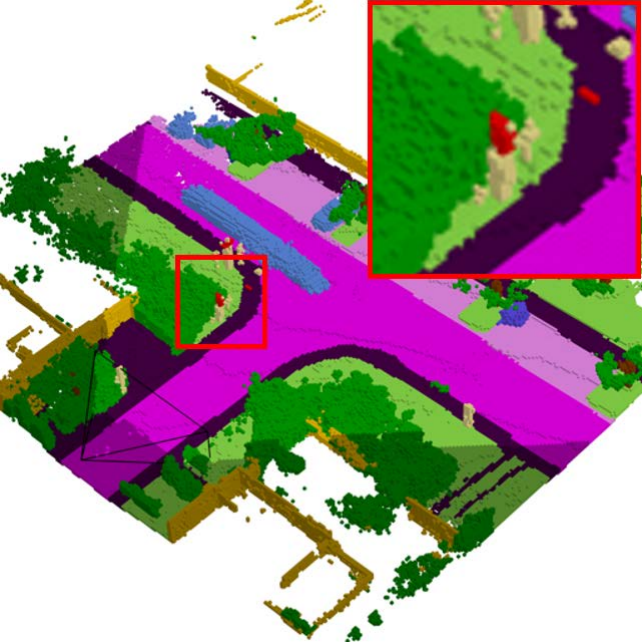}
		\\
		\includegraphics[width=.9\linewidth]{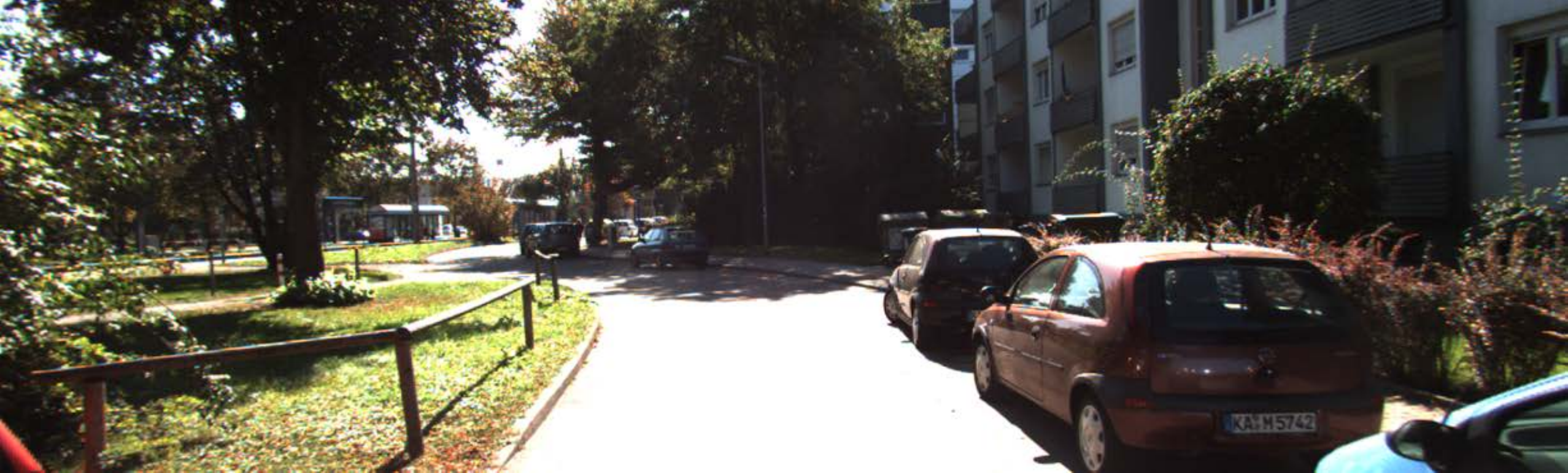} &
		\includegraphics[width=.9\linewidth]{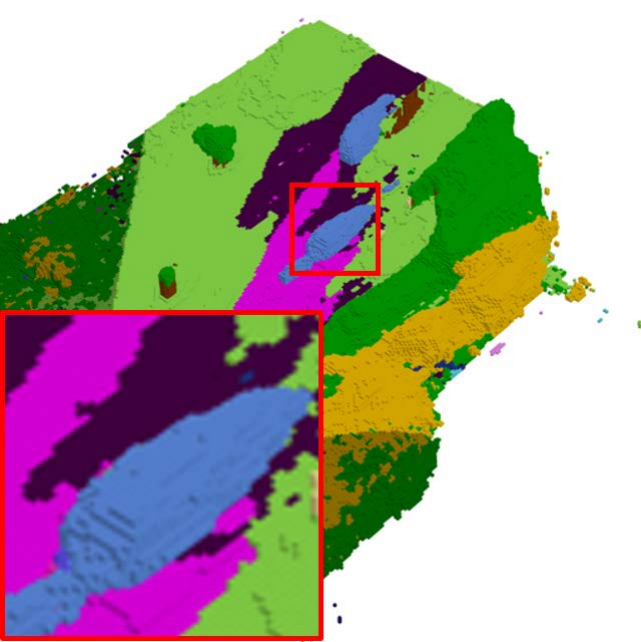} & 
		\includegraphics[width=.9\linewidth]{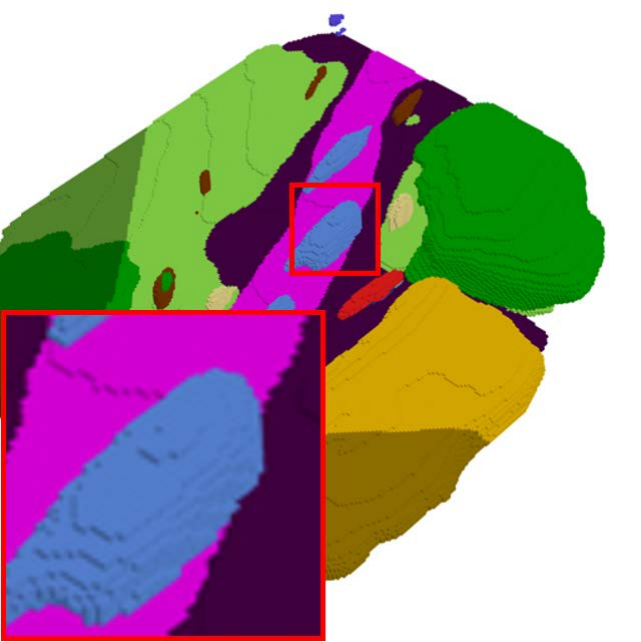} & 
		\includegraphics[width=.9\linewidth]{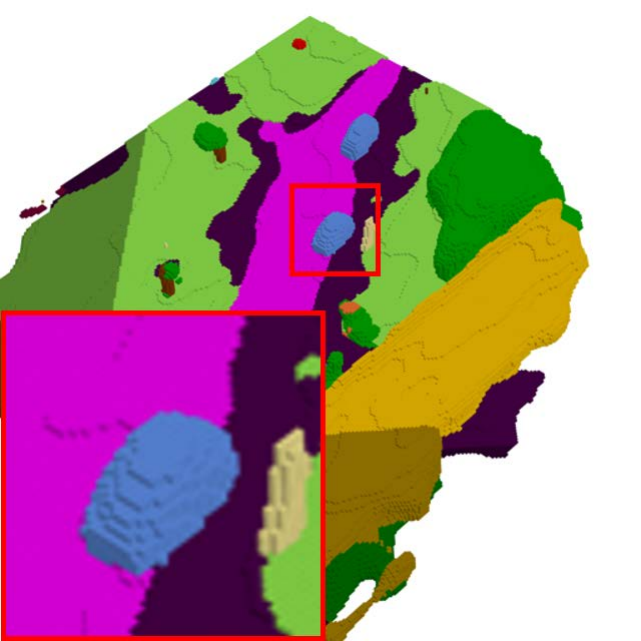} &
		\includegraphics[width=.9\linewidth]{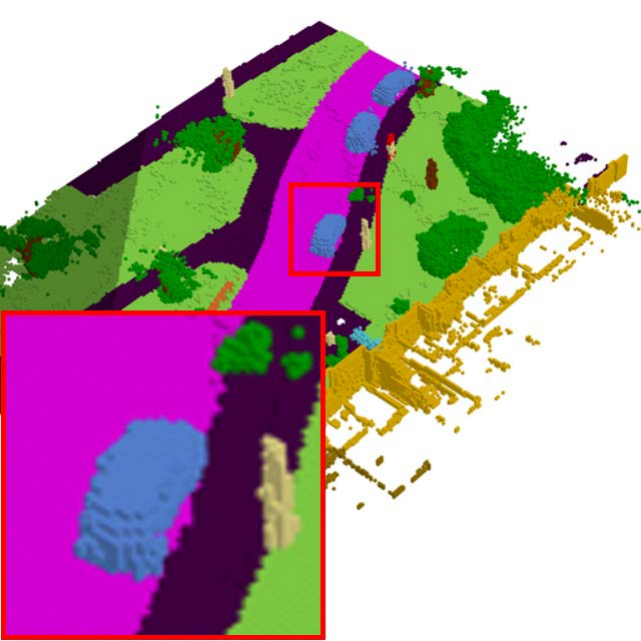}
		\\
		\includegraphics[width=.9\linewidth]{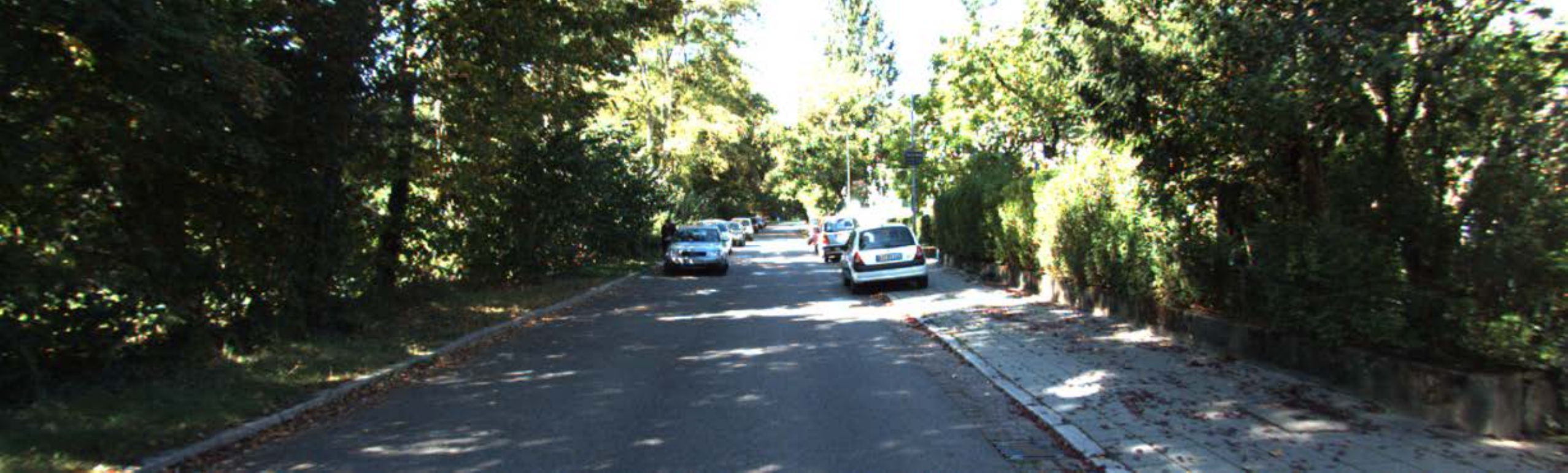} &
		\includegraphics[width=.9\linewidth]{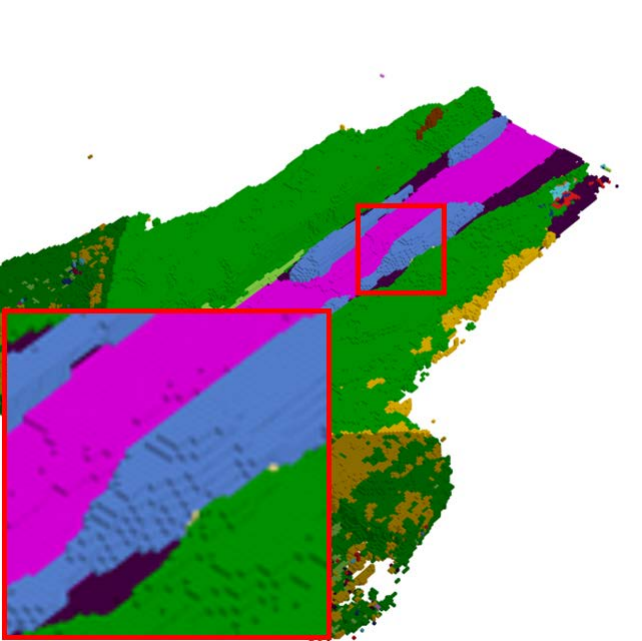} & 
		\includegraphics[width=.9\linewidth]{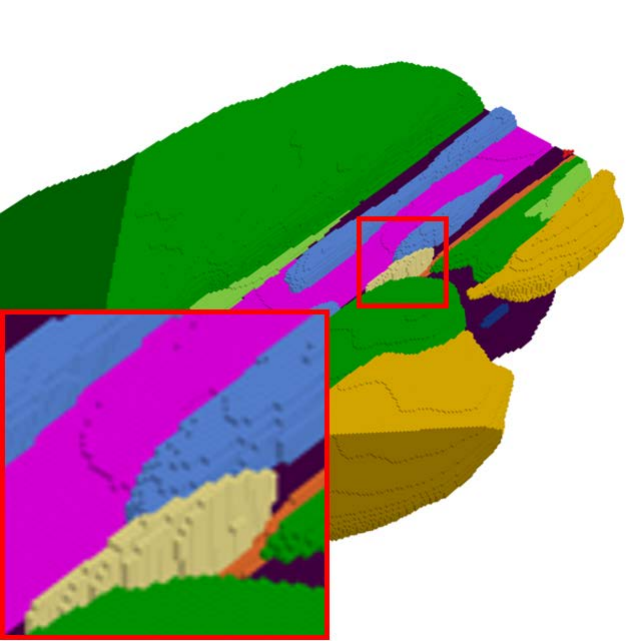} & 
		\includegraphics[width=.9\linewidth]{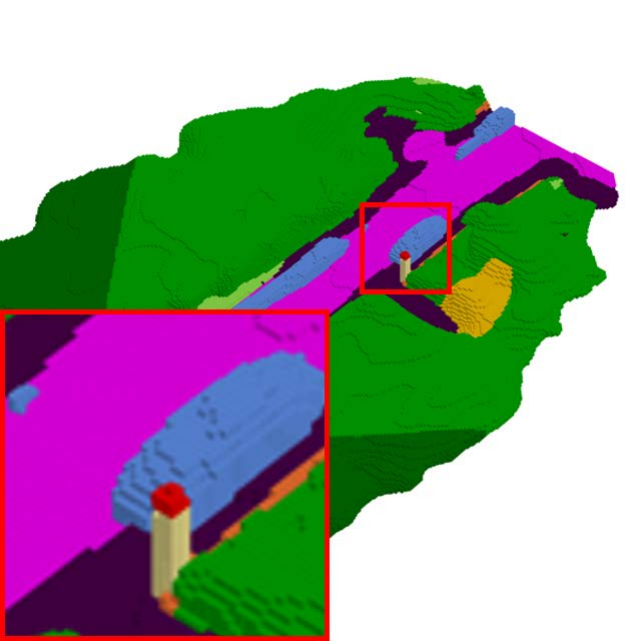} &
		\includegraphics[width=.9\linewidth]{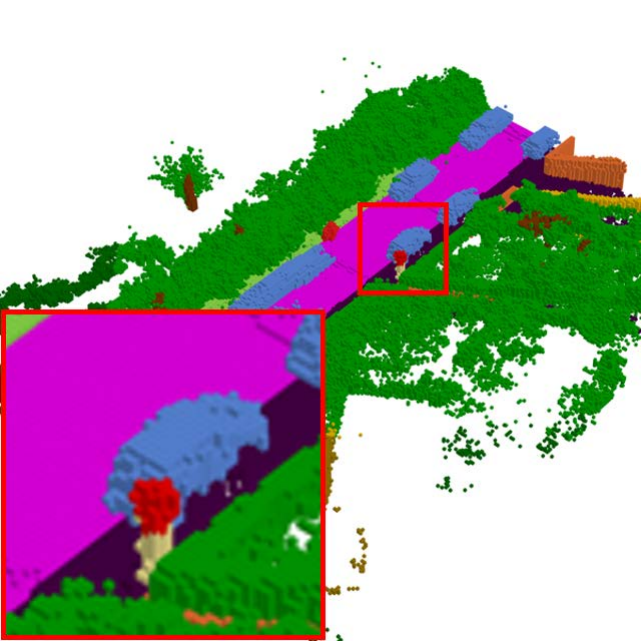}
		\\
		(a) RGB & (b) VoxFormer~\cite{VoxFormer} & (c) OccFormer~\cite{OccFormer} & (d) CGFormer (ours) & (e) Ground Truth
	\end{tabular}
	\caption{Qualitative visualization results on the SemanticKITTI~\cite{SemanticKITTI} validation set.}
	\label{fig:visualization}
	\vspace{-4mm}
\end{figure*}

%% file: src/conclusion.tex
\section{Conclusions}
\label{sec:conclusion}
In this paper, we present CGFormer, a novel neural network for Semantic Scene Completion. CGFormer dynamically generates distinct voxel queries, which serve as a good starting point for the attention layers, capturing the unique characteristics of various input images. To improve the accuracy of the estimated depth probability, we propose a simple yet efficient depth refinement module, with minimal computational burden. To boost the semantic and geometric representation abilities, CGFormer incorporates multiple representations (voxel and TPV) to encode the 3D volumes from both local and global perspectives. We experimentally show that our CGFormer achieves state-of-the-art performance on the SemanticKITTI and SSCBench-KITTI-360 benchmarks.

%% file: src/acknowledgments.tex
\newpage
\section*{Acknowledgments}
This work was supported in part by the National Key Research and Development Program of China under Grant No. 2023YFB3209800, in part by Zhejiang Provincial Natural Science Foundation of China under Grant No. LD24F020003, and in part by the National Natural Science Foundation of China under Grant No. 62301484.

%% file: src/appendix.tex
\newpage
\appendix

\section{Appendix / Supplemental Material}
In the appendix, we mainly provide implementation details and more experiment results.
\subsection{Datasets and Metrics}
\noindent\textbf{Datasets.} We evaluate our CGFormer on two datasets: SemanticKITTI~\cite{SemanticKITTI} and SSC-Bench-KITTI-360~\cite{SSCBench}. These datasets are derived from the KITTI Odometry~\cite{odometry} and KITTI-360~\cite{KITTI360} Benchmarks, respectively. The evaluation focuses on a specific spatial volume: $51.2m$ in front of the car, $25.6m$ to the left and right sides, and $6.4m$ above the car. Voxelization of this volume results in a set of 3D voxel grids with a resolution of $256 \times 256 \times 32$, where each voxel measures $0.2m \times 0.2m \times 0.2m$. SemanticKITTI provides RGB images with dimensions of $1226 \times 370$ as inputs, encompassing 20 unique semantic classes (19 semantic classes and 1 free class). The dataset includes 10 sequences for training, 1 sequence for validation, and 11 sequences for testing. SSC-Bench-KITTI-360~\cite{SSCBench} offers 7 sequences for training, 1 sequence for validation, and 1 sequence for testing. It contains 19 unique semantic classes (18 semantic classes and 1 free class), with input RGB images having a resolution of $1408 \times 376$.

\noindent\textbf{Metrics.} Following previous methods~\cite{MonoScene, VoxFormer, TPVFormer}, we report the intersection over union (IoU) and mean IoU (mIoU) metrics for occupied voxel grids and voxel-wise semantic predictions, respectively. The interplay between IoU and mIoU offers a comprehensive perspective on the model's effectiveness in capturing both geometry and semantic aspects of the scene.

\subsection{Implementation Details}
\noindent\textbf{Network Structures.} Consistent with previous researches~\cite{TPVFormer, MonoScene, surroundOcc}, we utilize a 2D UNet based on a pretrained EfficientNetB7~\cite{EfficientNet} as the image backbone. The CGVT generates a 3D feature volume with dimensions of $128 \times 128 \times 16$ and $128$ channels. The numbers of deformable attention layers for cross-attention and self-attention are $3$ and $2$ respectively. We use 8 sampling points around each reference point for the cross and self-attention head. The voxel-based branch of the LGE comprises $3$ stages with 2 residual blocks~\cite{ResNet} each. SwinT~\cite{swin} is employed as the 2D backbone in the TPV-based branch. Both are followed by feature pyramid networks (FPNs)~\cite{FPN} to aggregate multi-scale features for dynamic fusion. The final prediction has dimensions of $128 \times 128 \times 16$ and is upsampled to $256 \times 256 \times 32$ through trilinear interpolation to align the resolution with the ground truth.

\noindent\textbf{Training Setup.} We train CGFormer for 25 epochs on 4 NVIDIA 4090 GPUs, with a batch size of 4. It approximately consumes 19 GB of GPU memory on each GPU during the training phase. We employ the AdamW~\cite{AdamW} optimizer with $\beta_{1} = 0.9$, $\beta_{2} =0.99$ and set the maximum learning rate to $3\times 10^{-4}$. The cosine annealing learning rate strategy is adopted for the learning rate decay, where the cosine warmup strategy is applied for the first $5\%$ iterations.

\subsection{Results Using Monocular Inputs}
In alignment with previous methods~\cite{VoxFormer, Symphonize}, we evaluate the performance of our CGFormer using only a monocular RGB image as input. We replace the depth estimation network with AdaBins~\cite{Adabins} and present the results on the semantickitti validation set in the table~\ref{tab:monocular}. To better demonstrate the advantage of our CGFormer, we also include the results of VoxFormer, Symphonize, and OccFormer. Compared to the stereo-based methods when using only a monocular image (VoxFormer, Symphonize), CGFormer achieves superior performance in terms of both IoU and mIoU. Furthermore, our method also surpasses OccFormer, the state-of-the-art monocular method.

\begin{table}[!h]
	\centering
	\caption{The performance of the CGFormer with more lightweight backbone networks.}
	\resizebox{0.7\linewidth}{!}
	{
		\begin{tabular}{c|cc|cc}
			\toprule
			Backbone Networks & IoU & mIoU & Parameters & Training Memory\\
			\midrule
			EfficientNetB7, Swin Block &	\textbf{45.99} & \textbf{16.87} & 122.42 & 19330\\ 
			ResNet50, Swin Block & 45.99 & 16.79 & 80.46 & 19558\\
			ResNet50, ResBlock & 45.86 & 16.85 	& \textbf{54.8} & \textbf{18726}\\
			\bottomrule
		\end{tabular}
	}
	\setlength{\abovecaptionskip}{0cm}
	\setlength{\belowcaptionskip}{0cm}
	\label{tab:lightweight}
\end{table}

\begin{table}
	\centering
	\caption{Comparison of the performance using monocular inputs. For stereo-based methods, we replace the MobileStereoNet~\cite{MobileStereoNet} with Adabins~\cite{Adabins}.}
	\resizebox{0.3\linewidth}{!}
	{
		\begin{tabular}{c|cc}
			\toprule
			Method & IoU & mIoU \\
			\midrule
			VoxFormer-S~\cite{VoxFormer} & 38.68 & 10.67 \\
			VoxFormer-T~\cite{VoxFormer} & 38.08 & 11.27 \\
			Symphonize~\cite{Symphonize} & 38.37 & 12.20 \\
			OccFormer~\cite{OccFormer} & 36.50 & 13.46 \\
			CGFormer (ours) & \textbf{41.82} & \textbf{14.06} \\ 
			\bottomrule
		\end{tabular}
	}
	\setlength{\abovecaptionskip}{0cm}
	\setlength{\belowcaptionskip}{0cm}
	\label{tab:monocular}
	\vspace{-4mm}
\end{table}

\subsection{Reults with More Lightweight Backbone Networks}
We reanalyze the components of CGFormer, finding that replacing EfficientNetB7, used as the image backbone, and the Swin blocks, used in the TPV branch backbone, with more lightweight ResNet50 and residual blocks, respectively, can significantly reduce the number of parameters of our network. Besides, we also remove the predefined parameters as we find it doesn't influence the final performance. The results on the semantickitti validation set are presented in the Table~\ref{tab:lightweight}. Compared to the original architecture, CGFormer maintains stable performance regardless of the backbone networks used for the image encoder and TPV branch encoder, underscoring its effectiveness, robustness, and potential.

\input{tab/tab_memory_inference}

\subsection{Additional Quantitative Results}
For more comprehensive comparison, we list the results with input modality and image backbones in Table~\ref{tab:sem_kitti_test_complete} and Table~\ref{tab:kitti_360_test_complete}. Table~\ref{tab:sem_kitti_val} presents the comparison results of CGFormer with the state-of-the-art methods on the SemanticKITTI validation set. CGFormer outperforms all other methods in terms of both IoU and mIoU. Additionally, it ranks either first or second on most of the classes, demonstrating consistent performance across various semantic categories, as indicated in previous tables.

\subsection{Computational Cost}
In Table~\ref{tab:time}, we display the training memory and inference time of CGFormer, along with those of the comparison methods. Additionally, the table includes the corresponding IoU and mIoU metrics for comprehensive comparison. As shown in the table, CGFormer achieves the best performance in terms of both IoU and mIoU, with comparable training memory and inference time.

\subsection{Additional Qualitative Results}
We offer additional visualization results in Fig.\ref{fig:AdditionQualitative} and Fig.\ref{fig:AdditionQualitative2}. These examples are randomly selected from the SemanticKITTI~\cite{SemanticKITTI} validation set.

\input{tab/tab_sem_test_complete}
\input{tab/tab_kitti_360_test_complete}
\input{tab/tab_sem_val}

\subsection{Failure Cases}
\begin{figure*}
	\centering
	\newcolumntype{P}[1]{>{\centering\arraybackslash}m{#1}}
	\setlength{\tabcolsep}{0.001\textwidth}
	\footnotesize
	\begin{tabular}{P{0.3\textwidth} P{0.3\textwidth} P{0.3\textwidth}} 		
		\includegraphics[width=.7\linewidth]{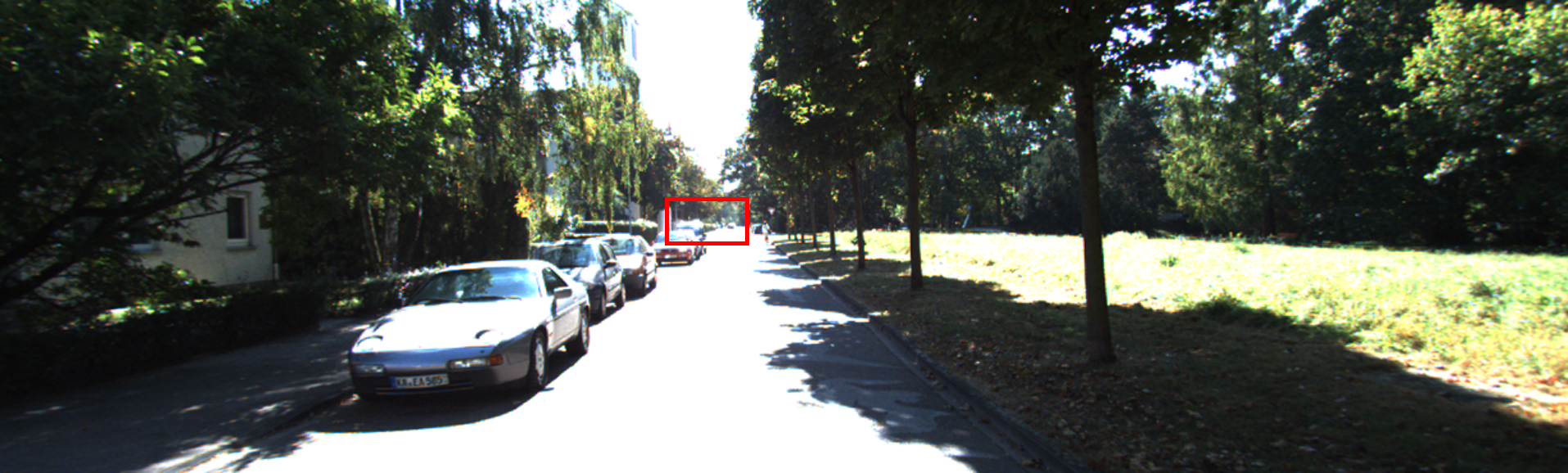} &  
		\includegraphics[width=.5\linewidth]{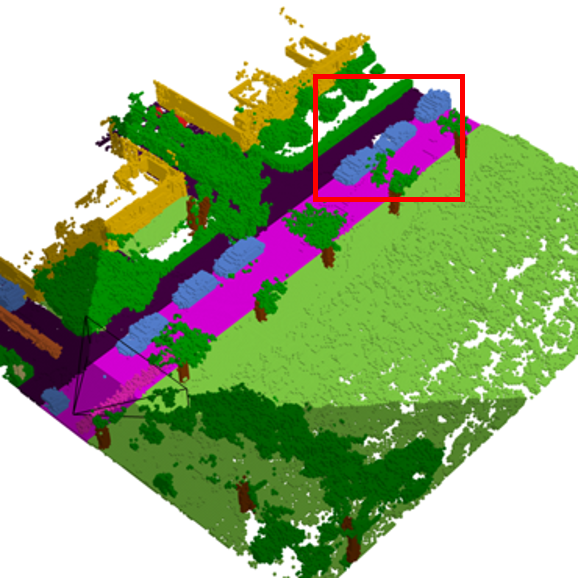} & 
		\includegraphics[width=.5\linewidth]{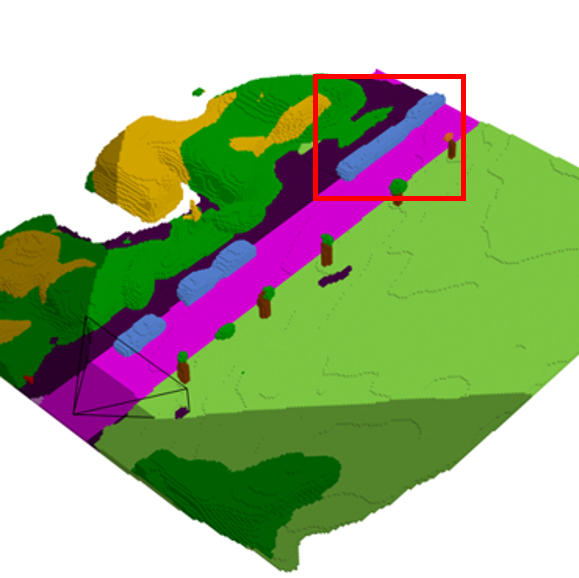} \\
		\includegraphics[width=.7\linewidth]{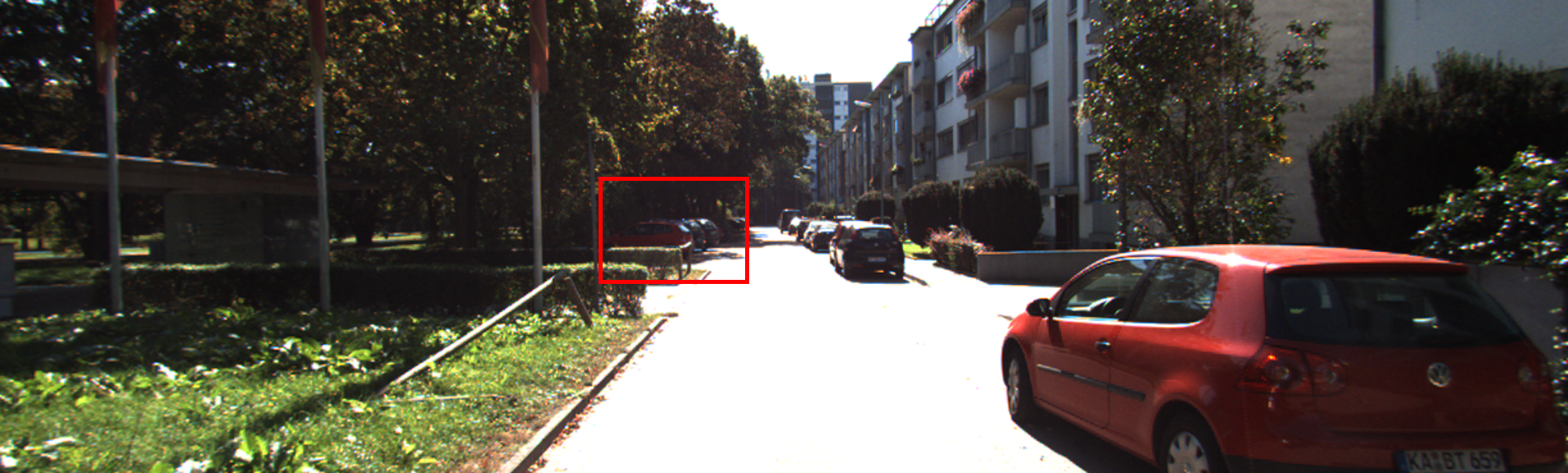} &  
		\includegraphics[width=.5\linewidth]{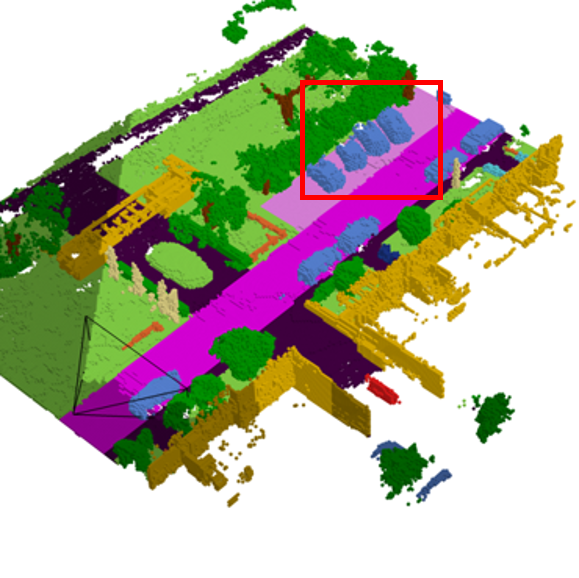} & 
		\includegraphics[width=.5\linewidth]{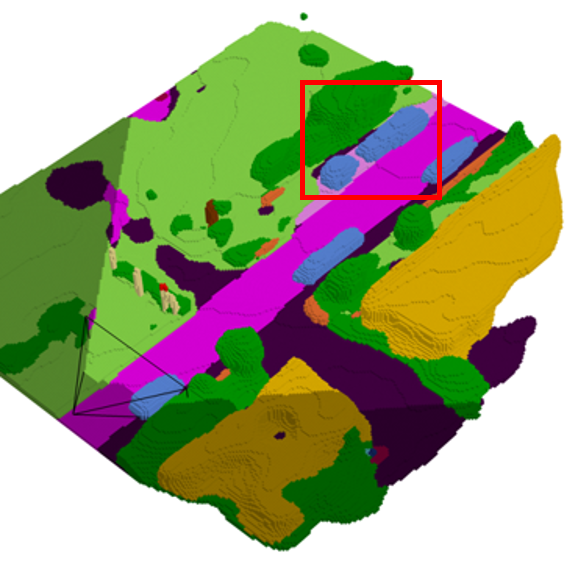}  
		\\
		(a) RGB & (b) Ground Truth & (c) CGFormer (ours)
	\end{tabular}
	\caption{Failure cases.}
	\label{fig:failure_cases}
	
\end{figure*}

We provide two failure cases in Fig.~\ref{fig:failure_cases}.

\subsection{Limitations}
While CGFormer exhibits strong performance on benchmarks, but the accuracy on most of the categories (\eg person, bicyclist, other vehicle) is unsatisfactory. Improving the performance on these instances could be beneficial for the downstream application tasks. Furthermore, there is a need to explore designing depth estimation networks under multi-view scenarios to extend the geometry-aware view transformation to these scenes. Despite these limitations, we are confident that CGFormer will contribute to advancing the field of 3D perception.

\input{fig/MoreQualitativeResults}
\input{fig/MoreQualitativeResults2}

%% file: tab/tab_memory_inference.tex
\begin{table}
	\centering
	\caption{Comparison of training memory and inference time with SOTA methods on the and SemanticKITTI test set. These metrics were measured on the NVIDIA 4090 GPU.}
	\resizebox{\linewidth}{!}
	{
		\begin{tabular}{c|cccccc}
			\toprule
			Method & TPVFormer~\cite{TPVFormer} & OccFormer~\cite{OccFormer} & VoxFormer~\cite{VoxFormer} & Symphonize~\cite{Symphonize} & StereoScene~\cite{StereoScene}  & CGFormer (ours)\\ 
			\midrule
			Training Meomry (M) & 18564 & 18080 & 18725 & 17757 & 19000 & 19330 \\
			Inference Time (ms) & 207 & 199 & 204 & 216 & 258 & 205 \\
			IoU & 34.25 & 34.53 & 42.95 & 42.19 & 43.34 & \textbf{44.41} \\
			mIoU & 11.26 & 12.20 & 12.20 & 15.04 & 15.36 & \textbf{16.63} \\
			\bottomrule
		\end{tabular}
	}
	\setlength{\abovecaptionskip}{0cm}
	\setlength{\belowcaptionskip}{0cm}
	\label{tab:time}
	\vspace{-4mm}
\end{table}

%% file: tab/tab_sem_test_complete.tex
\begin{table}
	\newcommand{\clsname}[2]{
		\rotatebox{90}{
			\hspace{-6pt}
			\textcolor{#2}{$\blacksquare$}
			\hspace{-6pt}
			\renewcommand\arraystretch{0.6}
			\begin{tabular}{l}
				#1                                      \\
				\hspace{-4pt} ~\tiny(\semkitfreq{#2}\%) \\
			\end{tabular}
	}}
	\centering
	\caption{Quatitative results on SemanticKITTI~\cite{SemanticKITTI} test set. $^\ast$ represents the reproduced results in ~\cite{TPVFormer, OccFormer}. The best and the second best results are in \textbf{bold} and \underline{underlined}, respectively. Our CGFormer outperforms temporal stereo-based (Stereo-T) methods or those methods with larger image backbones in terms of IoU and mIoU.}
	\resizebox{\linewidth}{!}
	{
		\begin{tabular}{c|c|c|cc|ccccccccccccccccccc}
			\toprule		
			Method 								 & 
			Input 								 & 
			Image Backbone 						 & 
			IoU 								 & 
			mIoU  								 &  
			\clsname{road}{road}                 & 
			\clsname{sidewalk}{sidewalk}         &        
			\clsname{parking}{parking}           & 
			\clsname{other-grnd.}{otherground}   & 
			\clsname{building}{building}         & 
			\clsname{car}{car} 					 & 
			\clsname{truck}{truck}               &
			\clsname{bicycle}{bicycle}           &
			\clsname{motorcycle}{motorcycle}     &
			\clsname{other-veh.}{othervehicle}   &
			\clsname{vegetation}{vegetation}     &
			\clsname{trunk}{trunk}               &
			\clsname{terrain}{terrain}           &
			\clsname{person}{person}             &
			\clsname{bicyclist}{bicyclist}       &
			\clsname{motorcyclist}{motorcyclist} &
			\clsname{fence}{fence}               &
			\clsname{pole}{pole}                 &
			\clsname{traf.-sign}{trafficsign}
			\\
			\midrule
			MonoScene$^\ast$~\cite{MonoScene} & Mono & EfficientNetB7 & 34.16 & 11.08 & 54.70 & 27.10 & 24.80 & 5.70
			& 14.40 & 18.80 & 3.30 & 0.50 & 0.70 & 4.40  & 14.90 & 2.40  & 19.50 & 1.00  & 1.40
			& 0.40  & 11.10 & 3.30 & 2.10         \\
			TPVFormer~\cite{TPVFormer}        & Mono & EfficientNetB7 &34.25 & 11.26 & 55.10 & 27.20 & 27.40 & 6.50
			& 14.80 & 19.20 & 3.70 & 1.00 & 0.50 & 2.30  & 13.90 & 2.60  & 20.40 & 1.10  & 2.40
			& 0.30  & 11.00 & 2.90 & 1.50 		  \\
			SurroundOcc~\cite{surroundOcc}    & Mono & EfficientNetB7 & 34.72 & 11.86 & 56.90 & 28.30 & 30.20 & 6.80 
			& 15.20 & 20.60 & 1.40 & 1.60 & 1.20 & 4.40  & 14.90 & 3.40  & 19.30 & 1.40  & 2.00
			& 0.10  & 11.30 & 3.90 & 2.40         \\
			OccFormer~\cite{OccFormer}        & Mono & EfficientNetB7 & 34.53 & 12.32 & 55.90 & 30.30 & \underline{31.50} & 6.50          
			& 15.70 & 21.60 & 1.20 & 1.50 & 1.70 & 3.20  & 16.80 & 3.90  & 21.30 & 2.20  & 1.10
			& 0.20  & 11.90 & 3.80 & 3.70         \\
			IAMSSC~\cite{IAMSSC}  & Mono & ResNet50 & 43.74 & 12.37 & 54.00 & 25.50 & 24.70 & 6.90 & 19.20 & 21.30 & 3.80 & 1.10 & 0.60 & 3.90 & 22.70 & 5.80 & 19.40 & 1.50 & 2.90 & 0.50 & 11.90 & 5.30 & 4.10 \\
			VoxFormer-S~\cite{VoxFormer}        & Stereo & ResNet50 & 42.95 & 12.20 & 53.90 & 25.30 & 21.10 & 5.60
			& 19.80 & 20.80 & 3.50 & 1.00 & 0.70 & 3.70  & 22.40 & 7.50  & 21.30 & 1.40  & 2.60 
			& 0.20  & 11.10 & 5.10 & 4.90         \\
			VoxFormer-T~\cite{VoxFormer}        & Stereo-T & ResNet50 & 43.21 & 13.41 & 54.10 & 26.90 & 25.10 & 7.30 & 23.50 & 21.70 & 3.60 & 1.90 & 1.60 & 4.10 & 24.40 & 8.10 & 24.20 & 1.60 & 1.10 & 0.00 & 13.10 & 6.60 & 5.70 \\
			DepthSSC~\cite{DepthSSC}           & Stereo & ResNet50 & \textbf{44.58} & 13.11 & 55.64 & 27.25 & 25.72 & 5.78
			& 20.46 & 21.94 & 3.74 & 1.35 & 0.98 & 4.17  & 23.37 & 7.64  & 21.56 & 1.34  & 2.79
			& 0.28  & 12.94 & 5.87 & 6.23         \\
			Symphonize~\cite{Symphonize}	   & Stereo & MaskDINO & 42.19 & 15.04 & 58.40 & 29.30 & 26.90 & \underline{11.70}
			& \underline{24.70} & 23.60 & 3.20 & 3.60 & \textbf{2.60} & 5.60  & 24.20 & 10.00 & 23.10 & \textbf{3.20}  & 1.90  
			& \textbf{2.00}  & 16.10 & \underline{7.70} & 8.00         \\
			HASSC-S~\cite{HASSC}  & Stereo & ResNet50 & 43.40 & 13.34 & 54.60 & 27.70 & 23.80 & 6.20 & 21.10 & 22.80 & 4.70 & 1.60 & 1.00 & 3.90 & 23.80 & 8.50 & 23.30 & 1.60 & \underline{4.00} & 0.30 & 13.10 & 5.80 & 5.50 \\
			HASSC-T~\cite{HASSC}  & Stereo-T & ResNet50 & 42.87 & 14.38 & 55.30 & 29.60 & 25.90 & 11.30 & 23.10 & 23.00 & 2.90 & 1.90 & 1.50 & 4.90 & 24.80 & 9.80 & 26.50 & 1.40 & 3.00 & 0.00 & 14.30 & 7.00 & 7.10 \\
			StereoScene~\cite{StereoScene}  & Stereo & EfficientNetB7 & 43.34 & 15.36 & \underline{61.90} & \underline{31.20} & 30.70 & 10.70 & 24.20 & 22.80 & 2.80 &  3.40 & \underline{2.40} & \underline{6.10} & 23.80 & 8.40 & \underline{27.00} & \underline{2.90} & 2.20 & 0.50 & 16.50 & 7.00 & 7.20 \\
			H2GFormer-S~\cite{H2GFormer}  & Stereo & ResNet50 & 44.20 & 13.72 & 56.40 & 28.60 & 26.50 & 4.90 & 22.80 & 23.40 & 4.80 & 0.80 & 0.90 & 4.10 & 24.60 & 9.10 & 23.80 & 1.20 & 2.50 & 0.10 & 13.30 & 6.40 & 6.30 \\
			H2GFormer-T~\cite{H2GFormer}  & Stereo-T & ResNet50 & 43.52 & 14.60 & 57.90 & 30.40 & 30.00 & 6.90 & 24.00 & 23.70 & \underline{5.20} & 0.60 & 1.20 & 5.00 & \textbf{25.20} & \underline{10.70} & 25.80 & 1.10 & 0.10 & 0.00 & 14.60 & 7.50 & \textbf{9.30} \\
			MonoOcc-S~\cite{MonoOcc}  & Stereo & ResNet50 & - & 13.80 & 55.20 & 27.80 & 25.10 & 9.70 & 21.40 & 23.20 & \underline{5.20} & 2.20 & 1.50 & 5.40 & 24.00 & 8.70 & 23.00 & 1.70 & 2.00 & 0.20 & 13.40 & 5.80 & 6.40 \\
			MonoOcc-L~\cite{MonoOcc}  & Stereo & InternImage-XL & - & \underline{15.63} & 59.10 & 30.90 & 27.10 & 9.80 & 22.90 & \underline{23.90} & \textbf{7.20} & \textbf{4.50} & \underline{2.40} & \textbf{7.70} & \underline{25.00} & 9.80 & 26.10 & 2.80 & \textbf{4.70} & \underline{0.60} & \underline{16.90} & 7.30 & \underline{8.40} \\
			\hline
			CGFormer (ours)  & Stereo & EfficientNetB7 & \underline{44.41} & \textbf{16.63} & \textbf{64.30} & \textbf{34.20} & \textbf{34.10} & \textbf{12.10} & \textbf{25.80} & \textbf{26.10} & 4.30 & \underline{3.70} & 1.30 & 2.70 & 24.50 & \textbf{11.20} & \textbf{29.30} & 1.70 & 3.60 & 0.40 & \textbf{18.70} & \textbf{8.70} & \textbf{9.30} \\
			\bottomrule
		\end{tabular}
	}
	\setlength{\abovecaptionskip}{0cm}
	\setlength{\belowcaptionskip}{0cm}
	\label{tab:sem_kitti_test_complete}
	\vspace{-4mm}
\end{table}

%% file: tab/tab_kitti_360_test_complete.tex
\begin{table}
	\newcommand{\clsname}[2]{
		\rotatebox{90}{
			\hspace{-6pt}
			\textcolor{#2}{$\blacksquare$}
			\hspace{-6pt}
			\renewcommand\arraystretch{0.6}
			\begin{tabular}{l}
				#1                                       \\
				\hspace{-4pt} ~\tiny(\sscbkitfreq{#2}\%) \\
			\end{tabular}
	}}
	\centering
	\caption{Quantitative results on SSCBench-KITTI360 test set. The results for counterparts are provided in \cite{SSCBench}. The best and the second best results for all camera-based methods are in \textbf{bold} and \underline{underlined}, respectively. The best results from the LiDAR-based methods are in \textcolor{red}{red}.}
	\resizebox{\linewidth}{!}
	{
		\begin{tabular}{c|c|c|cc|cccccccccccccccccc}
			\toprule
			Method                            			&
			Input                                       &
			Image Backbone                              &
			IoU                                 		&
			mIoU                                		&
			\clsname{car}{car}                      	&
			\clsname{bicycle}{bicycle}              	&
			\clsname{motorcycle}{motorcycle}        	&
			\clsname{truck}{truck}                   	&
			\clsname{other-veh.}{othervehicle}      	&
			\clsname{person}{person}                	&
			\clsname{road}{road}                    	&
			\clsname{parking}{parking}              	&
			\clsname{sidewalk}{sidewalk}            	&
			\clsname{other-grnd.}{otherground}      	&
			\clsname{building}{building}            	&
			\clsname{fence}{fence}                  	&
			\clsname{vegetation}{vegetation}        	&
			\clsname{terrain}{terrain}              	&
			\clsname{pole}{pole}                    	&
			\clsname{traf.-sign}{trafficsign}       	&
			\clsname{other-struct.}{otherstructure} 	&
			\clsname{other-obj.}{otherobject}
			\\
			\midrule
			\multicolumn{21}{l}{\textit{LiDAR-based methods}}                                                                                                                                                                                                                                                                                                                                               \\
			\hline
			SSCNet~\cite{SSCNet}        & LiDAR & - & \textcolor{red}{53.58} & 16.95 & \textcolor{red}{31.95} & 0.00 & 0.17 & 10.29 & 0.00 & 0.07 & \textcolor{red}{65.70}
			& \textcolor{red}{17.33} & \textcolor{red}{41.24} & 3.22 & \textcolor{red}{44.41} & 6.77 & \textcolor{red}{43.72} & \textcolor{red}{28.87} & 0.78  & 0.75 & 8.69 & 0.67 \\
			LMSCNet~\cite{LMSCNet}      & LiDAR & - & 47.35 & 13.65 & 20.91 & 0.00 & 0.00 & 0.26  & 0.58 & 0.00 & 62.95        
			& 13.51 & 33.51 & 0.20 & 43.67 & 0.33 & 40.01 & 26.80 & 0.00  & 0.00 & 3.63 & 0.00 \\
			\specialrule{0.7pt}{0pt}{0pt}
			\multicolumn{21}{l}{\textit{Camera-based methods}}                                                                                                                                                                                                                                                                                                                                              \\
			\hline
			MonoScene~\cite{MonoScene}  & Mono & EfficientNetB7 & 37.87 & 12.31 & 19.34 & 0.43 & 0.58 & 8.02  & 2.03 & 0.86 & 48.35        
			& 11.38 & 28.13 & 3.32 & 32.89 & 3.53 & 26.15 & 16.75 & 6.92 & 5.67 & 4.20 & 3.09  \\
			TPVFormer~\cite{TPVFormer}  & Mono & EfficientNetB7 & 40.22 & 13.64 & 21.56 & 1.09 & 1.37 & 8.06  & 2.57 & 2.38 & 52.99        
			& 11.99 & 31.07 & 3.78 & 34.83 & 4.80 & 30.08 & 17.52 & 7.46 & 5.86 & 5.48 & 2.70  \\
			OccFormer~\cite{OccFormer}  & Mono & EfficientNetB7 & 40.27 & 13.81 & 22.58 & 0.66 & 0.26 & 9.89  & 3.82 & 2.77 & 54.30       
			& 13.44 & 31.53 & 3.55 & 36.42 & 4.80 & 31.00 & 19.51 & 7.77 & 8.51 & 6.95 & 4.60  \\
			VoxFormer~\cite{VoxFormer}  & Stereo & ResNet50 & 38.76 & 11.91 & 17.84 & 1.16 & 0.89 & 4.56  & 2.06 & 1.63 & 47.01
			& 9.67  & 27.21 & 2.89 & 31.18 & 4.97 & 28.99 & 14.69 & 6.51 & 6.92 & 3.79 & 2.43  \\
			IAMSSC~\cite{IAMSSC}  & Mono & ResNet50 & 41.80 & 12.97 & 18.53 & \underline{2.45} & 1.76 & 5.12 & 3.92 & 3.09 & 47.55 & 10.56 & 28.35 & 4.12 & 31.53 & 6.28 & 29.17 & 15.24 & 8.29 & 7.01 & 6.35 & 4.19 \\
			DepthSSC~\cite{DepthSSC}    & Stereo & ResNet50 & 40.85 & 14.28 & 21.90 & 2.36 & \underline{4.30} & 11.51 & 4.56 & 2.92 & 50.88
			& 12.89 & 30.27 & 2.49 & \underline{37.33} & 5.22 & 29.61 & \underline{21.59} & 5.97 & 7.71 & 5.24 & 3.51  \\
			Symphonies~\cite{Symphonize}  & Stereo & MaskDINO & \underline{44.12} & \underline{18.58} & \textbf{30.02} & 1.85 & \textbf{5.90} & \textbf{25.07} & \textbf{12.06} & \textbf{8.20} & \underline{54.94}
			& \underline{13.83} & \underline{32.76} & \textbf{6.93} & 35.11 & \textbf{8.58} & \underline{38.33} & 11.52 & \underline{14.01} & \underline{9.57} & \textbf{14.44} & \textbf{11.28} \\
			\hline
			CGFormer (ours) & Stereo & EfficientNetB7 & \textbf{48.07} & \textbf{20.05} & \underline{29.85} & \textbf{3.42} & 3.96 & \underline{17.59} & \underline{6.79} & \underline{6.63} & \textbf{63.85} & \textbf{17.15} & \textbf{40.72} & \underline{5.53} & \textbf{42.73} & \underline{8.22} & \textbf{38.80} & \textbf{24.94} & \textbf{16.24} & \textbf{17.45} & \underline{10.18} & \underline{6.77} \\
			\bottomrule
		\end{tabular}
	}
	\setlength{\abovecaptionskip}{0cm}
	\setlength{\belowcaptionskip}{0cm}
	\label{tab:kitti_360_test_complete}
	\vspace{-4mm}
\end{table}

%% file: tab/tab_sem_val.tex
\begin{table}
	\newcommand{\clsname}[2]{
		\rotatebox{90}{
			\hspace{-6pt}
			\textcolor{#2}{$\blacksquare$}
			\hspace{-6pt}
			\renewcommand\arraystretch{0.6}
			\begin{tabular}{l}
				#1                                      \\
				\hspace{-4pt} ~\tiny(\semkitfreq{#2}\%) \\
			\end{tabular}
	}}
	\centering
	\caption{Quatitative results on SemanticKITTI~\cite{SemanticKITTI} validation set.  $^\ast$ represents the reproduced results in ~\cite{TPVFormer, OccFormer, DepthSSC}. The best and the second best results are in \textbf{bold} and \underline{underlined}, respectively.}
	\resizebox{\linewidth}{!}
	{
		\begin{tabular}{c|c|c|cc|ccccccccccccccccccc}
			\toprule		
			Method 								 & 
			Input 								 & 
			Image Backbone 						 & 
			IoU 								 & 
			mIoU  								 &  
			\clsname{road}{road}                 & 
			\clsname{sidewalk}{sidewalk}         & 
			\clsname{parking}{parking}           & 
			\clsname{other-grnd.}{otherground}   & 
			\clsname{building}{building}         & 
			\clsname{car}{car} 					 & 
			\clsname{truck}{truck}               &
			\clsname{bicycle}{bicycle}           &
			\clsname{motorcycle}{motorcycle}     &
			\clsname{other-veh.}{othervehicle}   &
			\clsname{vegetation}{vegetation}     &
			\clsname{trunk}{trunk}               &
			\clsname{terrain}{terrain}           &
			\clsname{person}{person}             &
			\clsname{bicyclist}{bicyclist}       &
			\clsname{motorcyclist}{motorcyclist} &
			\clsname{fence}{fence}               &
			\clsname{pole}{pole}                 &
			\clsname{traf.-sign}{trafficsign}
			\\
			\midrule
			MonoScene$^\ast$~\cite{MonoScene} & Mono     & EfficientNetB7 & 36.86 & 11.08 & 56.52 & 26.72 & 14.27 & 0.46  & 14.09 & 23.26
			& 6.98  & 0.61  & 0.45  & 1.48  & 17.89 & 2.81  & 29.64 & 1.86  & 1.20  & 0.00  & 5.84  & 4.14  & 2.25 \\
			TPVFormer~\cite{TPVFormer}        & Mono     & EfficientNetB7 & 35.61 & 11.36 & 56.50 & 25.87 & 20.60 & 0.85  & 13.88 & 23.81 
			& 8.08  & 0.36  & 0.05  & 4.35  & 16.92 & 2.26  & 30.38 & 0.51  & 0.89  & 0.00  & 5.94  & 3.14  & 1.52 \\
			OccFormer~\cite{OccFormer}        & Mono     & EfficientNetB7 & 36.50 & 13.46 & \textbf{58.85} & 26.88 & 19.61 & 0.31  & 14.40 & 25.09          
			& \textbf{25.53} & 0.81  & 1.19  & 8.52  & 19.63 & 3.93  & 32.62 & 2.78  & 2.82  & 0.00  & 5.61  & 4.26  & 2.86 \\
			IAMSSC~\cite{IAMSSC}              & Mono     & ResNet50 & 44.29 & 12.45 & 54.55 & 25.85 & 16.02 & 0.70  & 17.38 & 26.26
			& 8.74  & 0.60  & 0.15  & 5.06  & 24.63 & 4.95  & 30.13 & 1.32  & 3.46  & 0.01  & 6.86  & 6.35  & 3.56 \\
			VoxFormer-S~\cite{VoxFormer}      & Stereo   & ResNet50  & 44.02 & 12.35 & 54.76 & 26.35 & 15.50 & 0.70  & 17.65 & 25.79
			& 5.63  & 0.59  & 0.51  & 3.77  & 24.39 & 5.08  & 29.96 & 1.78  & 3.32  & 0.00  & 7.64  & 7.11  & 4.18 \\
			VoxFormer-T~\cite{VoxFormer}      & Stereo-T & ResNet50 & 44.15 & 13.35 & 53.57 & 26.52 & 19.69 & 0.42 & 19.54 & 26.54 & 7.26 & 1.28 & 0.56 & 7.81 & 26.10 & 6.10 & 33.06 & 1.93 & 1.97 & 0.00 & 7.31 & 9.15 & 4.94 \\
			DepthSSC~\cite{DepthSSC}          & Stereo   & ResNet50 & \underline{45.84} & 13.28 & 55.38 & 27.04 & 18.76 & 0.92  & 19.23 & 25.94
			& 6.02  & 0.35  & 1.16  & 7.50  & 26.37 & 4.52  & 30.19 & 2.58  & \textbf{6.32}  & 0.00  & 8.46  & 7.42  & 4.09 \\
			Symphonize~\cite{Symphonize} 	  & Stereo   & MaskDINO & 41.92 & \underline{14.89} & 56.37 & 27.58 & 15.28 & 0.95  & 21.64 & \underline{28.68}
			& \underline{20.44} & \underline{2.54}  & \textbf{2.82}  & \textbf{13.89} & 25.72 & 6.60  & 30.87 & \textbf{3.52}  & 2.24  & 0.00  & 8.40  & 9.57  & 5.76 \\
			HASSC-S~\cite{HASSC}              & Stereo   & ResNet50 & 44.82 & 13.48 & 57.05 & 28.25 & 15.90 & \underline{1.04} & 19.05 & 27.23 & 9.91 & 0.92 & 0.86 & 5.61 & 25.48 & 6.15 & 32.94 & \underline{2.80} & \underline{4.71} & 0.00 & 6.58 & 7.68 & 4.05 \\
			HASSC-T~\cite{HASSC}              & Stereo-T & ResNet50 & 44.58 & 14.74 & 55.30 & 29.60 & \textbf{25.90} & \textbf{11.30} & \underline{23.10} & 23.00 & 2.90 & 1.90 & 1.50 & 4.90 & 24.80 & \textbf{9.80} & 26.50 & 1.40 & 3.00 & 0.00 & \textbf{14.30} & 7.00 & \underline{7.10} \\ 
			H2GFormer-S~\cite{H2GFormer} & Stereo & ResNet50 & 44.57 & 13.73 & 56.08 & 29.12 & 17.83 & 0.45 & 19.74 & 28.21 & 10.00 & 0.50 & 0.47 & 7.39 & 26.25 & 6.80 & 34.42 & 1.54 & 2.88 & 0.00 & 7.24 & 7.88 & 4.68 \\
			H2GFormer-T~\cite{H2GFormer} & Stereo-T & ResNet50 & 44.69 & 14.29 & 57.00 & \underline{29.37} & \underline{21.74} & 0.34 & 20.51 & 28.21 & 6.80 & 0.95 & 0.91 & \underline{9.32} & \textbf{27.44} & 7.80 & \underline{36.26} & 1.15 & 0.10 & 0.00 & 7.98 & \underline{9.88} & 5.81 \\
			\hline
			CGFormer (ours) 					  & Stereo    & EfficientNetB7   & \textbf{45.99} & \textbf{16.87} & \textbf{65.51} & \textbf{32.31} & 20.82 & 0.16  & \textbf{23.52} & \textbf{34.32} 
			& 19.44 & \textbf{4.61} & \underline{2.71}  & 7.67  & \underline{26.93} & \underline{8.83}  & \textbf{39.54} & 2.38  & 4.08  & 0.00  & \underline{9.20} & \textbf{10.67} & \textbf{7.84} \\
			\bottomrule
		\end{tabular}
	}
	\setlength{\abovecaptionskip}{0cm}
	\setlength{\belowcaptionskip}{0cm}
	\label{tab:sem_kitti_val}
\end{table}

%% file: fig/MoreQualitativeResults.tex
\begin{figure*}
	\centering
	\newcolumntype{P}[1]{>{\centering\arraybackslash}m{#1}}
	\renewcommand{\arraystretch}{1.0}
	\footnotesize
	\begin{tabular}{P{0.17\textwidth} P{0.18\textwidth} P{0.18\textwidth} P{0.19\textwidth} P{0.17\textwidth}}	
		\includegraphics[width=.9\linewidth]{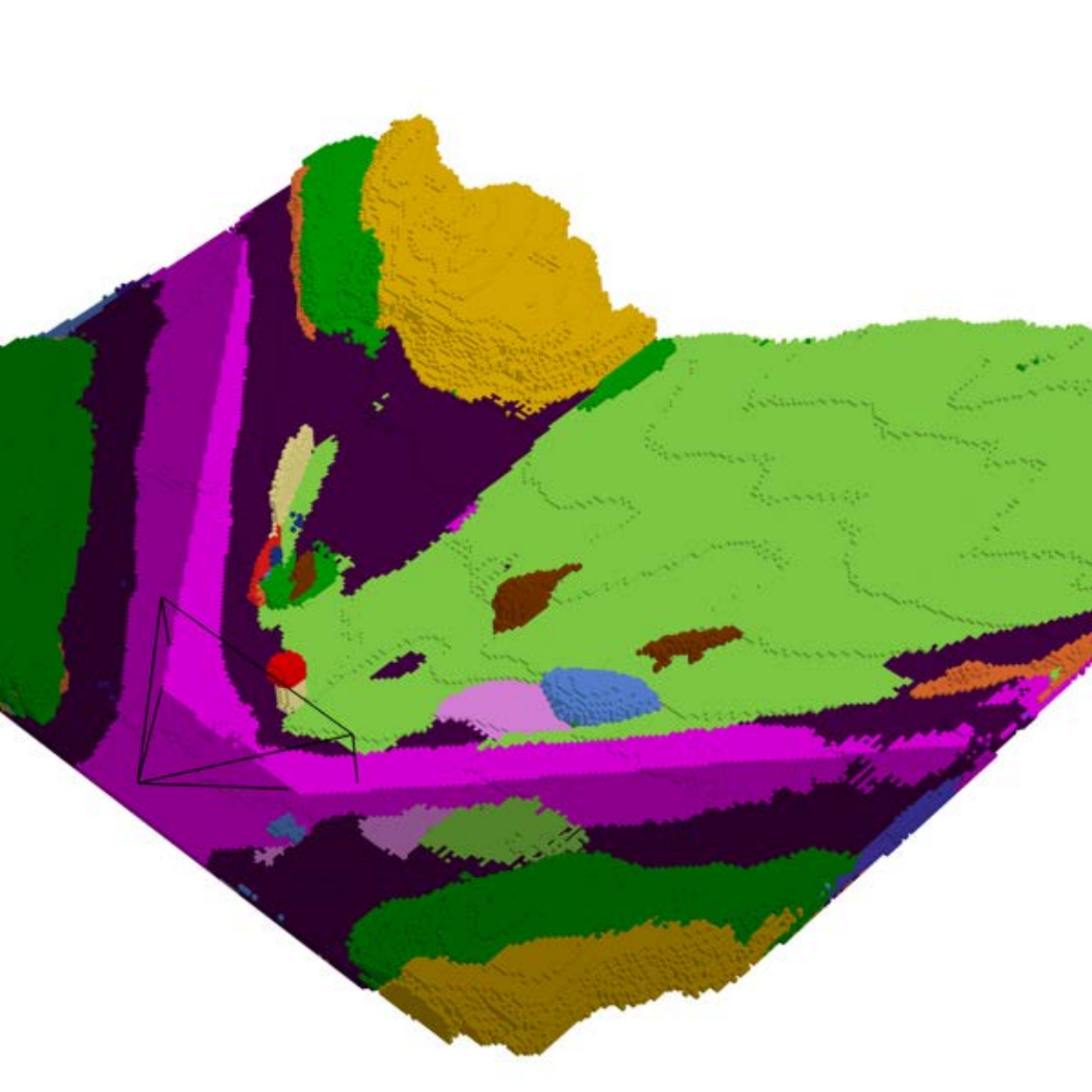} &  
		\includegraphics[width=.9\linewidth]{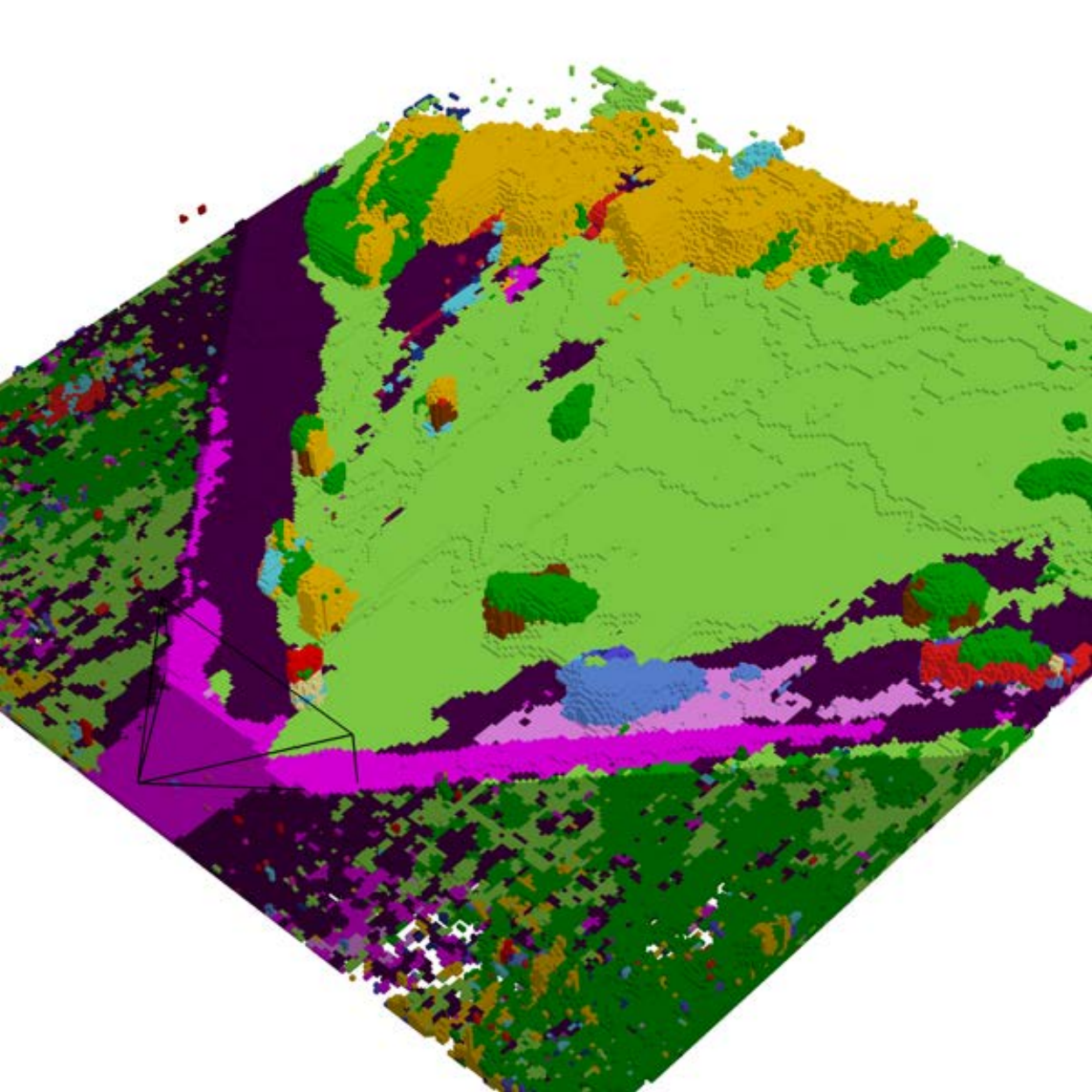} & 
		\includegraphics[width=.9\linewidth]{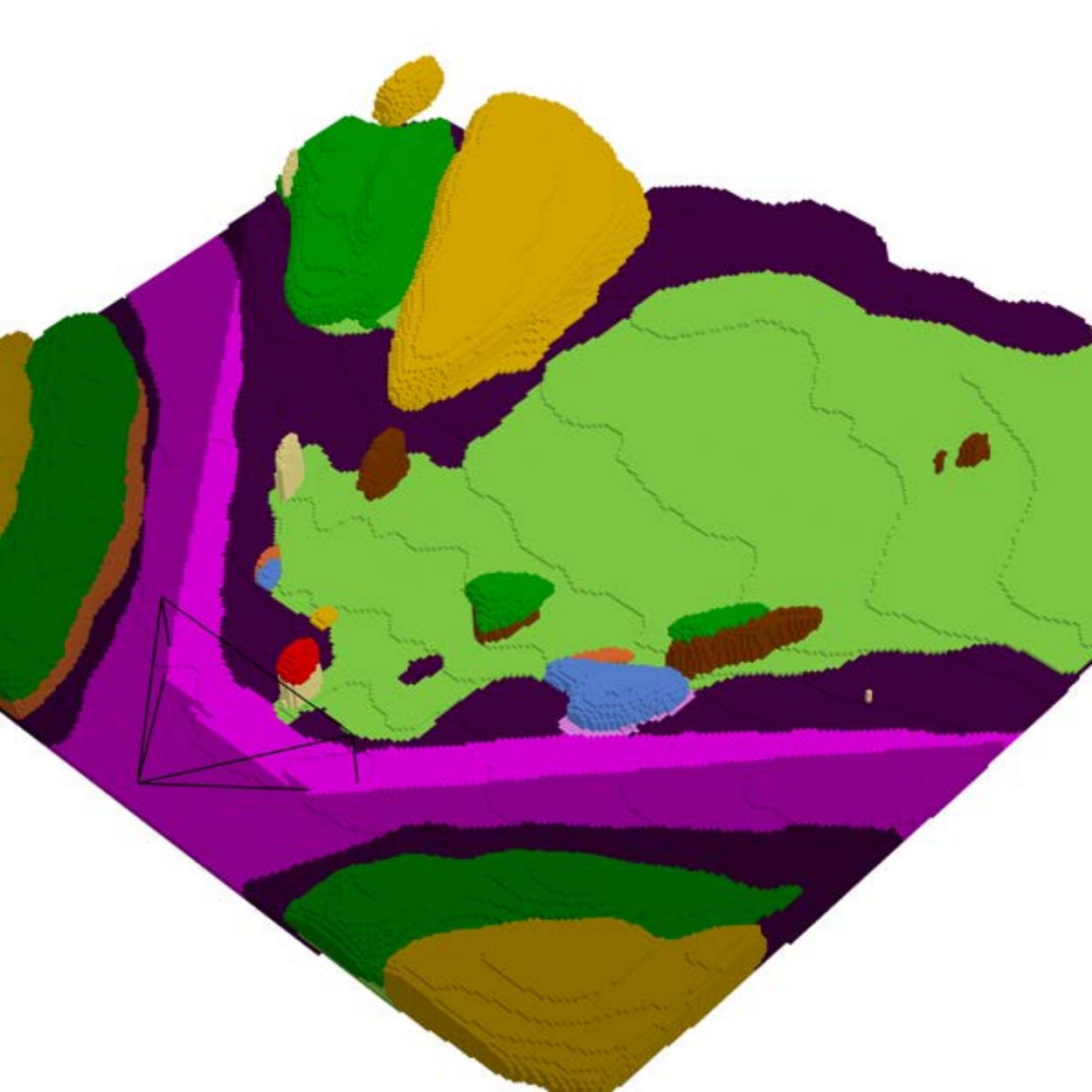} & 
		\includegraphics[width=.9\linewidth]{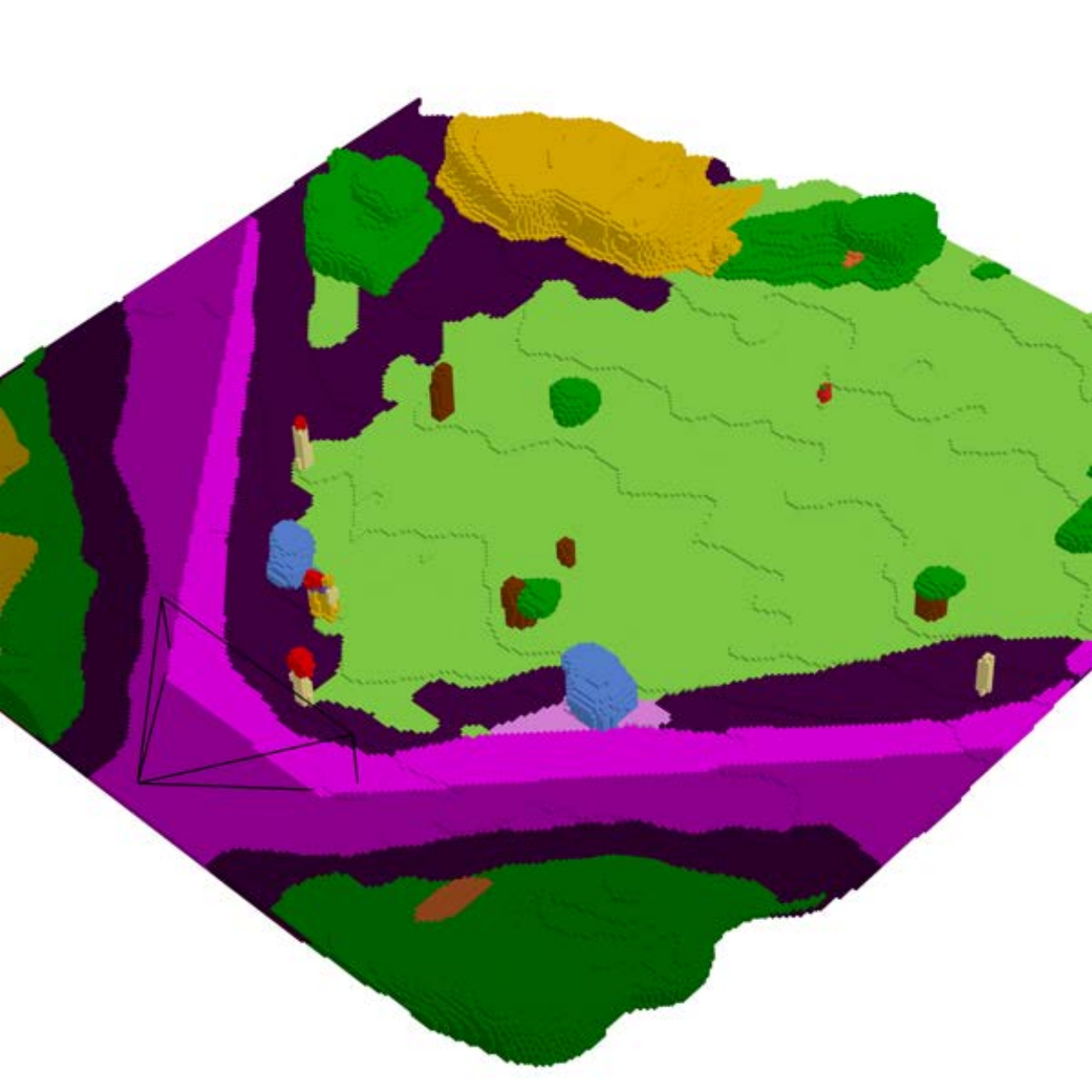} & 
		\includegraphics[width=.9\linewidth]{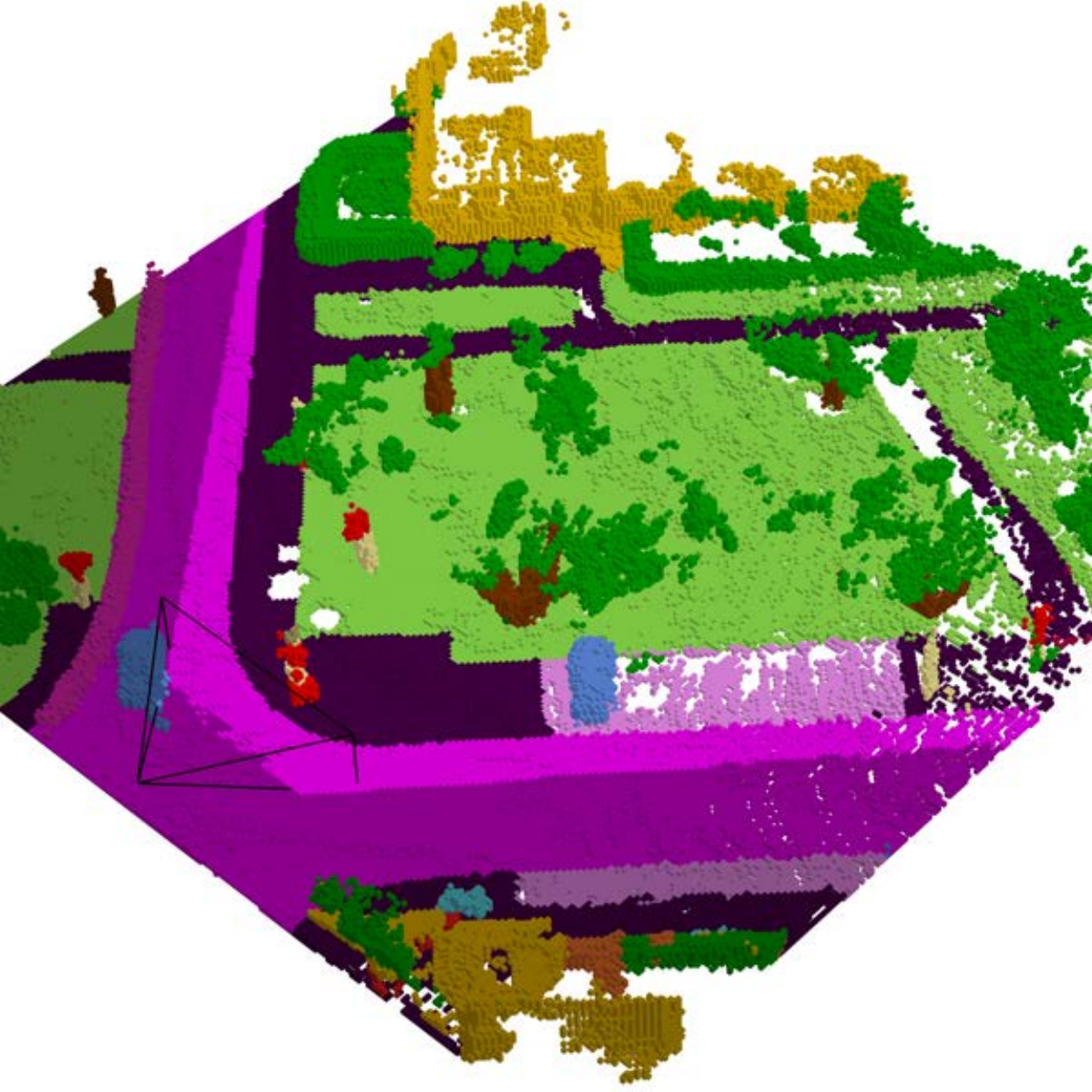} 
		\\[-0.1em]
		\includegraphics[width=.9\linewidth]{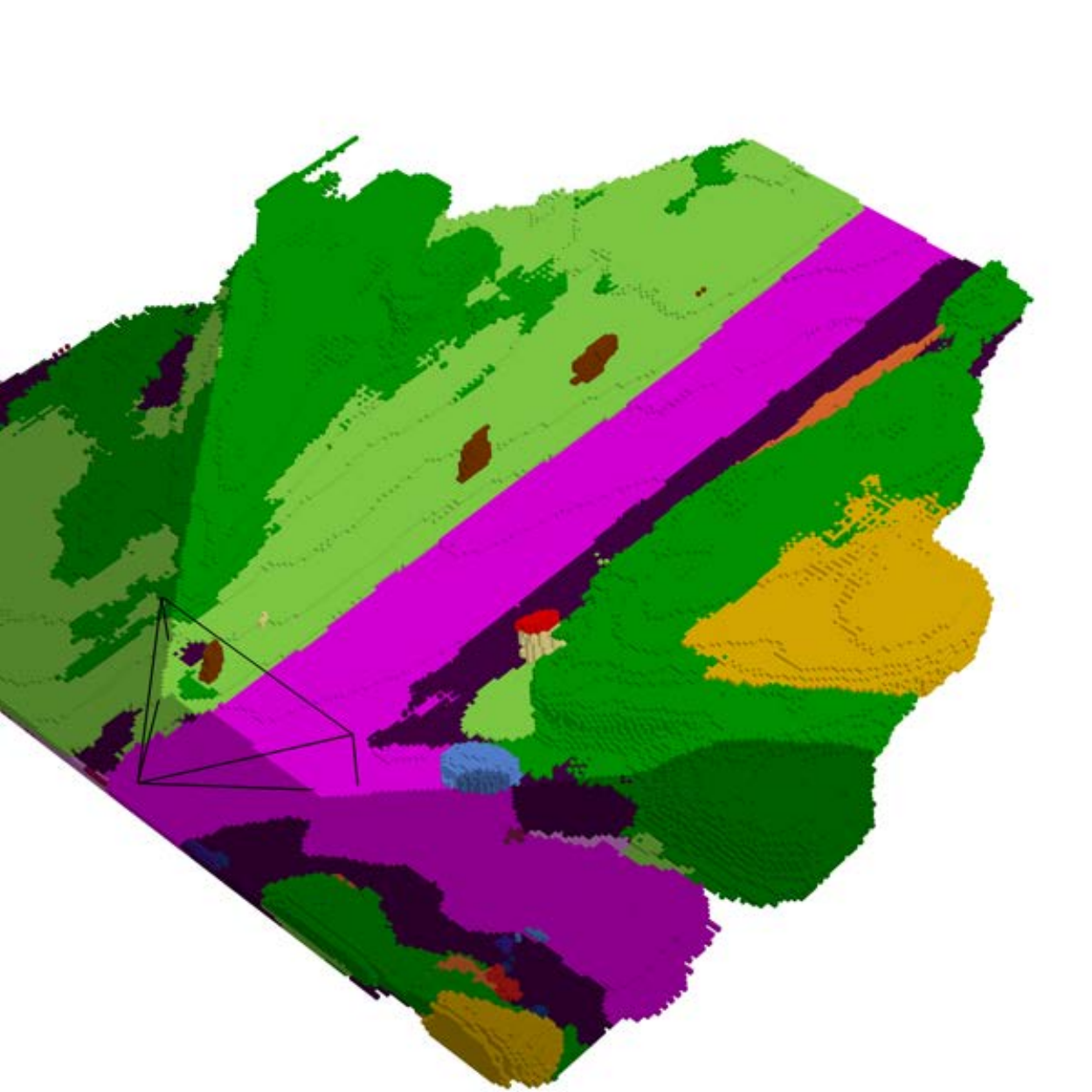} &  
		\includegraphics[width=.9\linewidth]{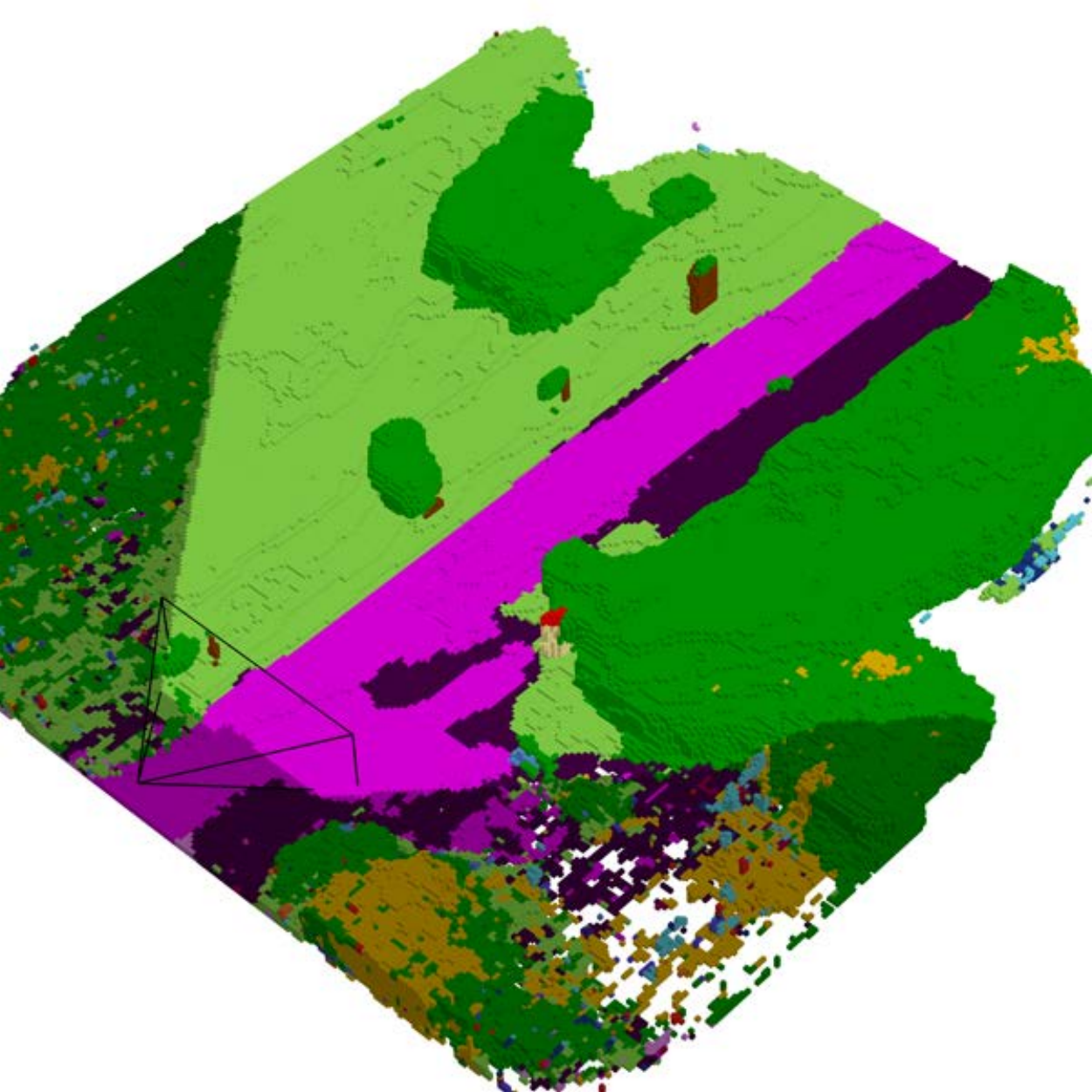} & 
		\includegraphics[width=.9\linewidth]{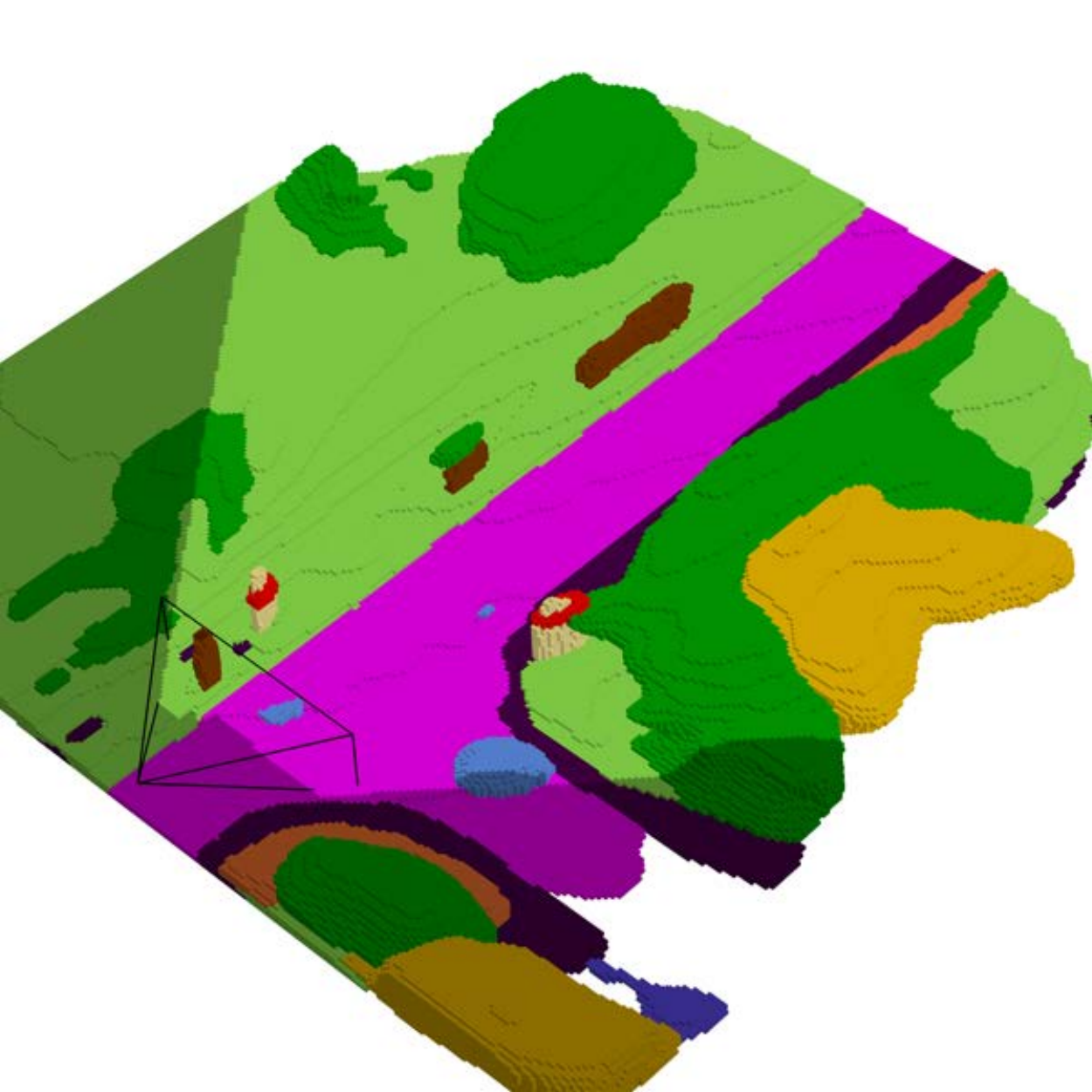} & 
		\includegraphics[width=.9\linewidth]{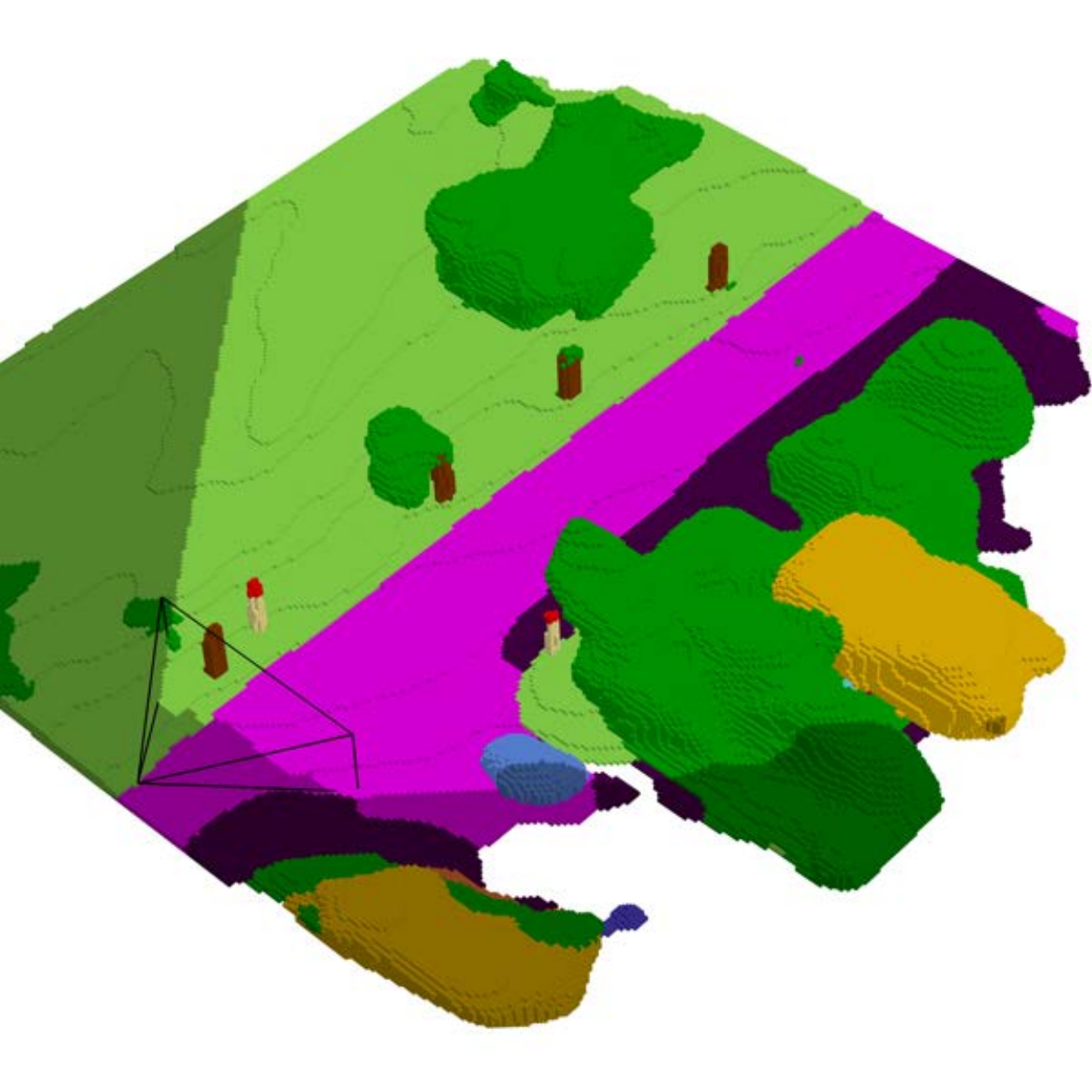} & 
		\includegraphics[width=.9\linewidth]{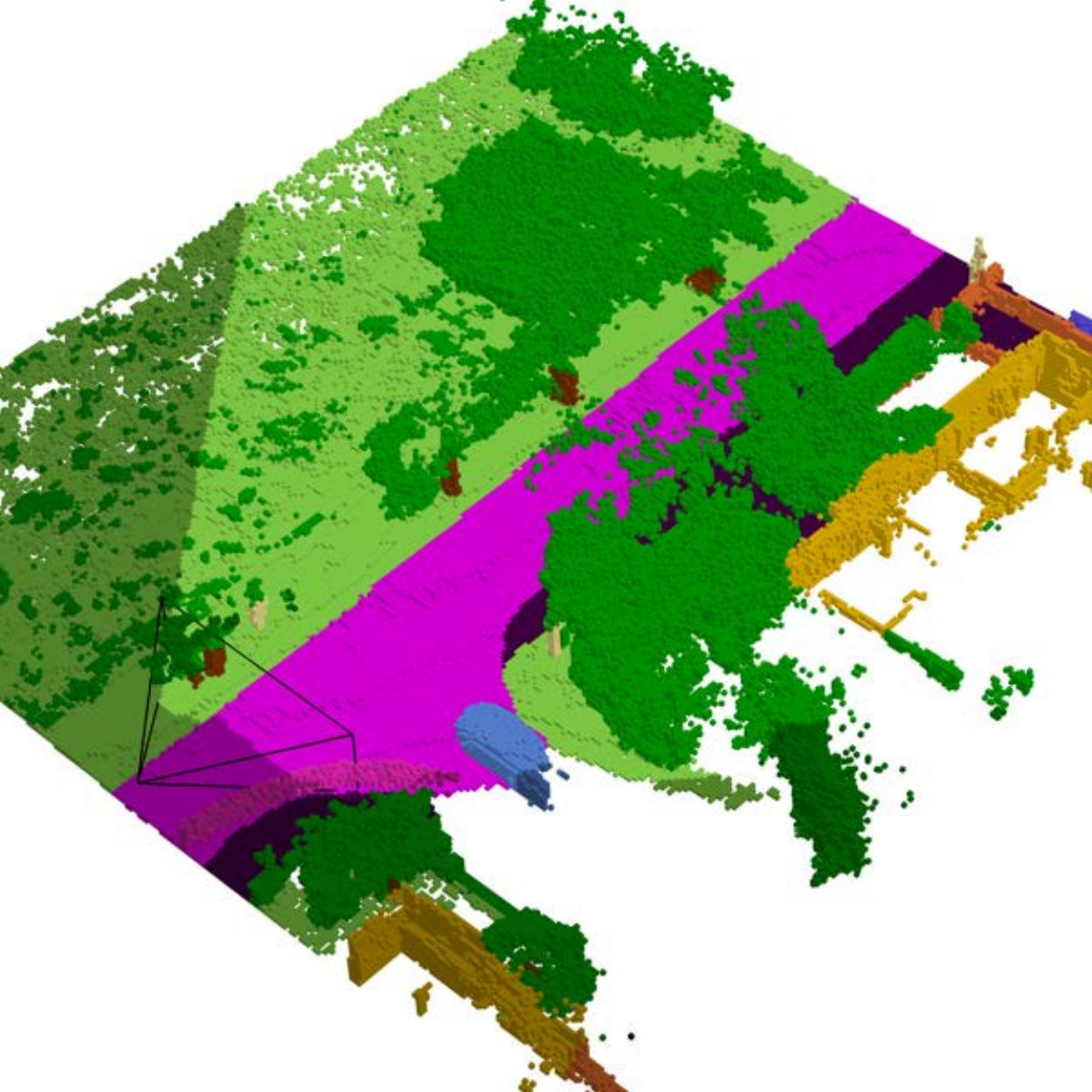} 
		\\[-0.1em]
		\includegraphics[width=.9\linewidth]{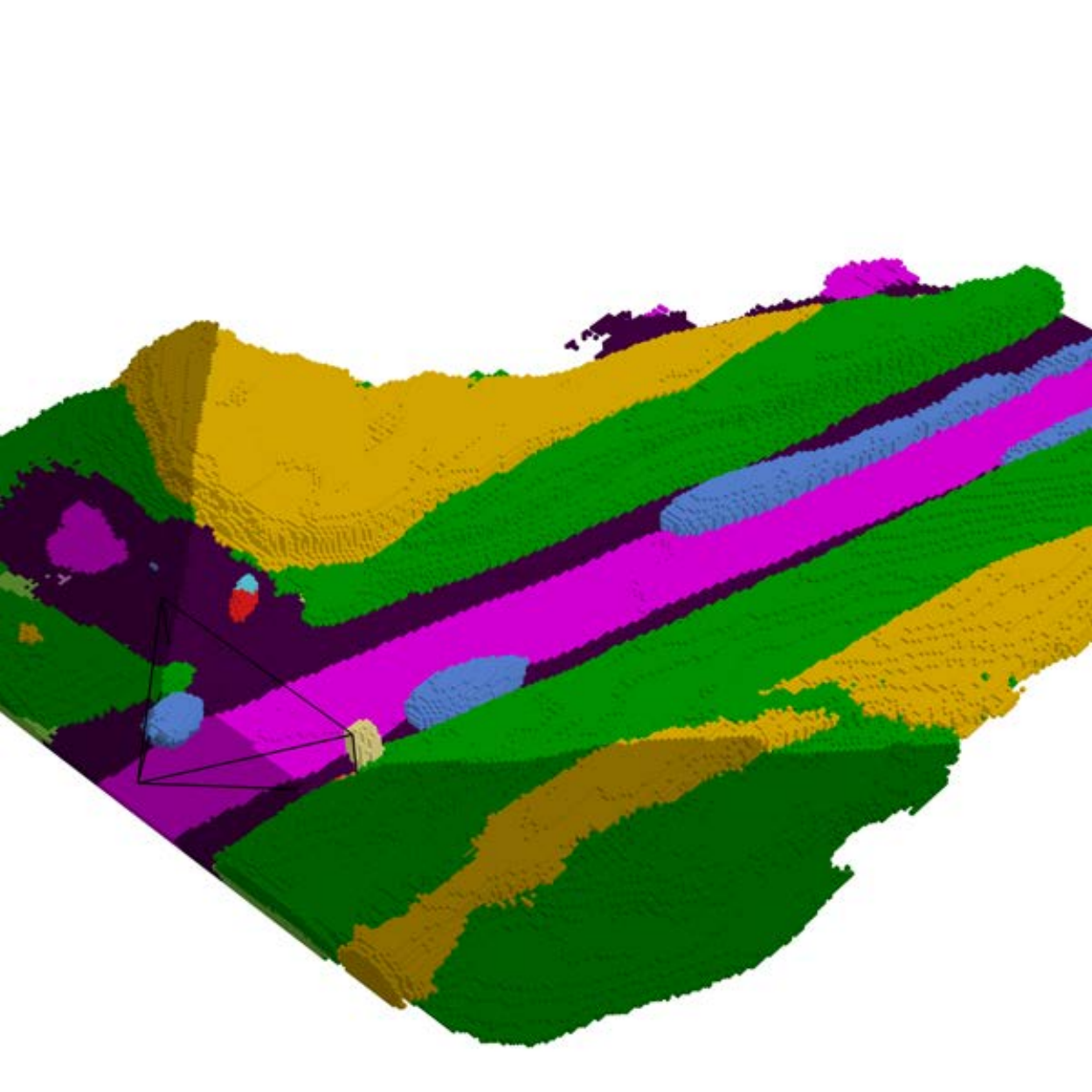} &  
		\includegraphics[width=.9\linewidth]{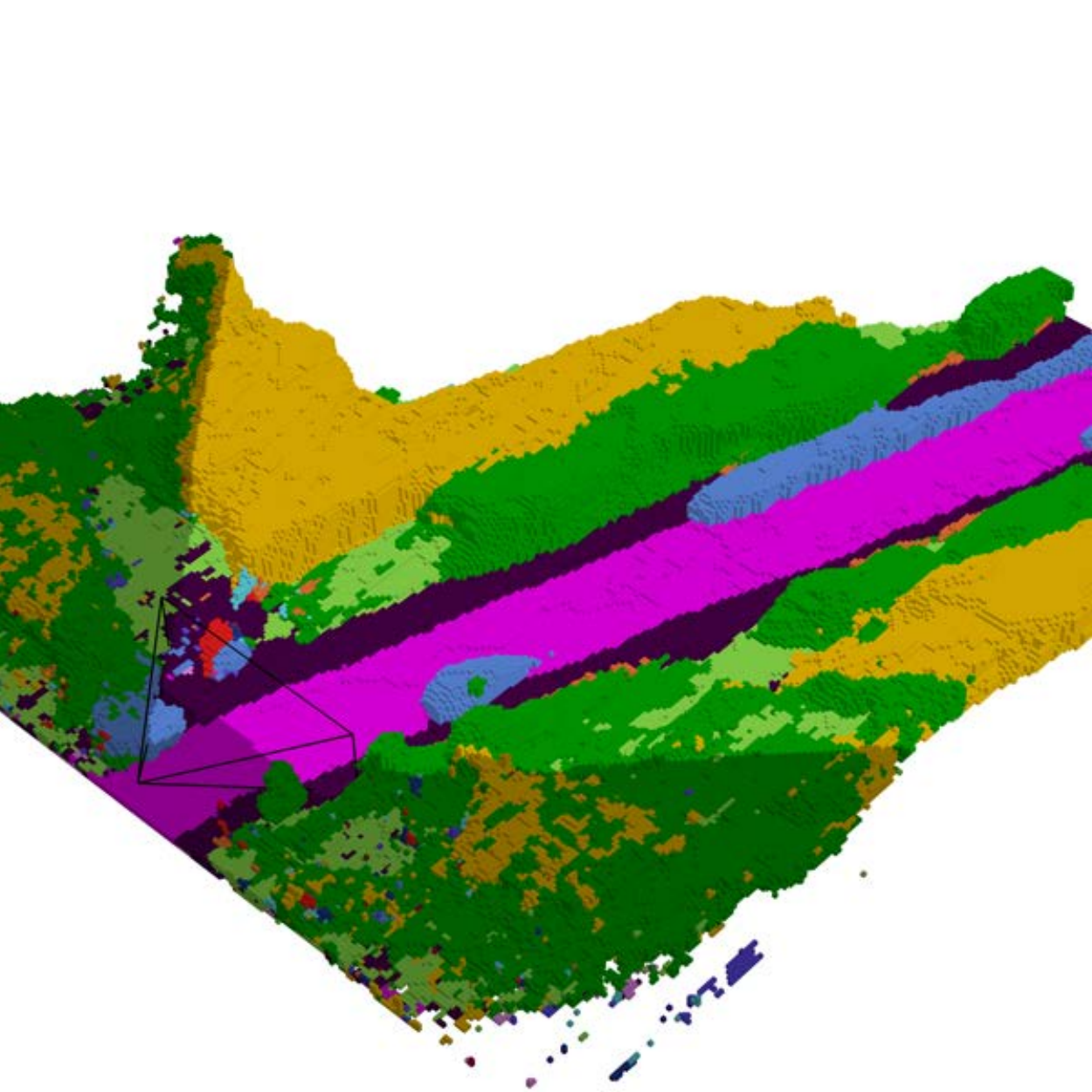} & 
		\includegraphics[width=.9\linewidth]{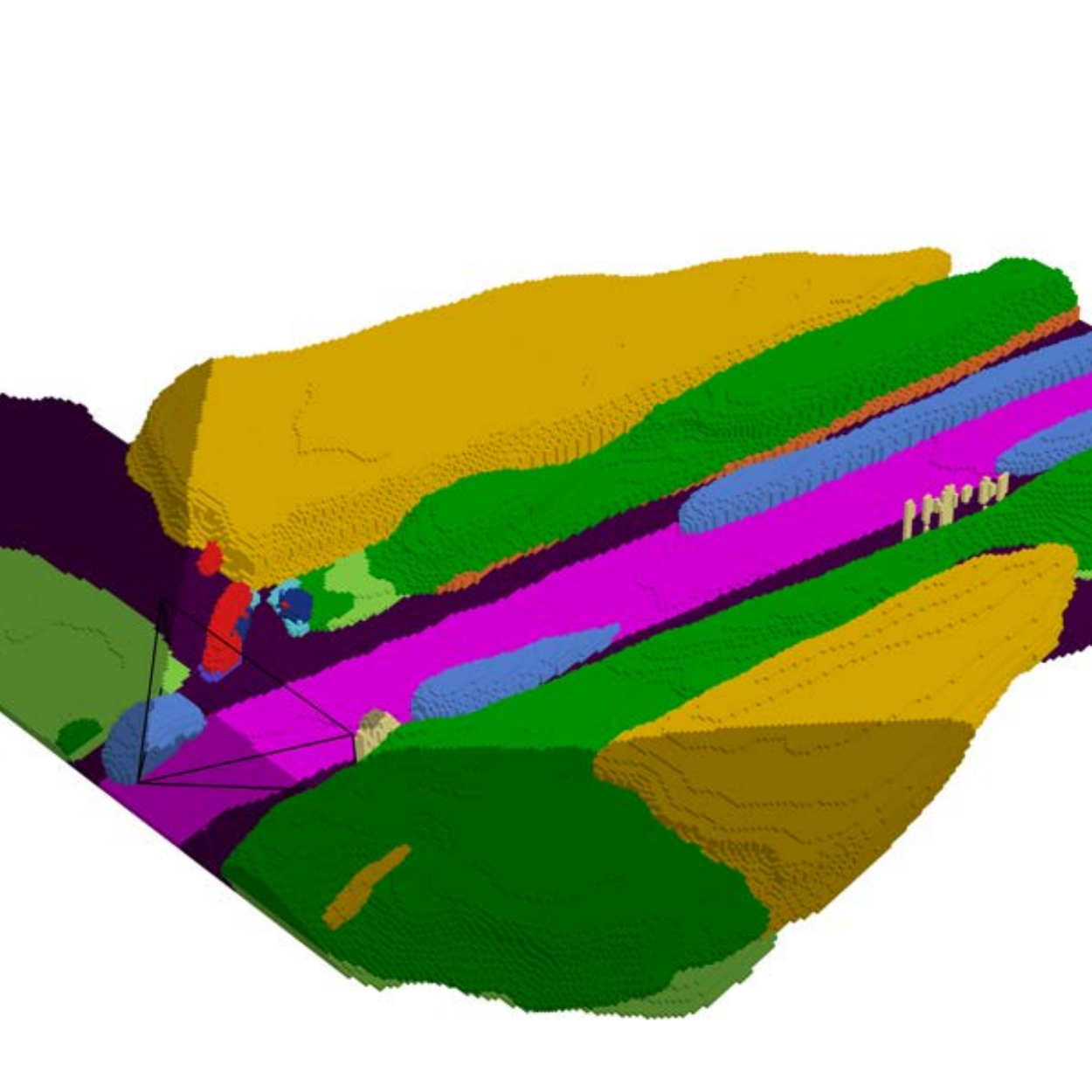} & 
		\includegraphics[width=.9\linewidth]{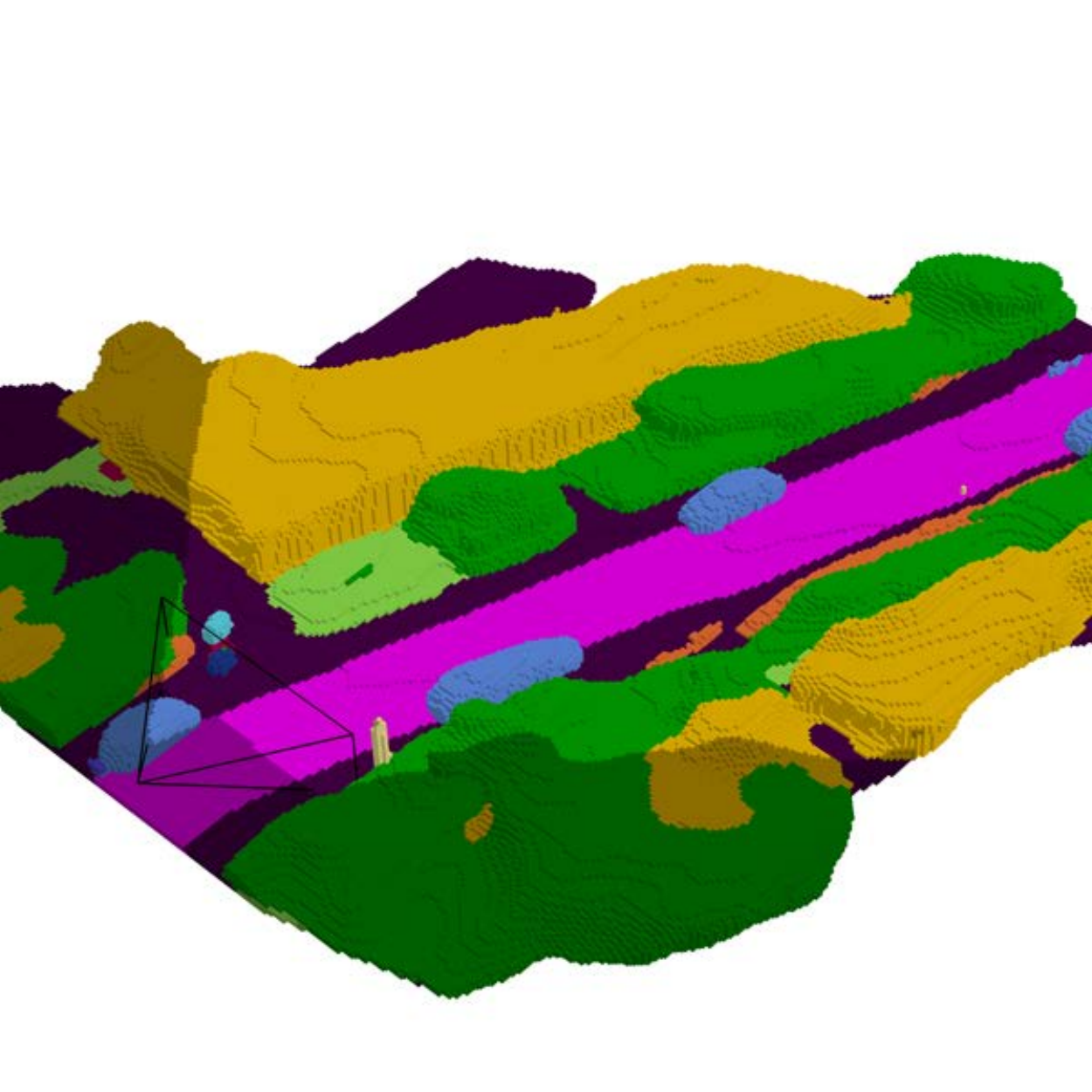} & 
		\includegraphics[width=.9\linewidth]{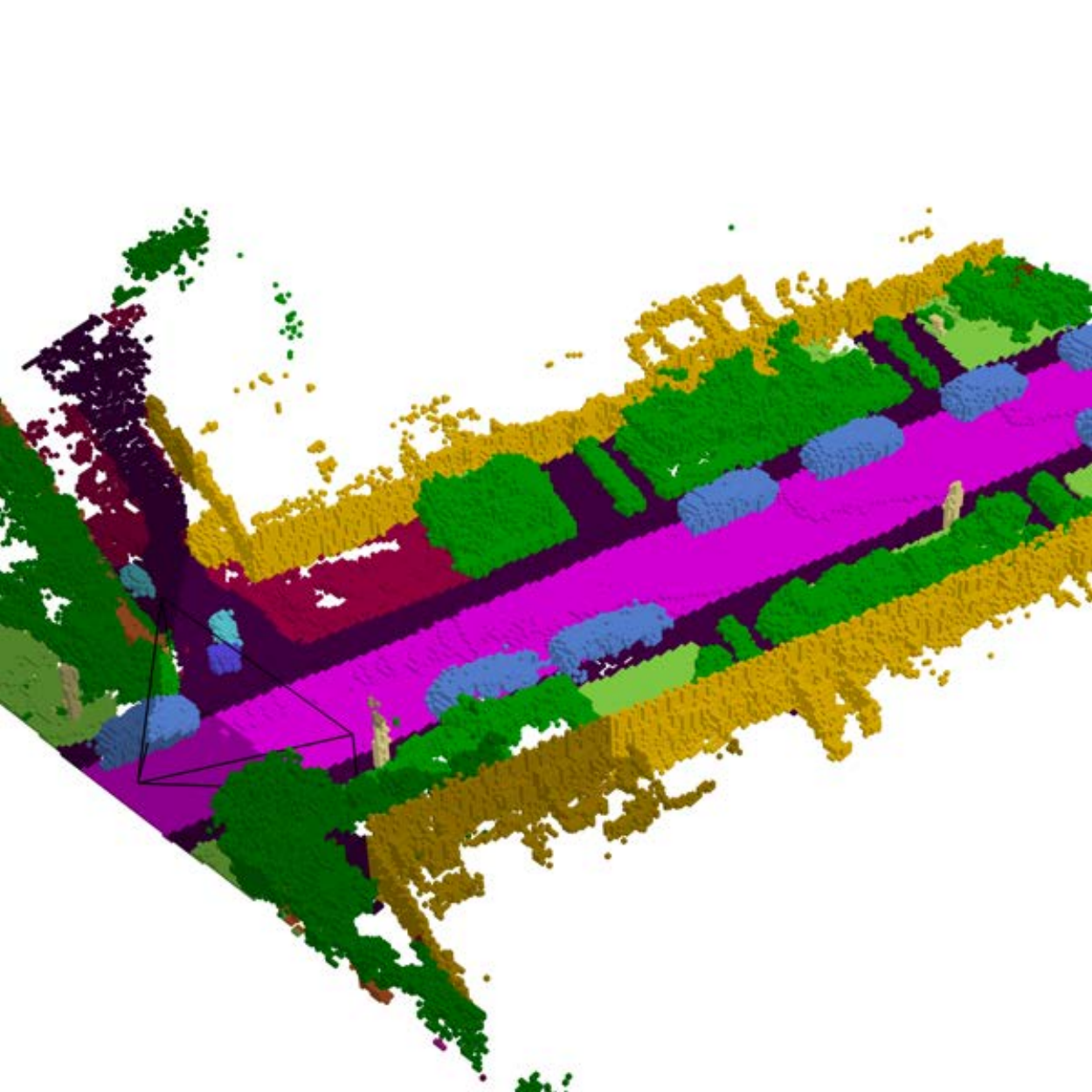} 
		\\[-0.1em]
		\includegraphics[width=.9\linewidth]{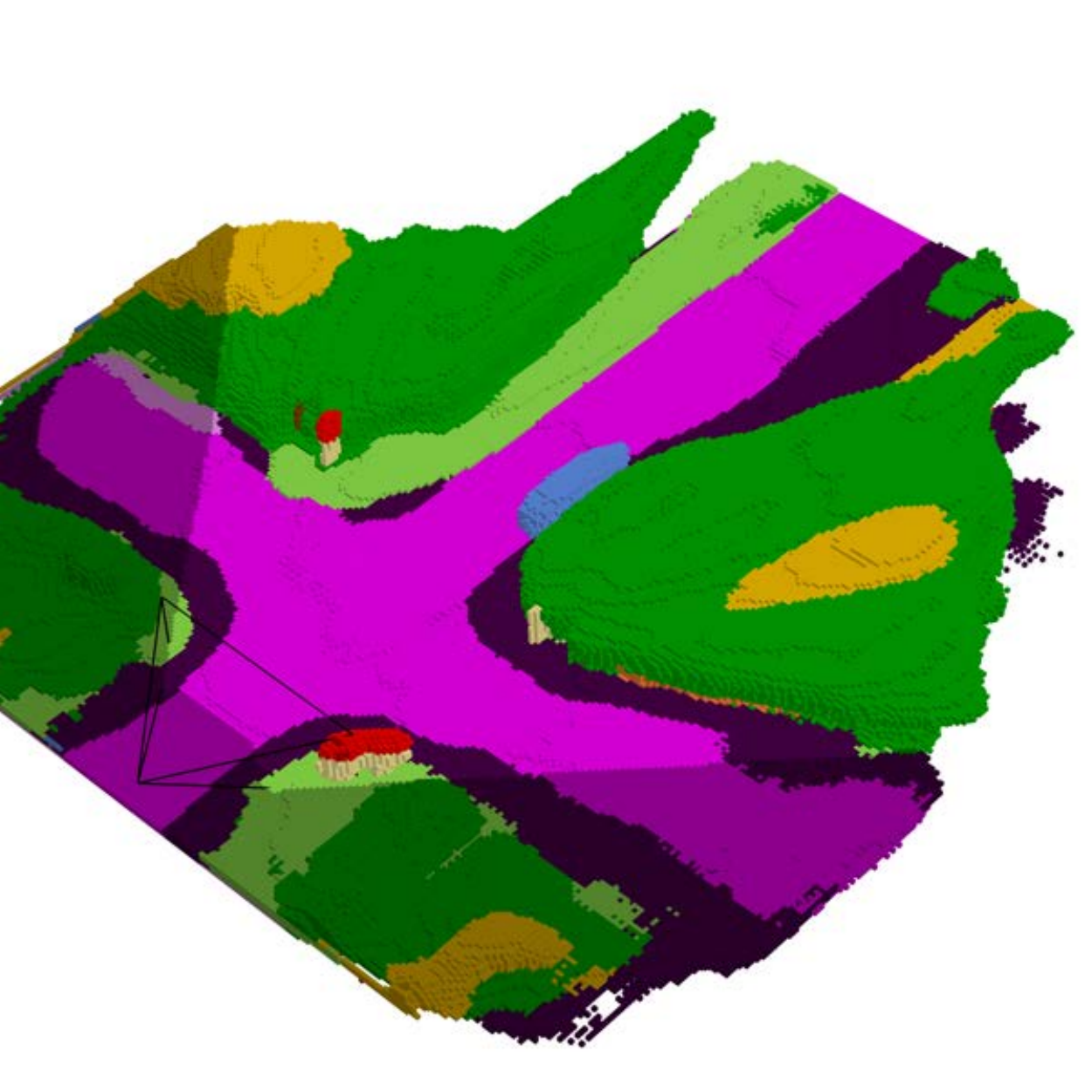} &  
		\includegraphics[width=.9\linewidth]{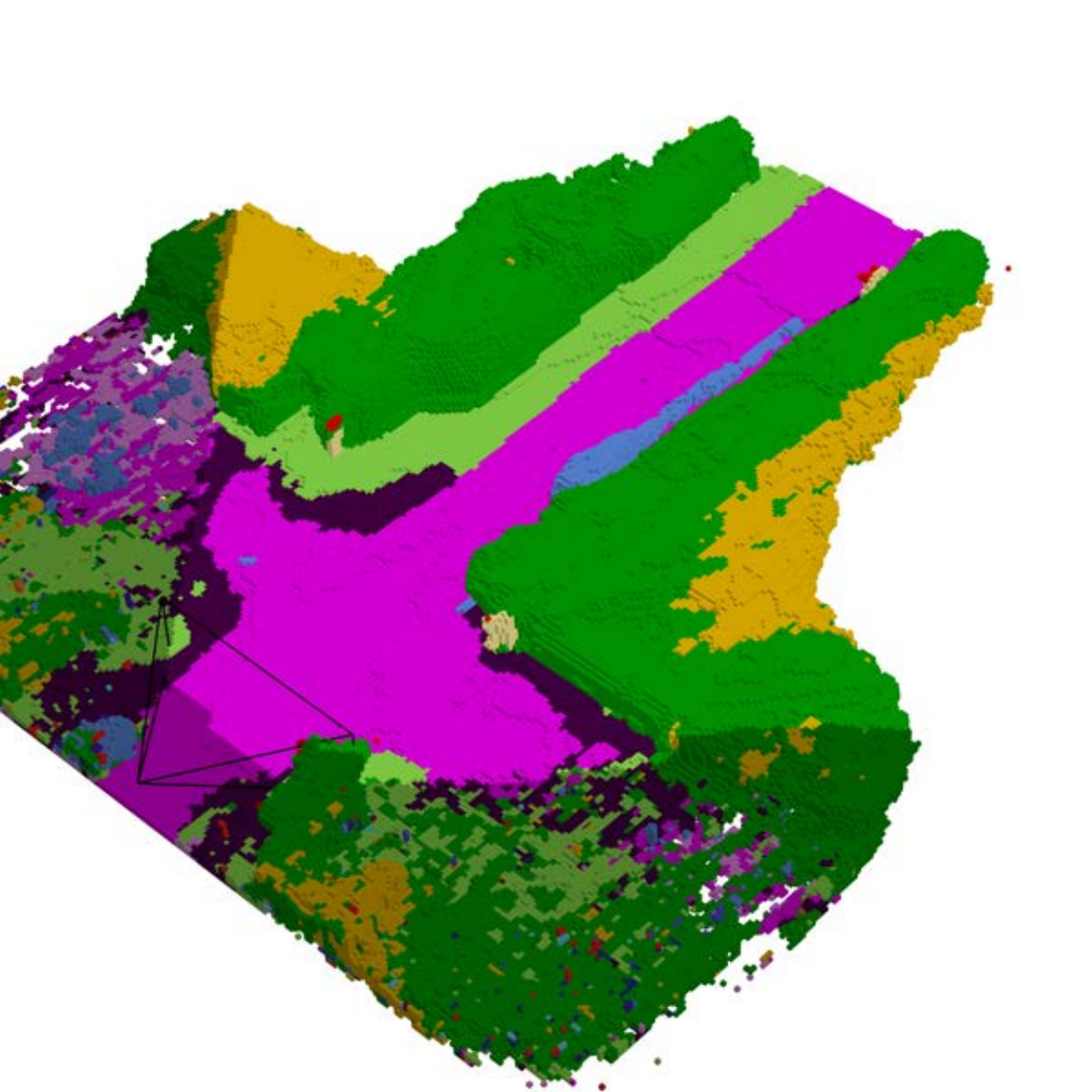} & 
		\includegraphics[width=.9\linewidth]{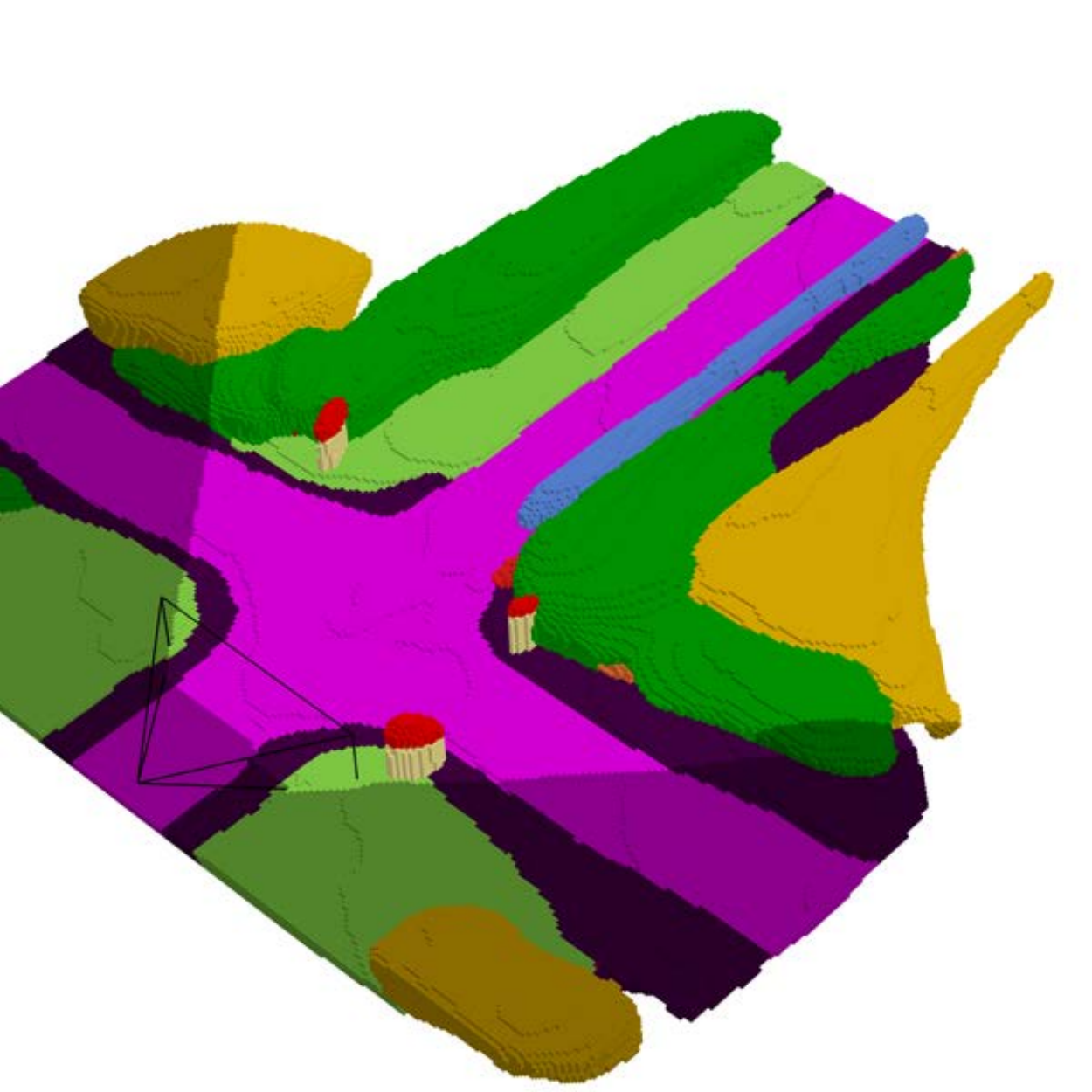} & 
		\includegraphics[width=.9\linewidth]{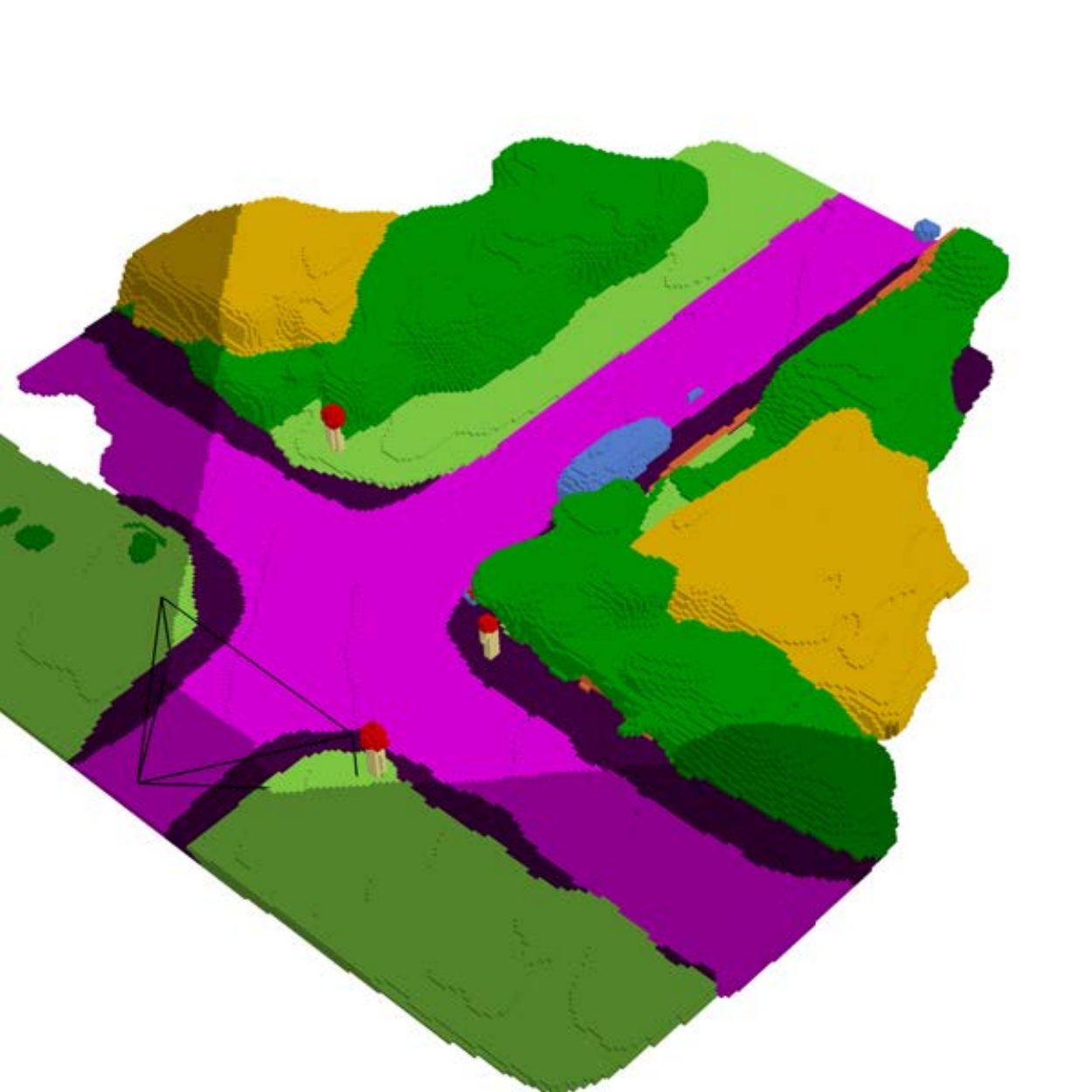} & 
		\includegraphics[width=.9\linewidth]{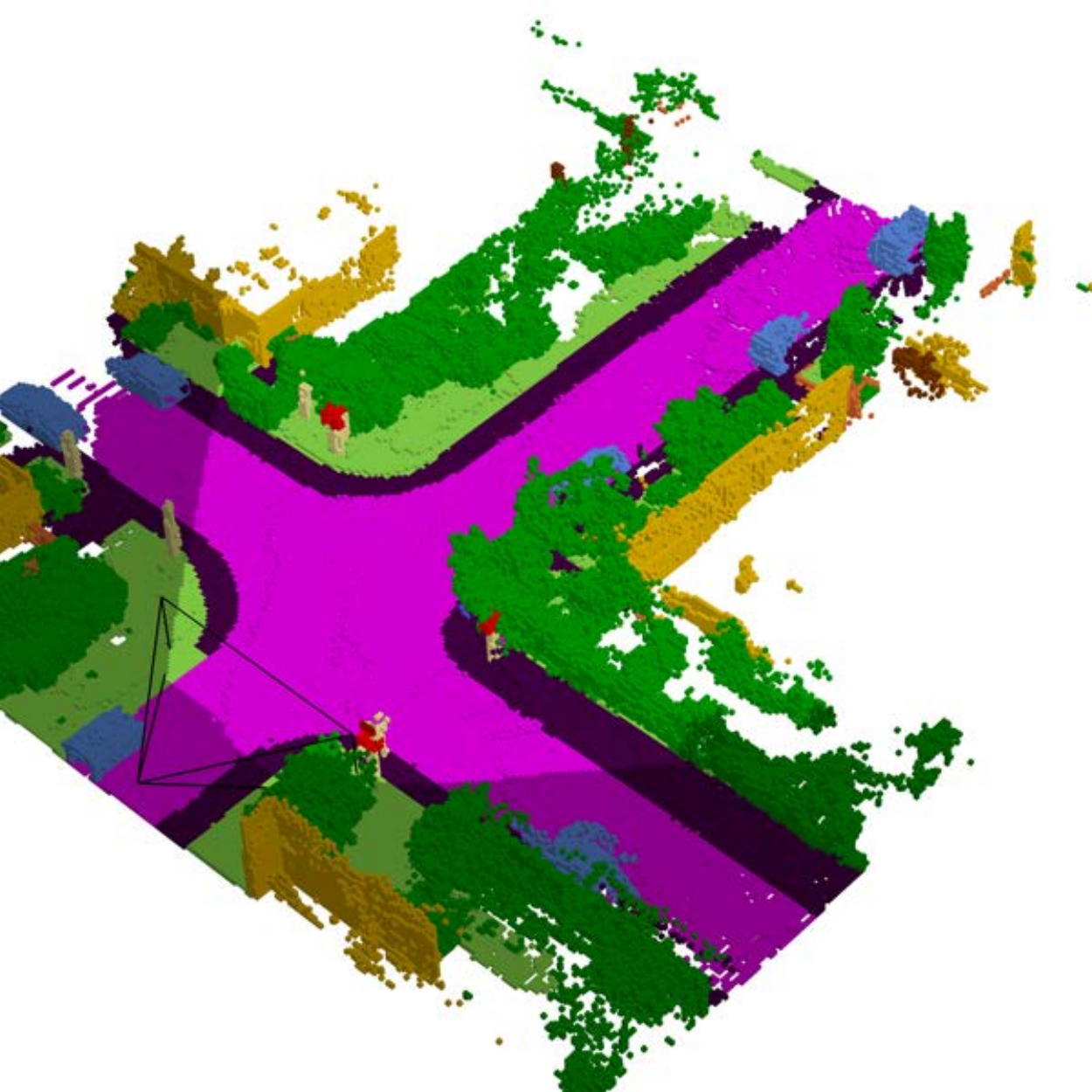} 
		\\[-0.1em]
		\includegraphics[width=.9\linewidth]{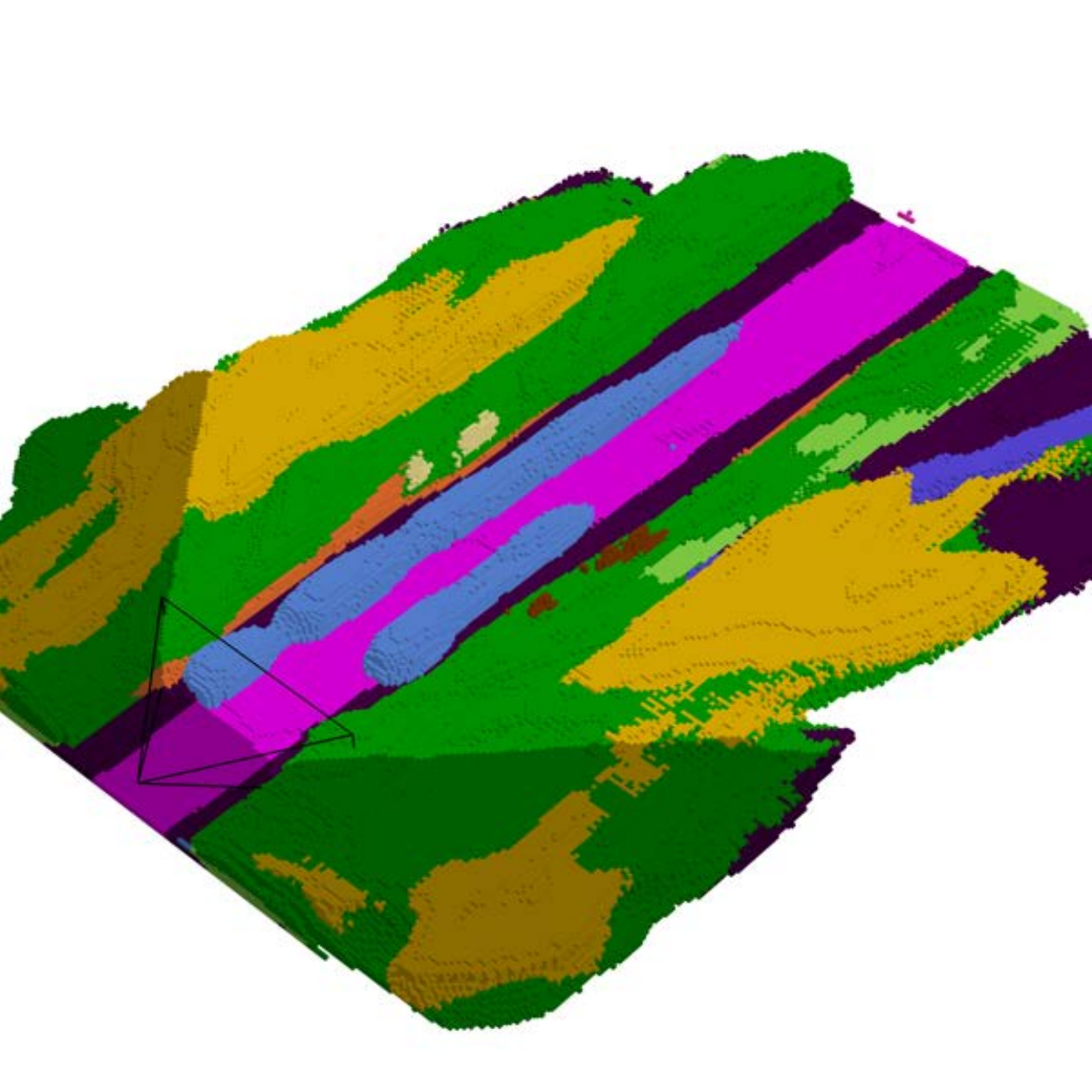} &  
		\includegraphics[width=.9\linewidth]{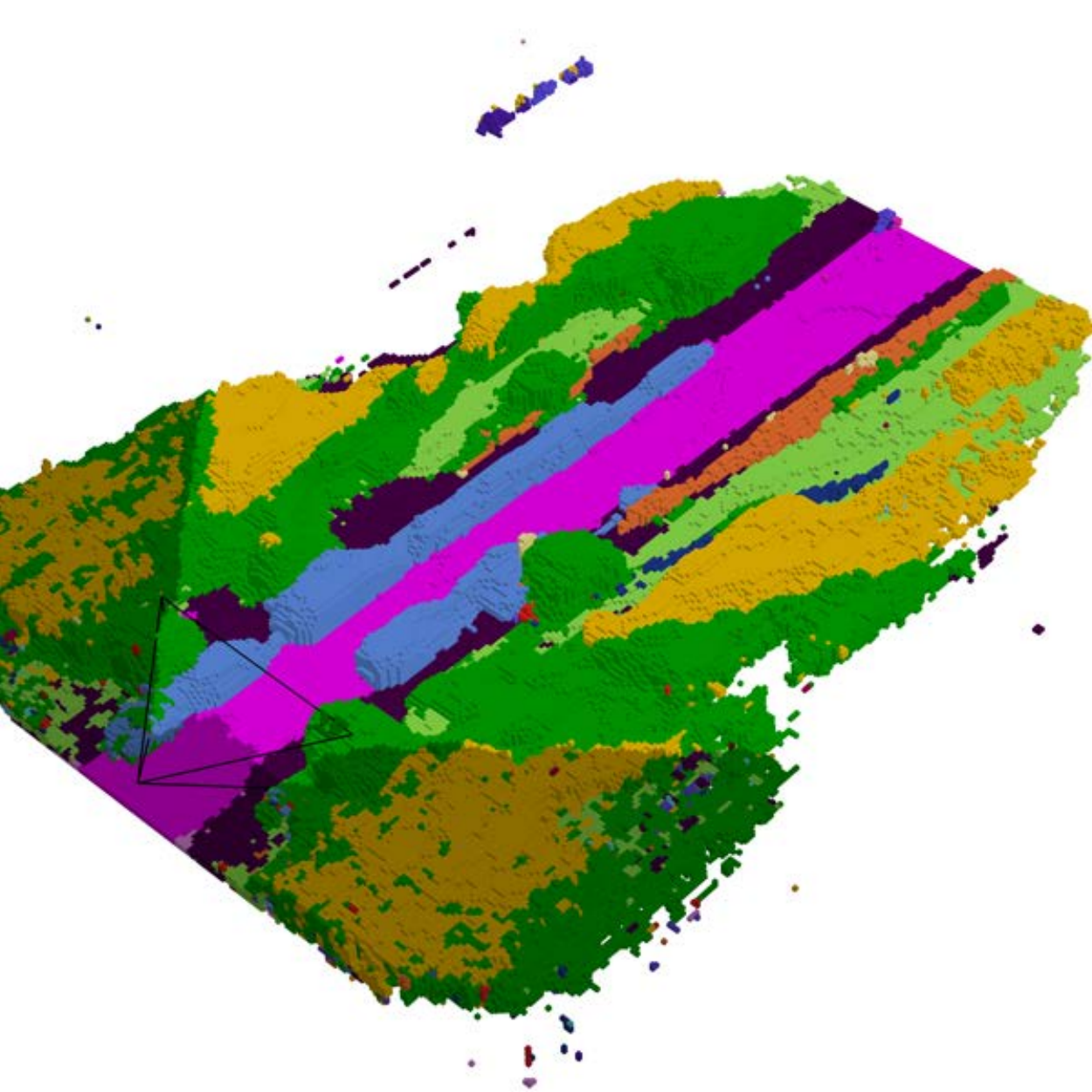} & 
		\includegraphics[width=.9\linewidth]{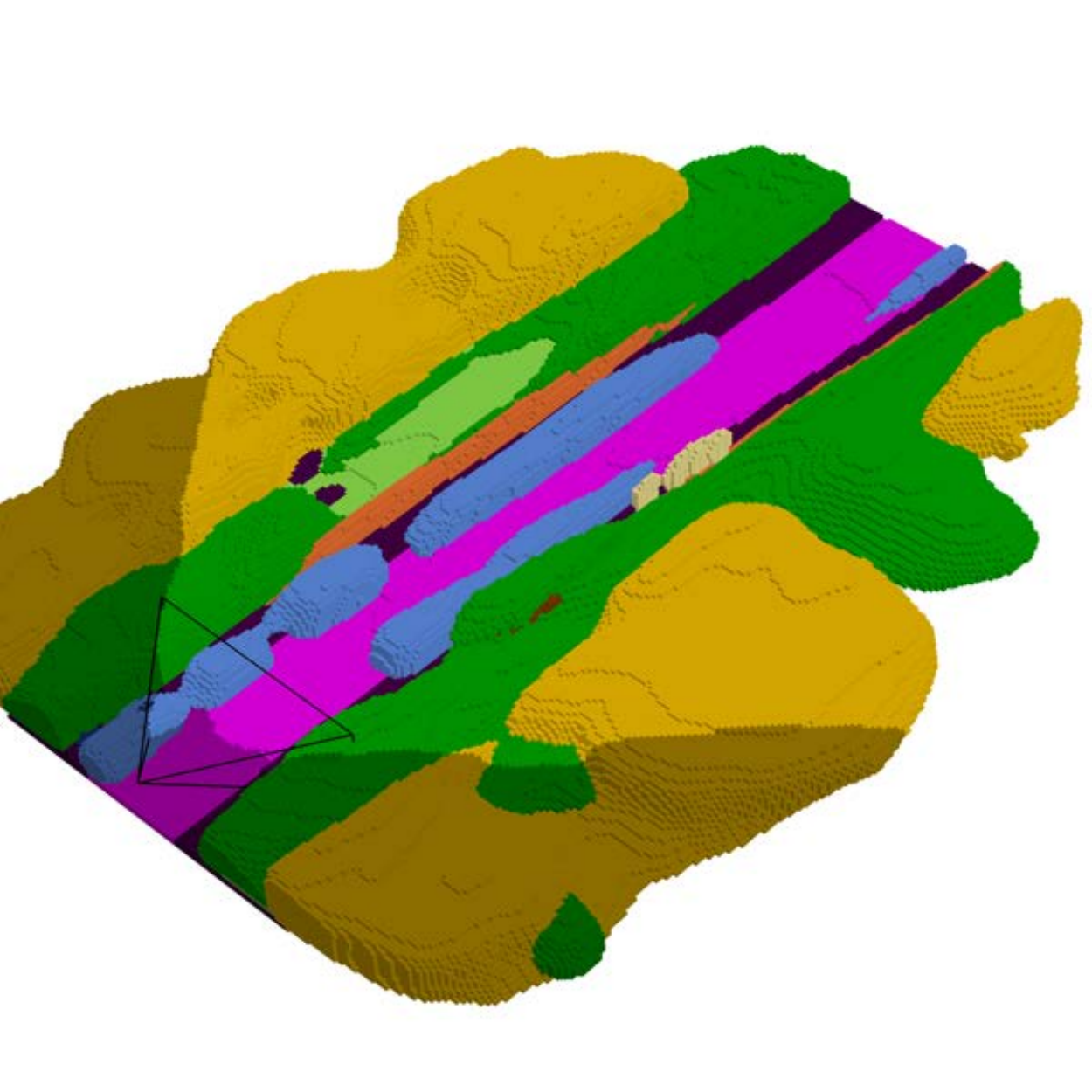} & 
		\includegraphics[width=.9\linewidth]{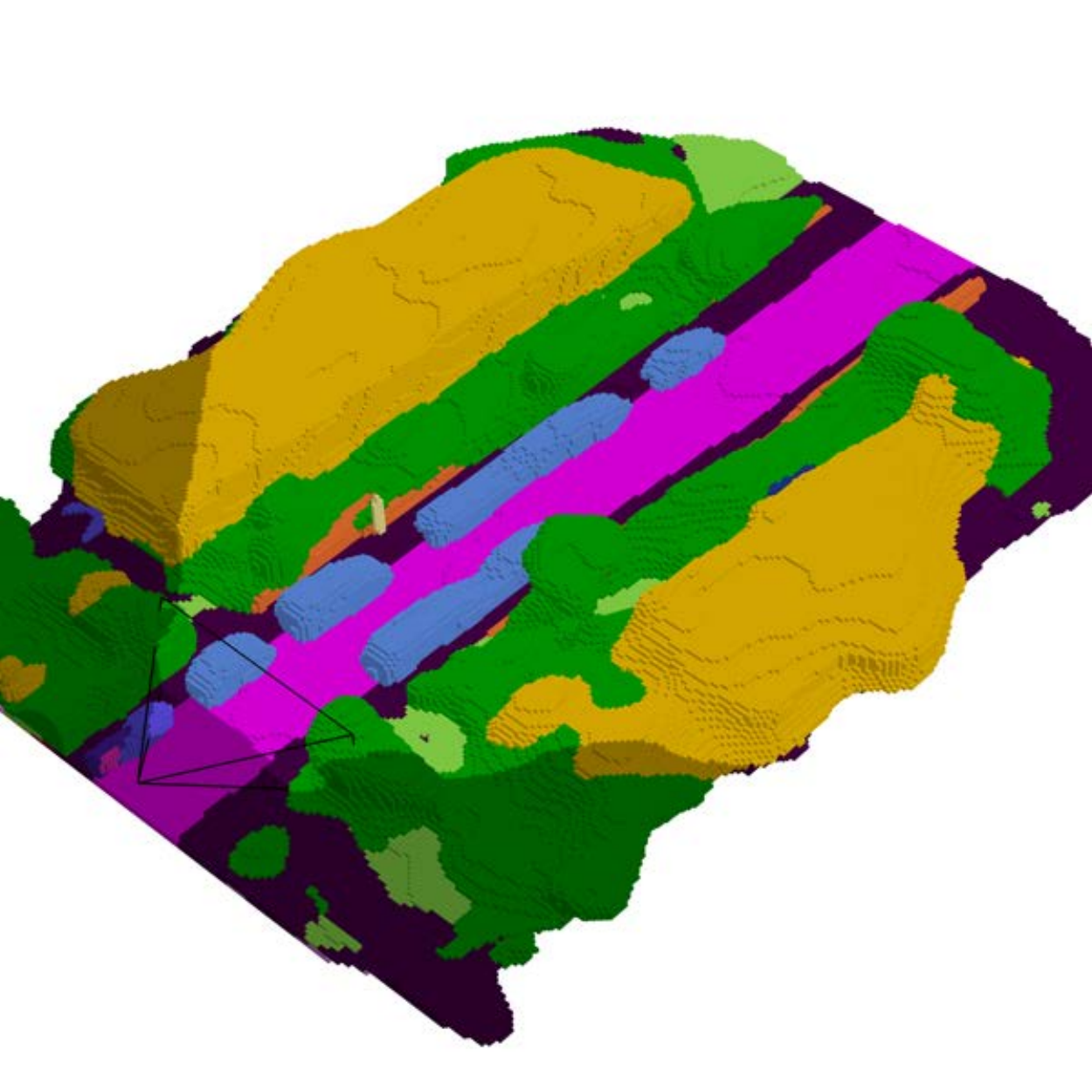} & 
		\includegraphics[width=.9\linewidth]{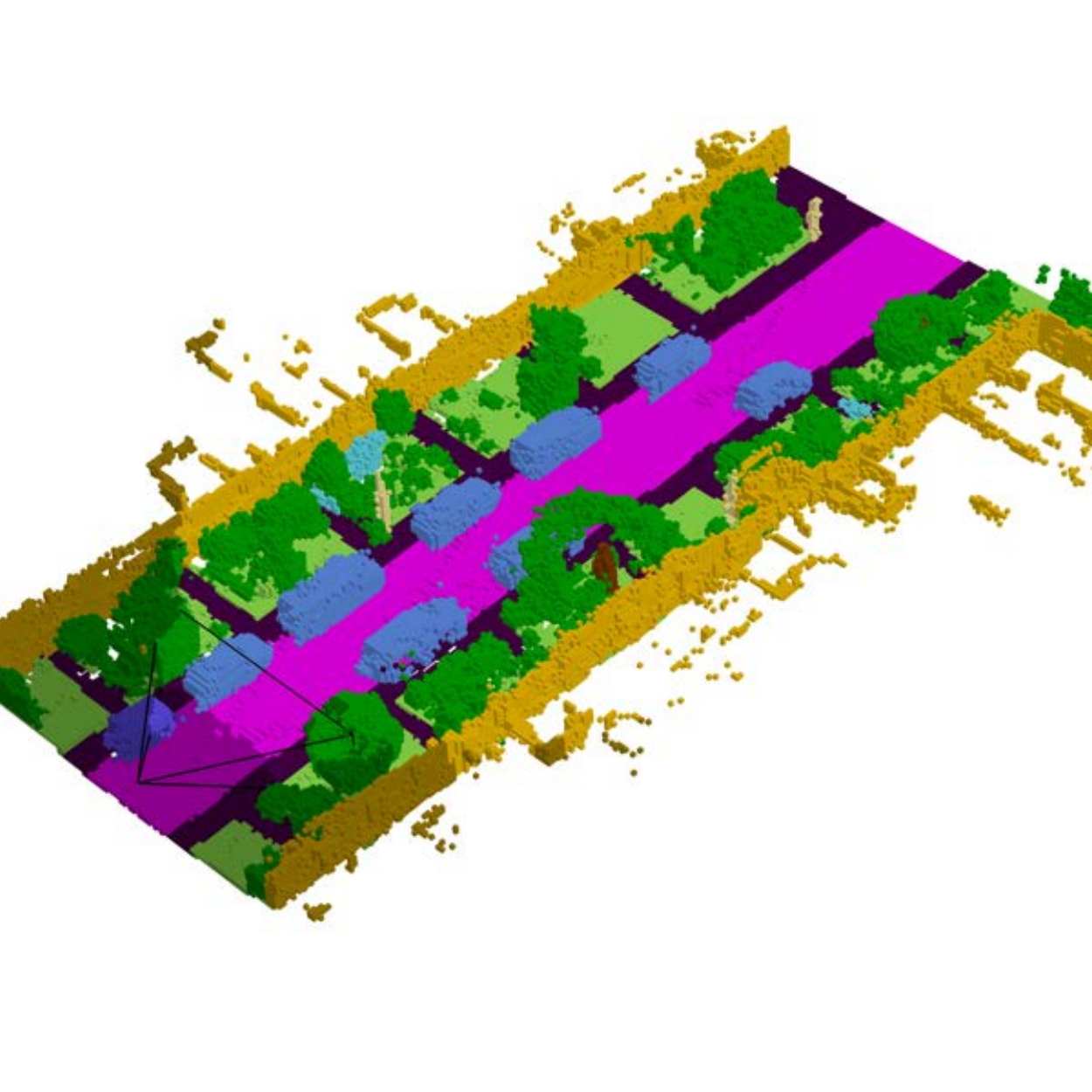} 
		\\[-0.1em]
		\includegraphics[width=.9\linewidth]{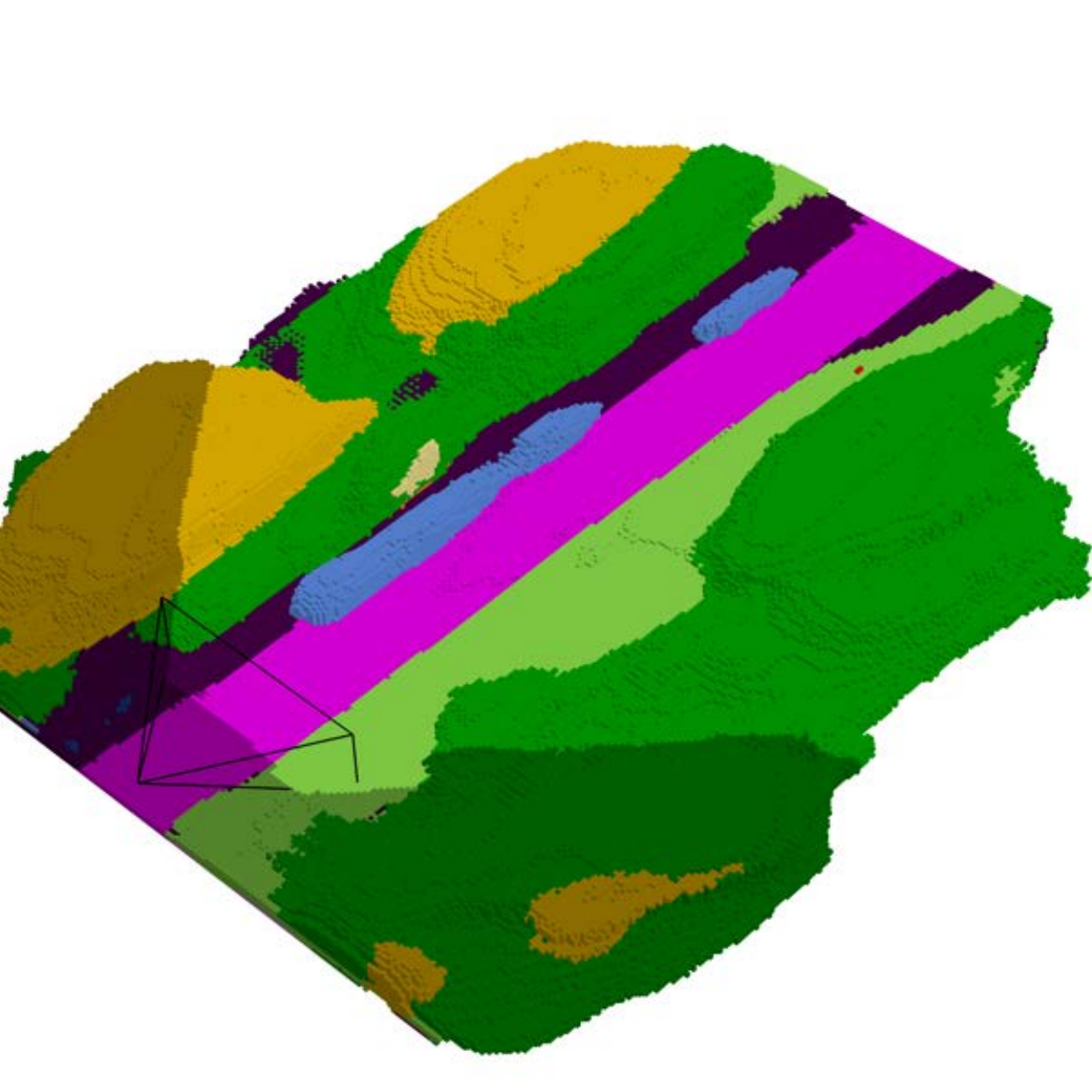} &  
		\includegraphics[width=.9\linewidth]{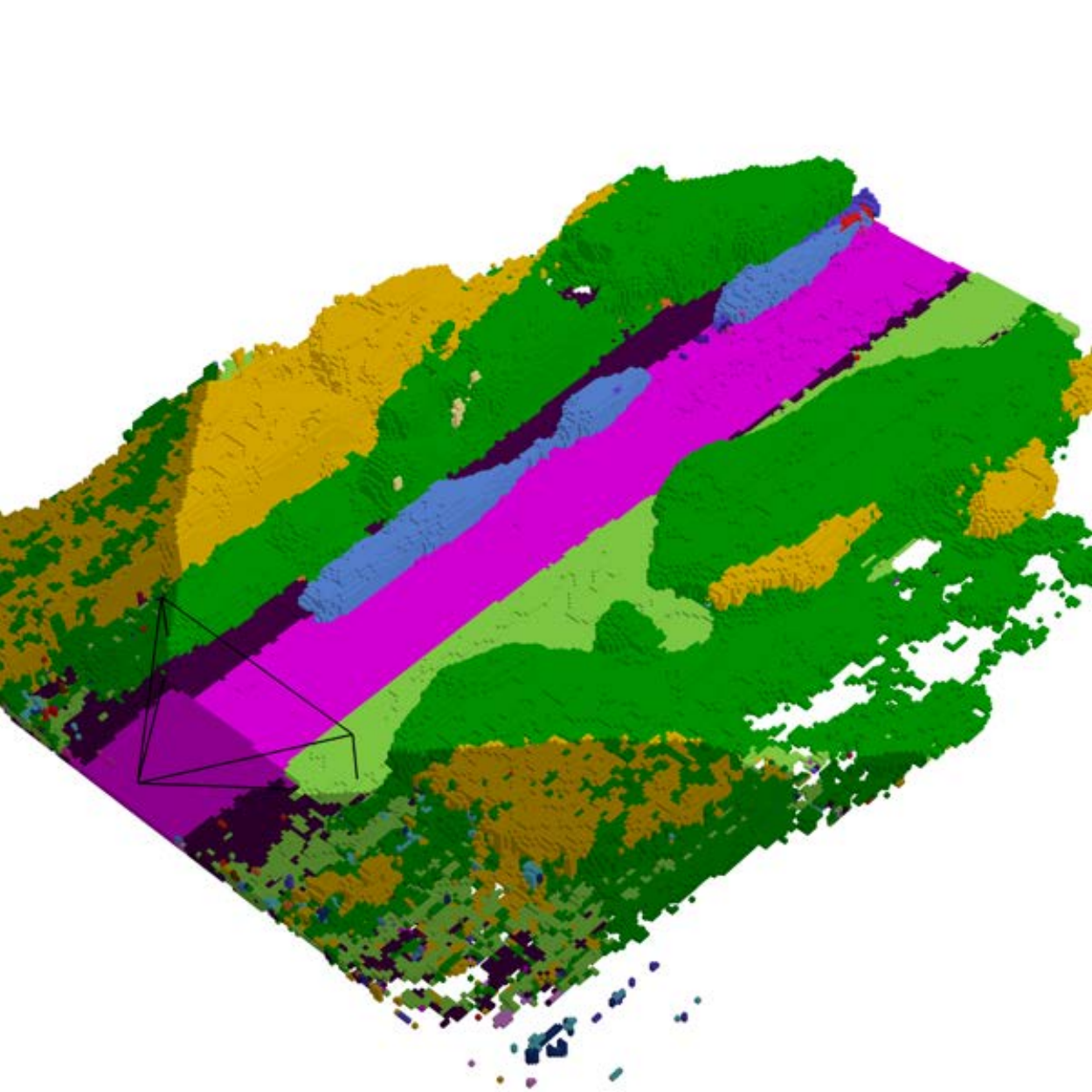} & 
		\includegraphics[width=.9\linewidth]{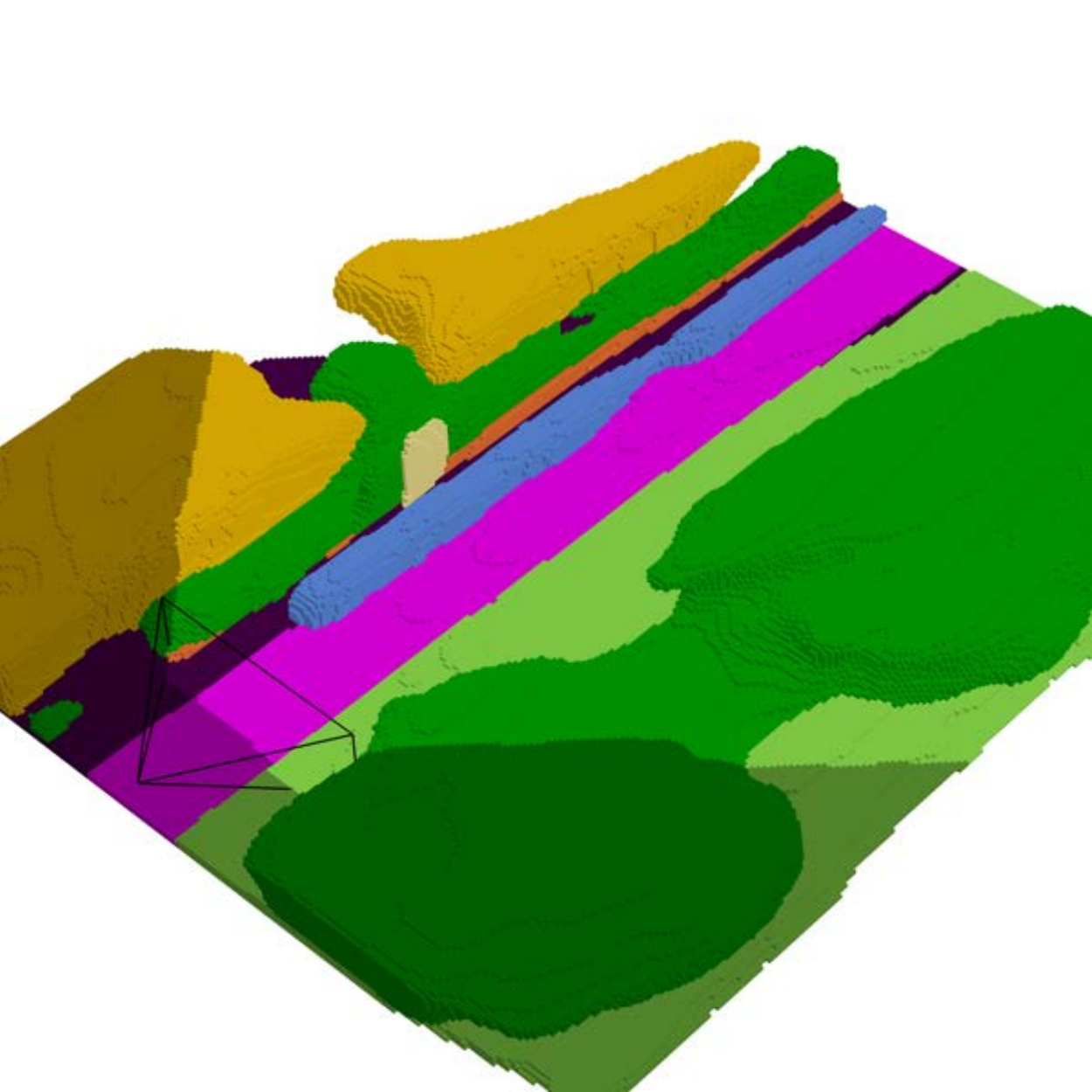} & 
		\includegraphics[width=.9\linewidth]{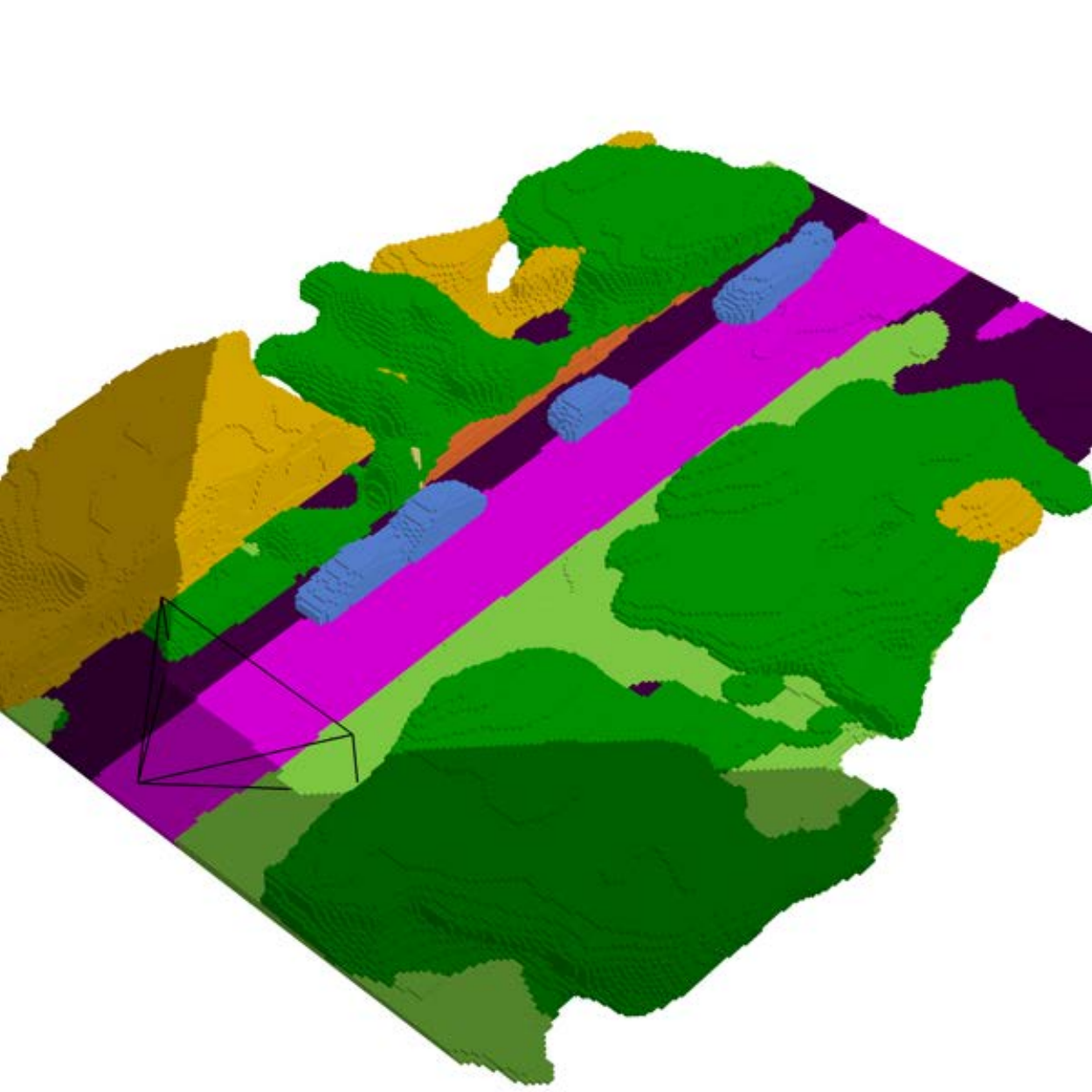} & 
		\includegraphics[width=.9\linewidth]{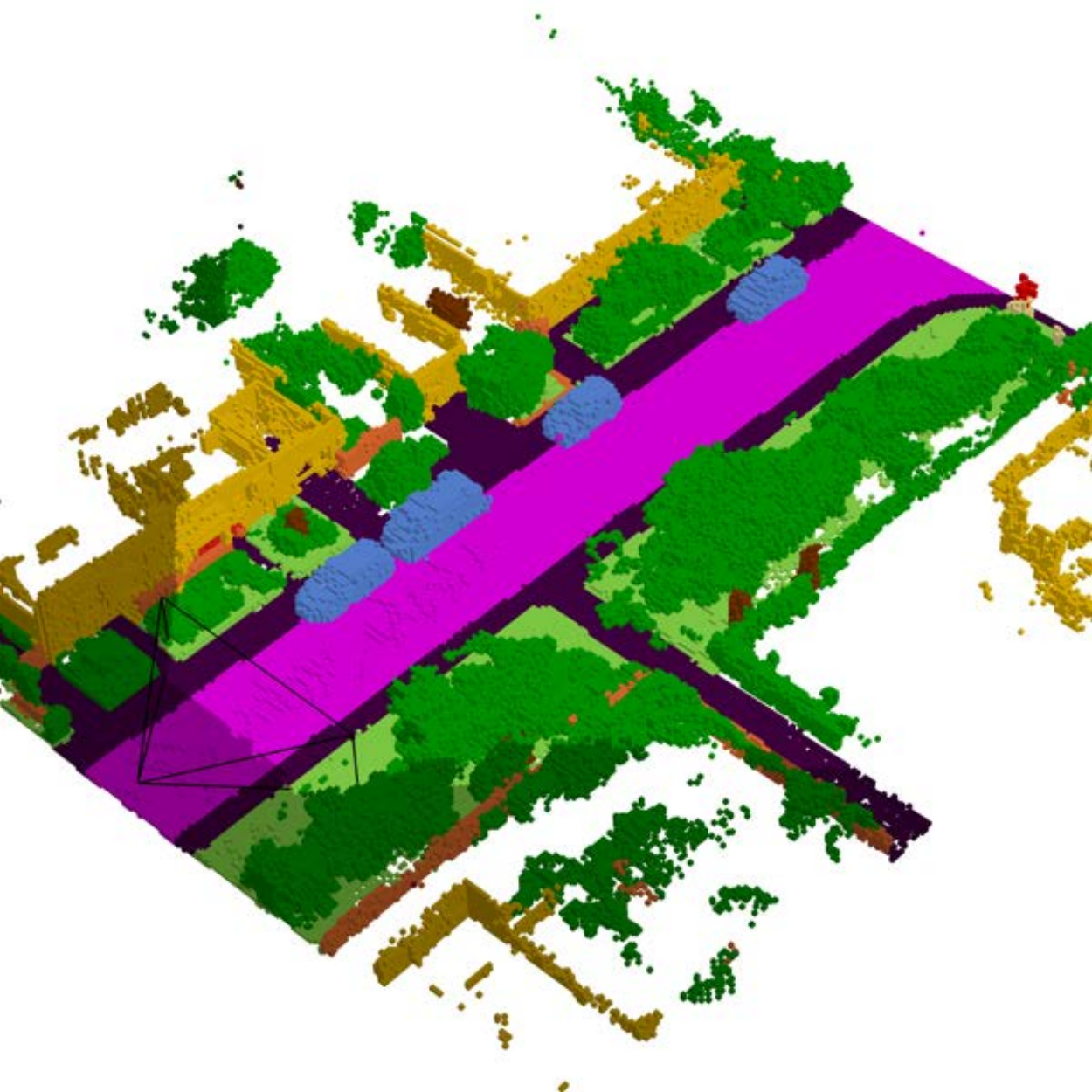} 
		\\[-0.1em]
		\includegraphics[width=.9\linewidth]{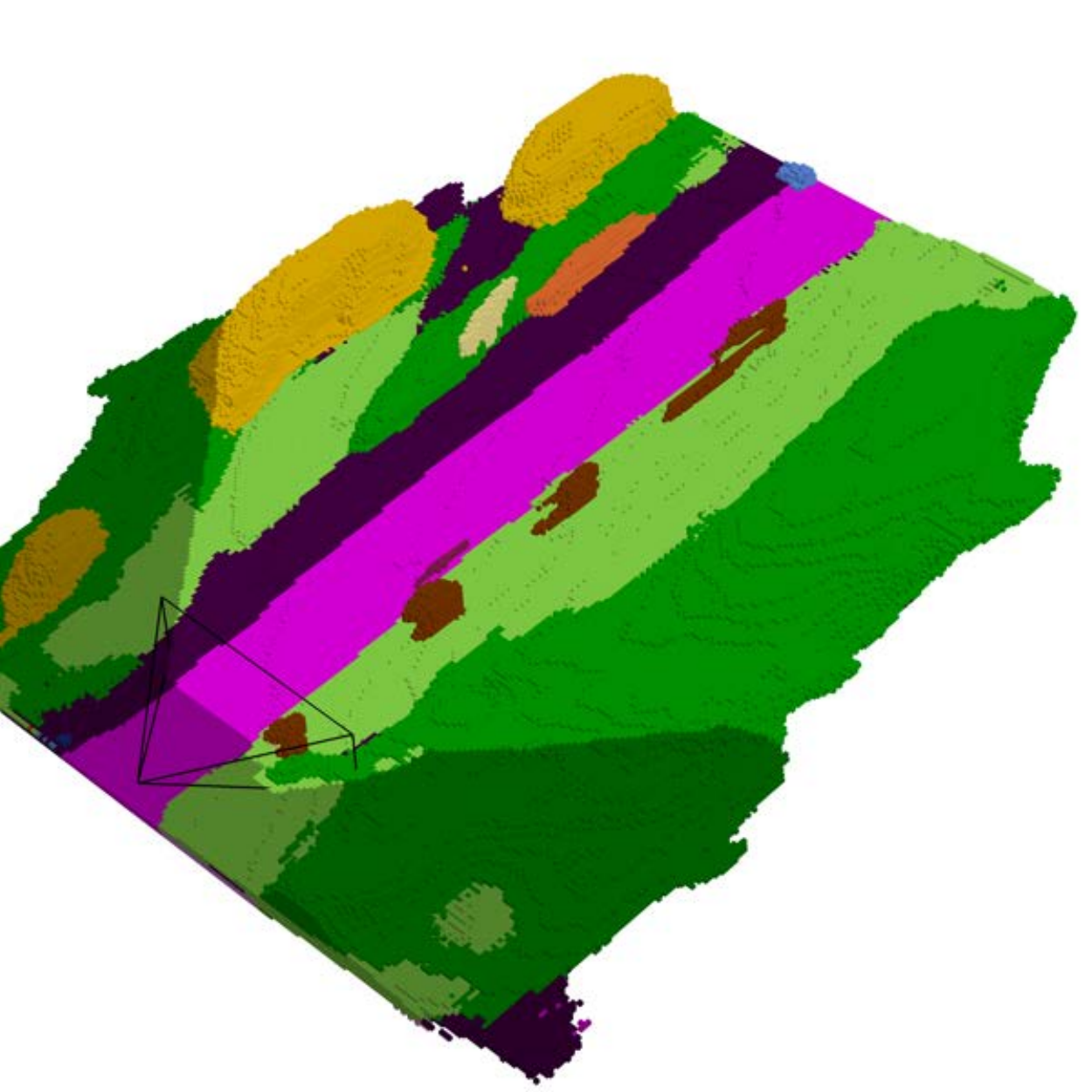} &  
		\includegraphics[width=.9\linewidth]{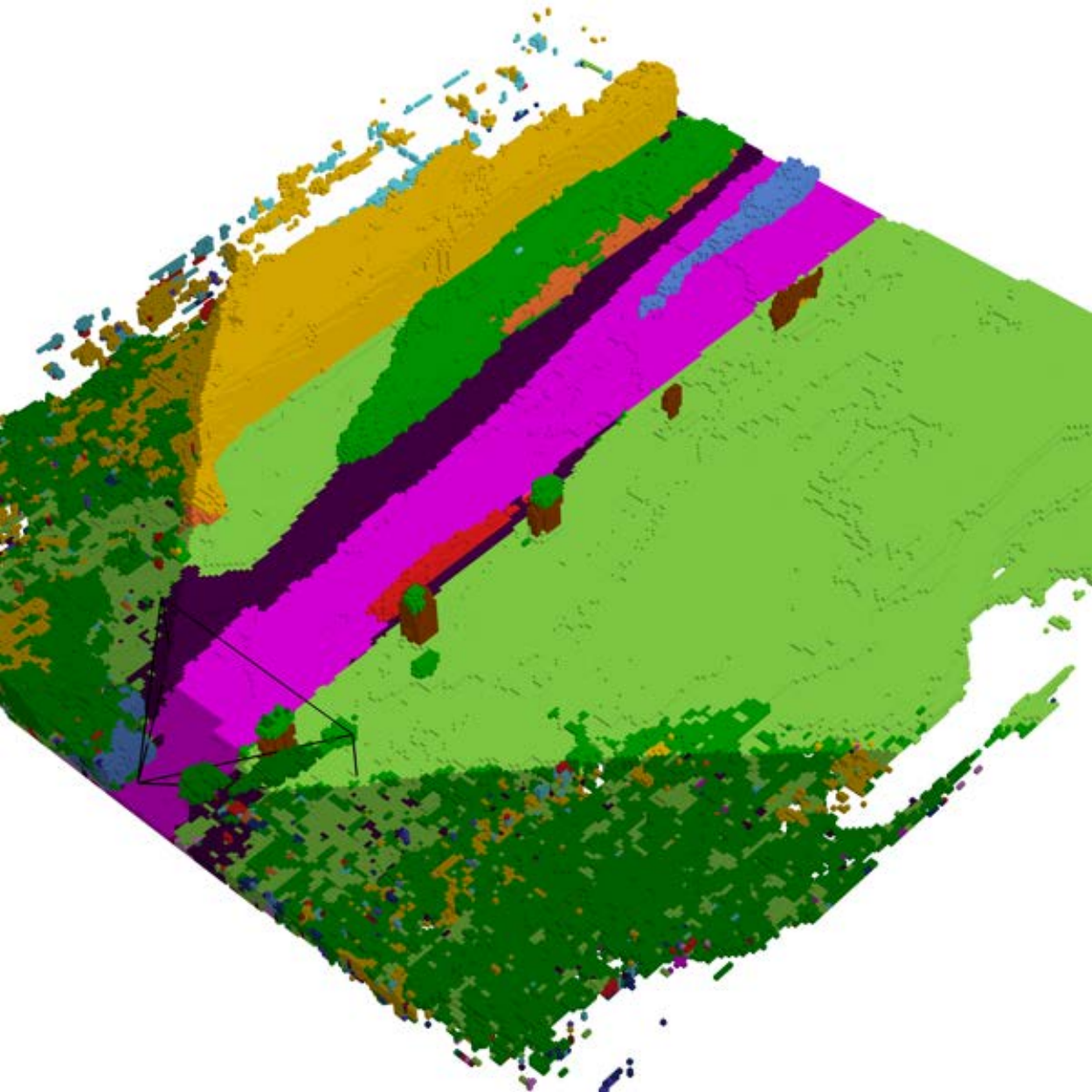} & 
		\includegraphics[width=.9\linewidth]{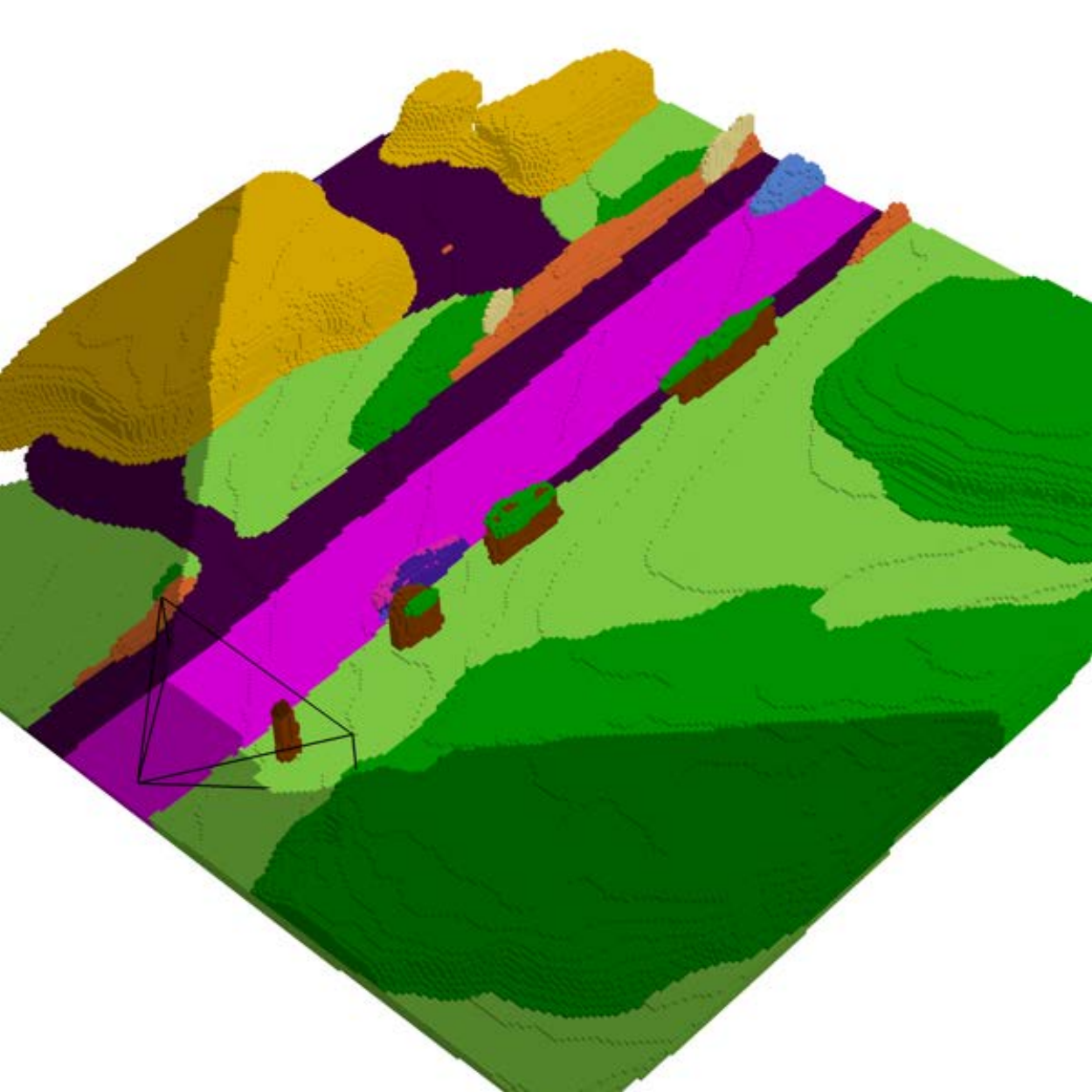} & 
		\includegraphics[width=.9\linewidth]{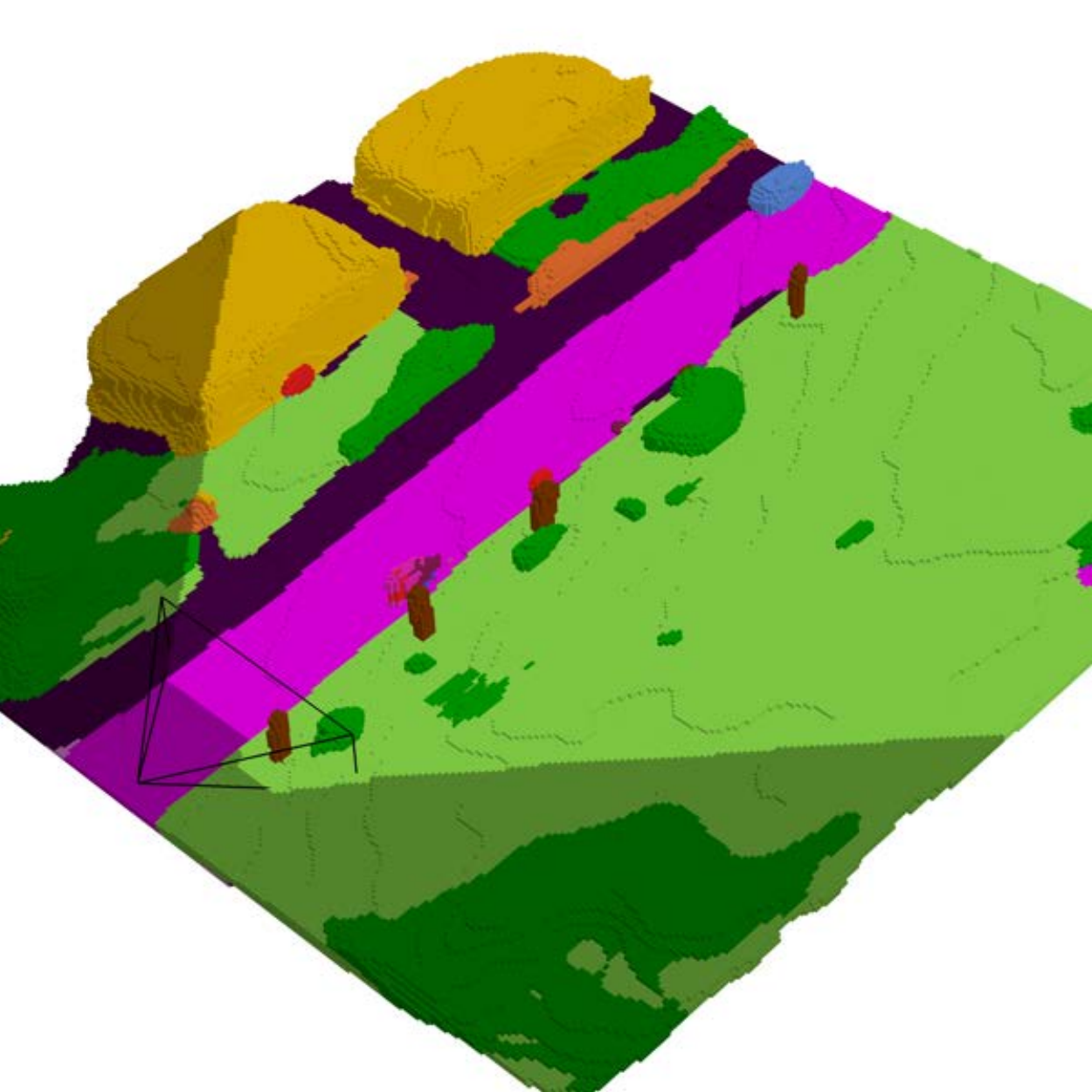} & 
		\includegraphics[width=.9\linewidth]{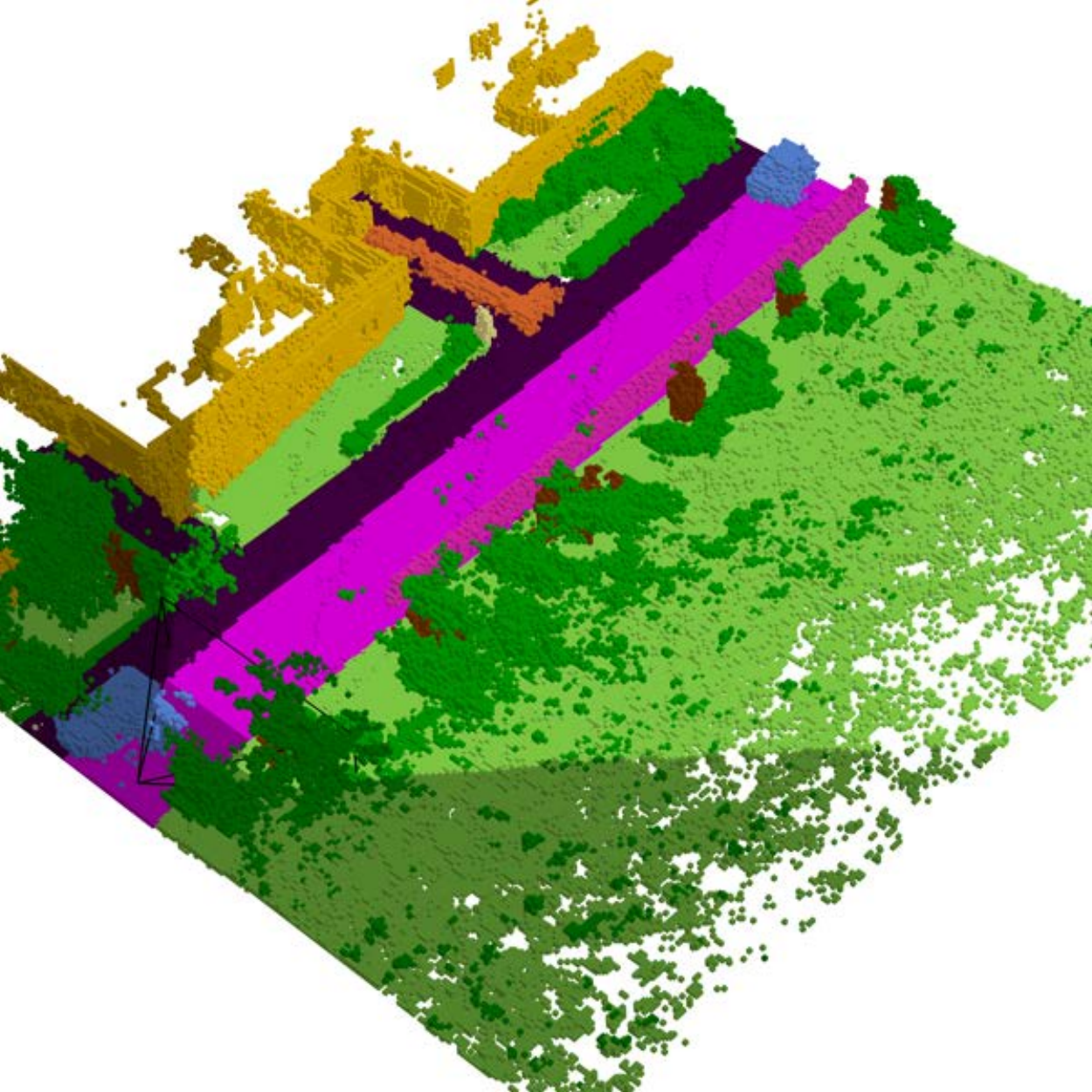} 
		\\[-0.1em]
	(a) MonoScene~\cite{MonoScene} & (b) VoxFormer~\cite{VoxFormer} & (c) OccFormer~\cite{OccFormer} & (d) CGFormer (ours) & (e) Ground Truth
	\end{tabular}
	\caption{More qualitative comparison results on the SemanticKITTI~\cite{SemanticKITTI} validation set.}
	\label{fig:AdditionQualitative}
	\vspace{-4mm}
\end{figure*} 

%% file: fig/MoreQualitativeResults2.tex
\begin{figure*}
	\centering
	\newcolumntype{P}[1]{>{\centering\arraybackslash}m{#1}}
	\renewcommand{\arraystretch}{1.0}
	\footnotesize
	\begin{tabular}{P{0.17\textwidth} P{0.18\textwidth} P{0.18\textwidth} P{0.19\textwidth} P{0.17\textwidth}}	
		\includegraphics[width=.9\linewidth]{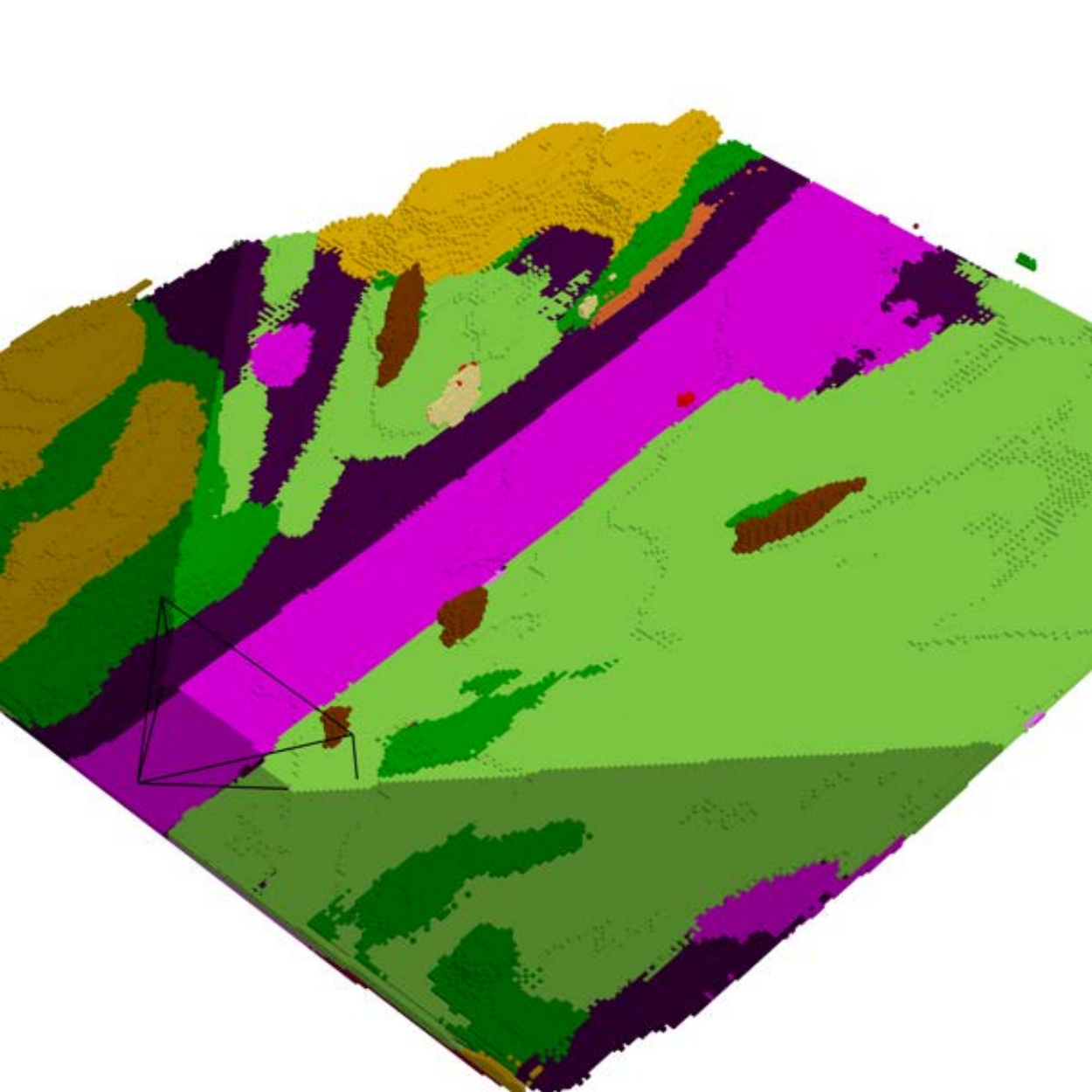} &  
		\includegraphics[width=.9\linewidth]{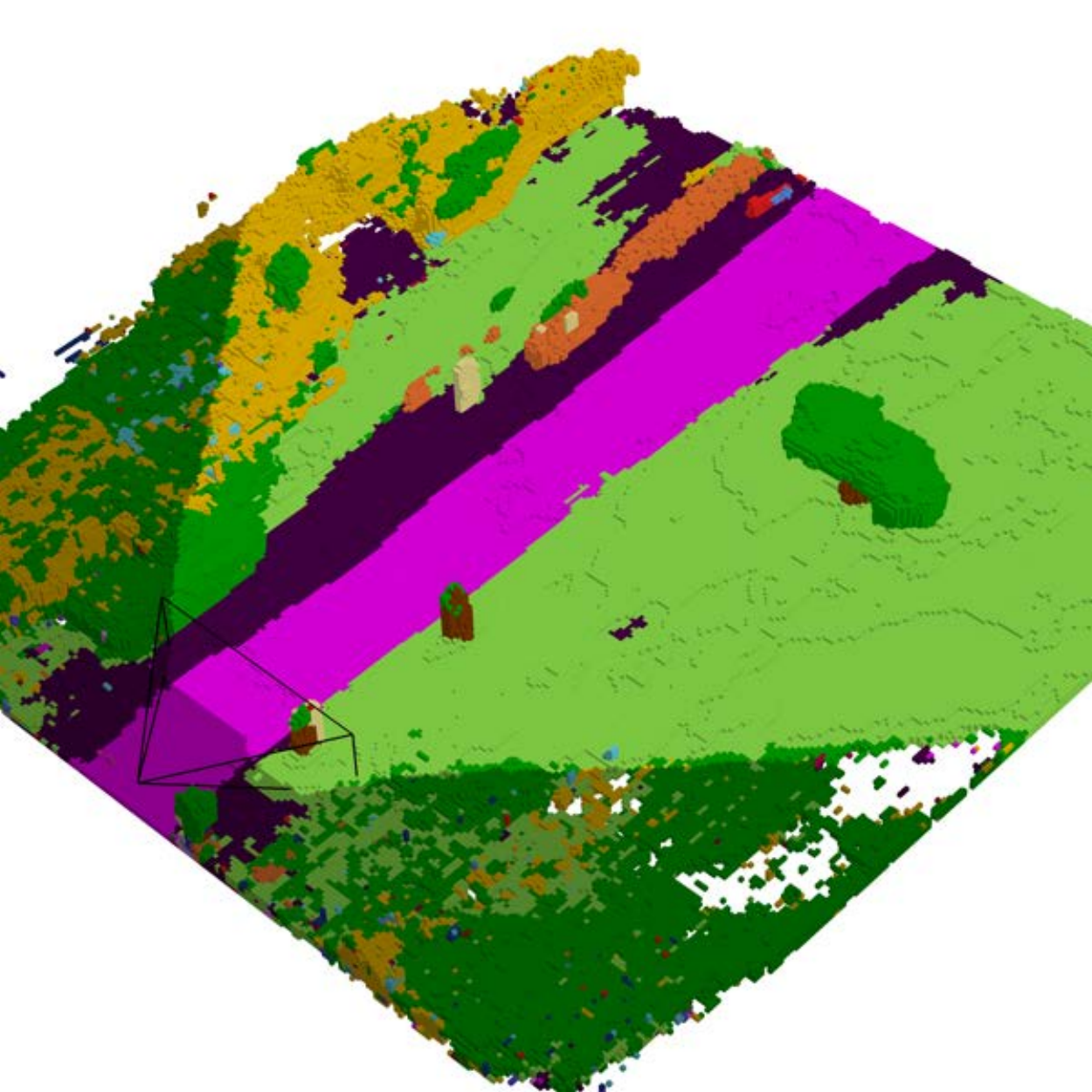} & 
		\includegraphics[width=.9\linewidth]{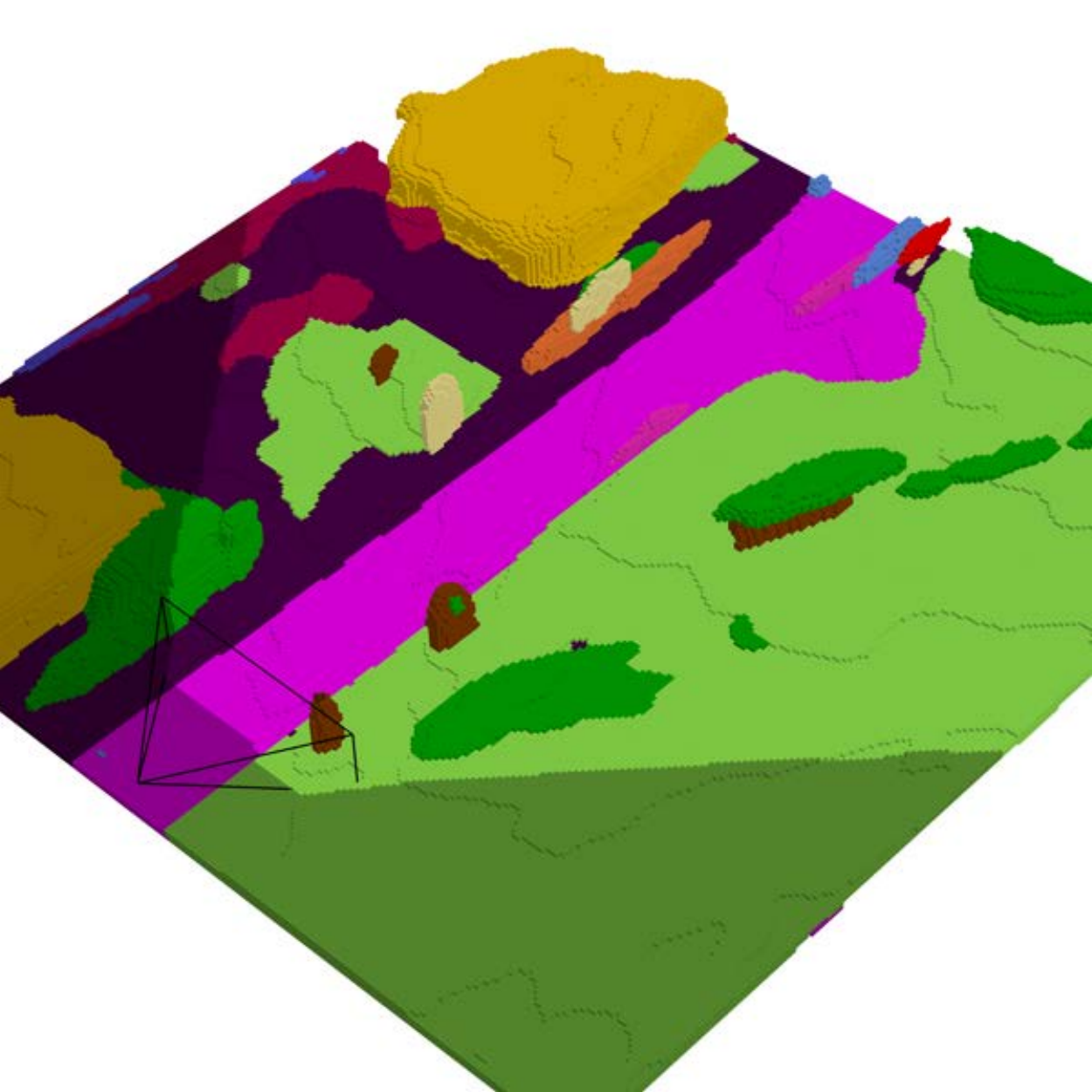} & 
		\includegraphics[width=.9\linewidth]{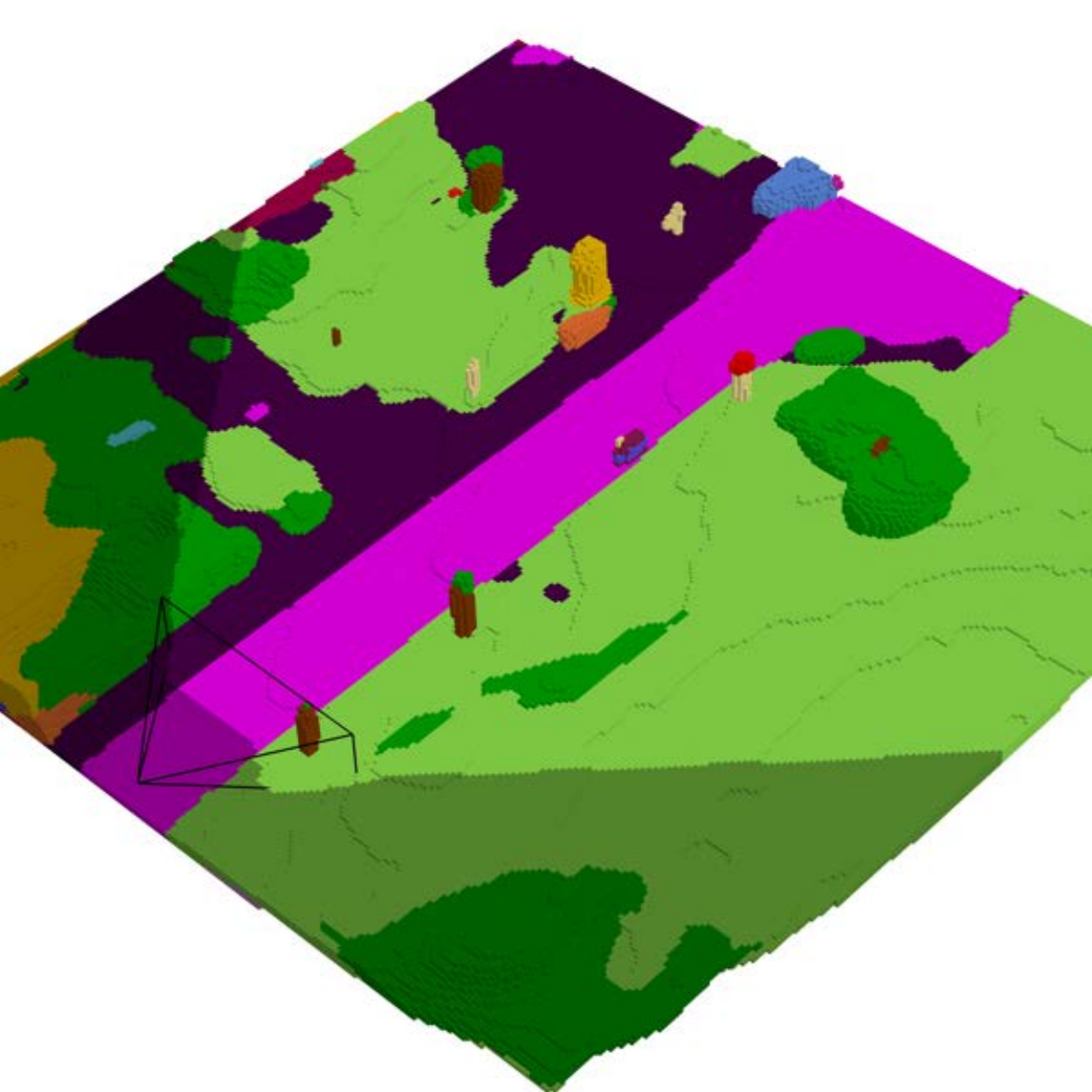} & 
		\includegraphics[width=.9\linewidth]{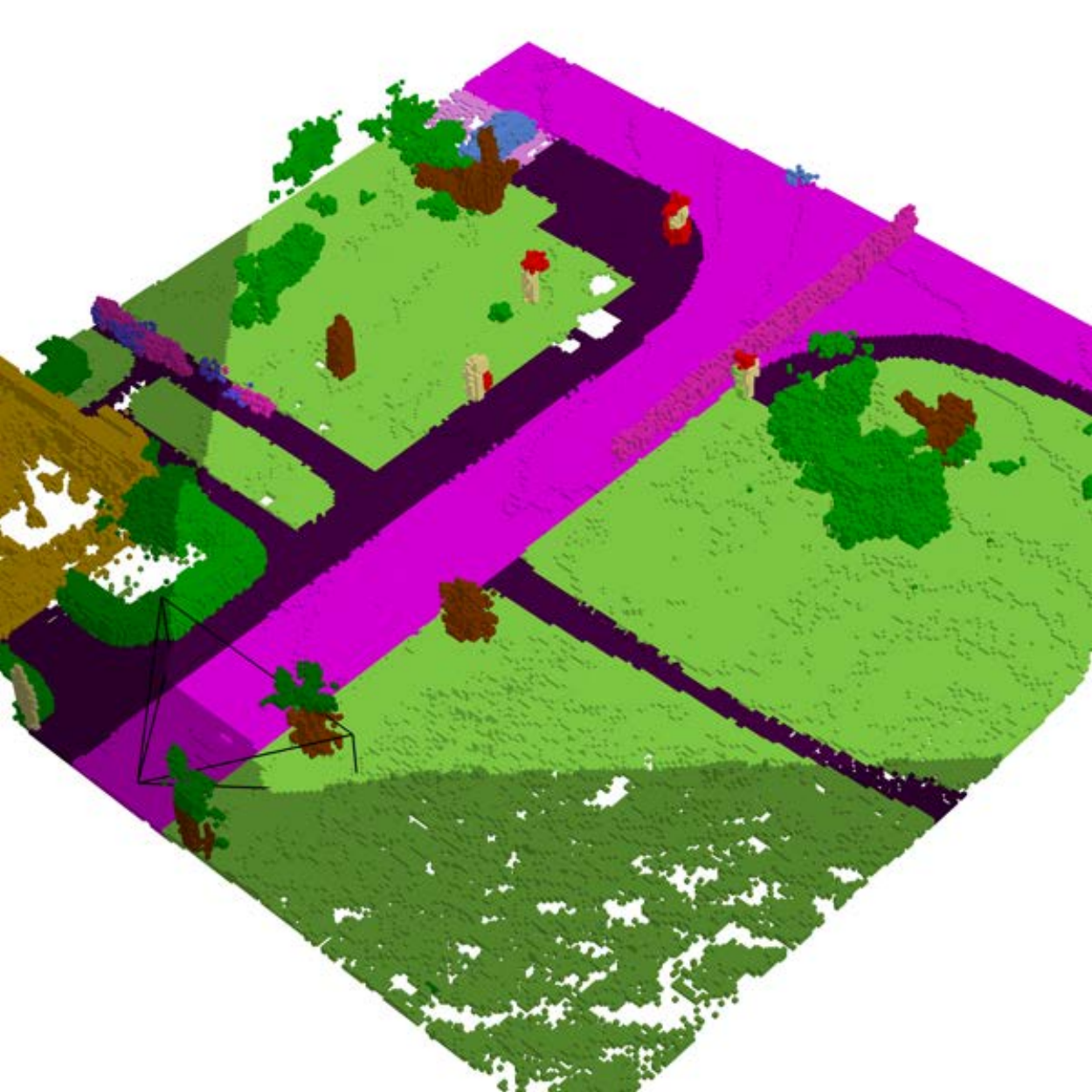} 
		\\[-0.1em]
		\includegraphics[width=.9\linewidth]{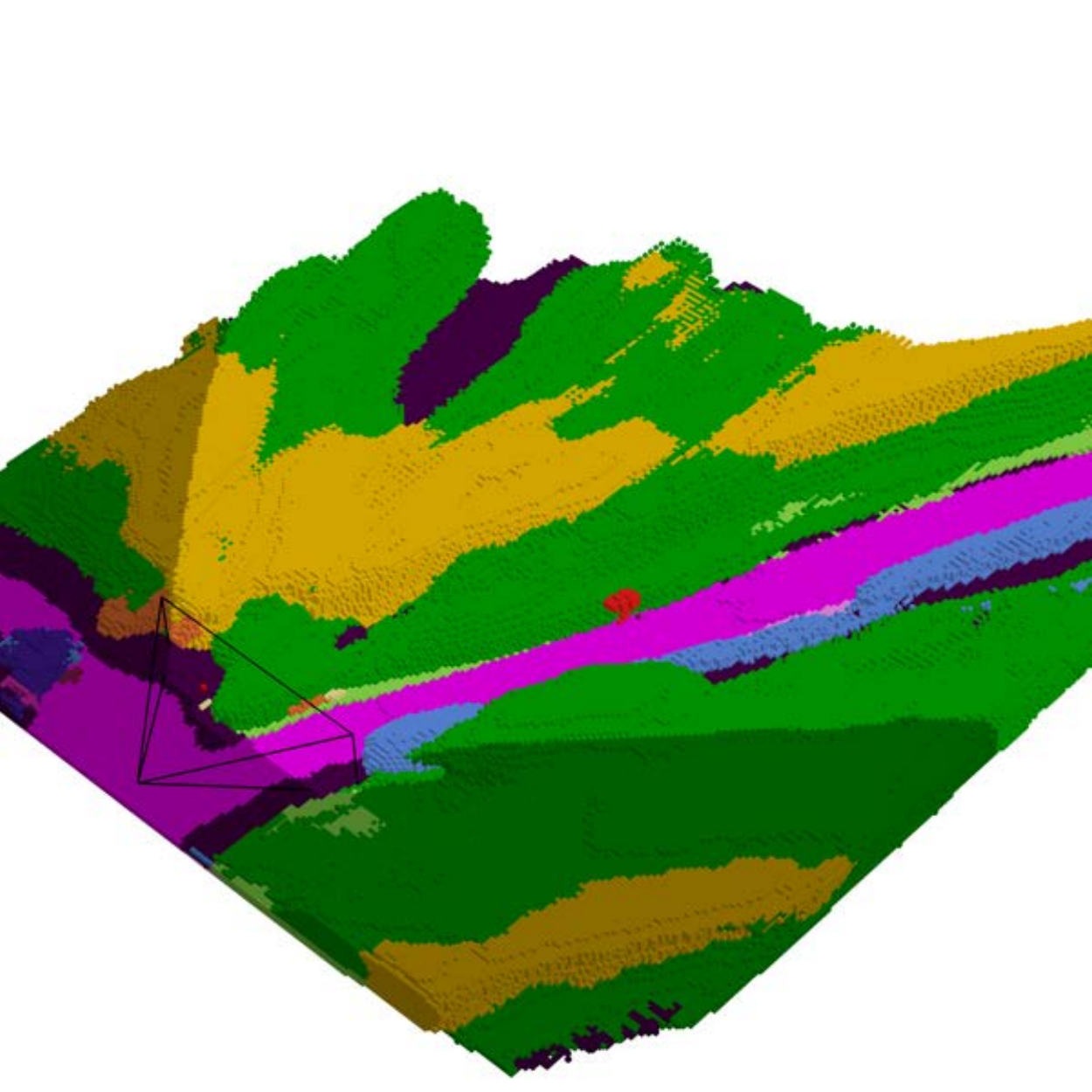} &  
		\includegraphics[width=.9\linewidth]{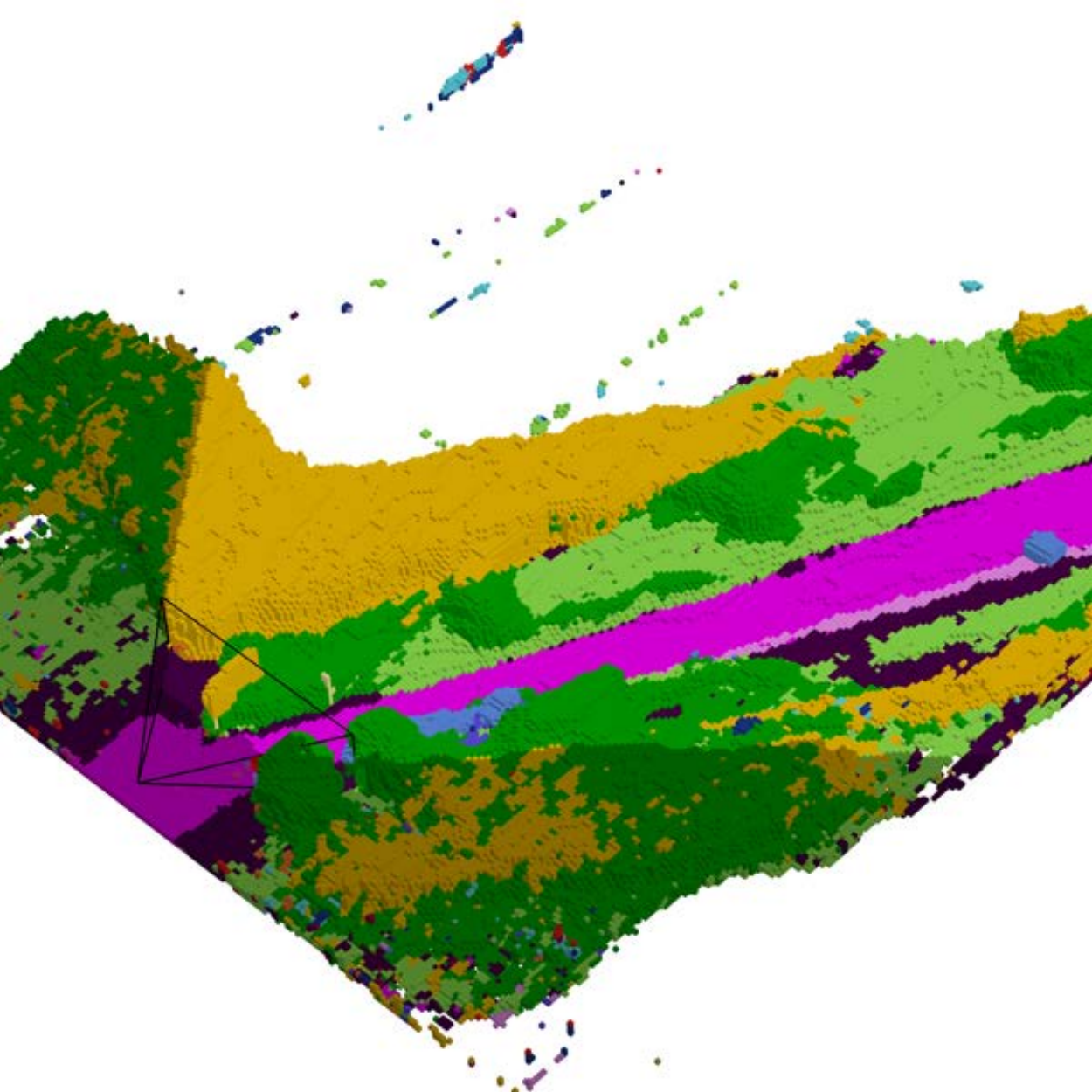} & 
		\includegraphics[width=.9\linewidth]{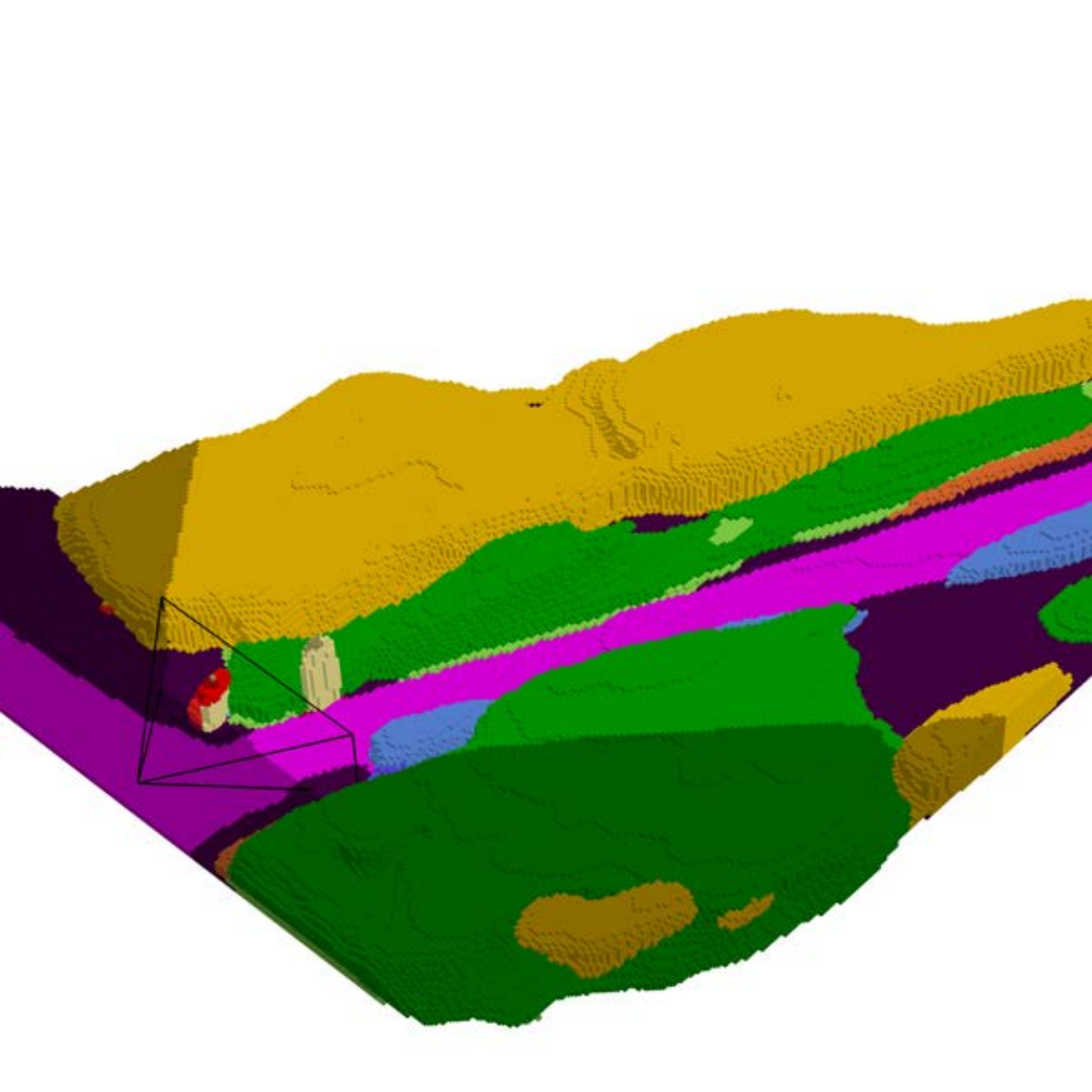} & 
		\includegraphics[width=.9\linewidth]{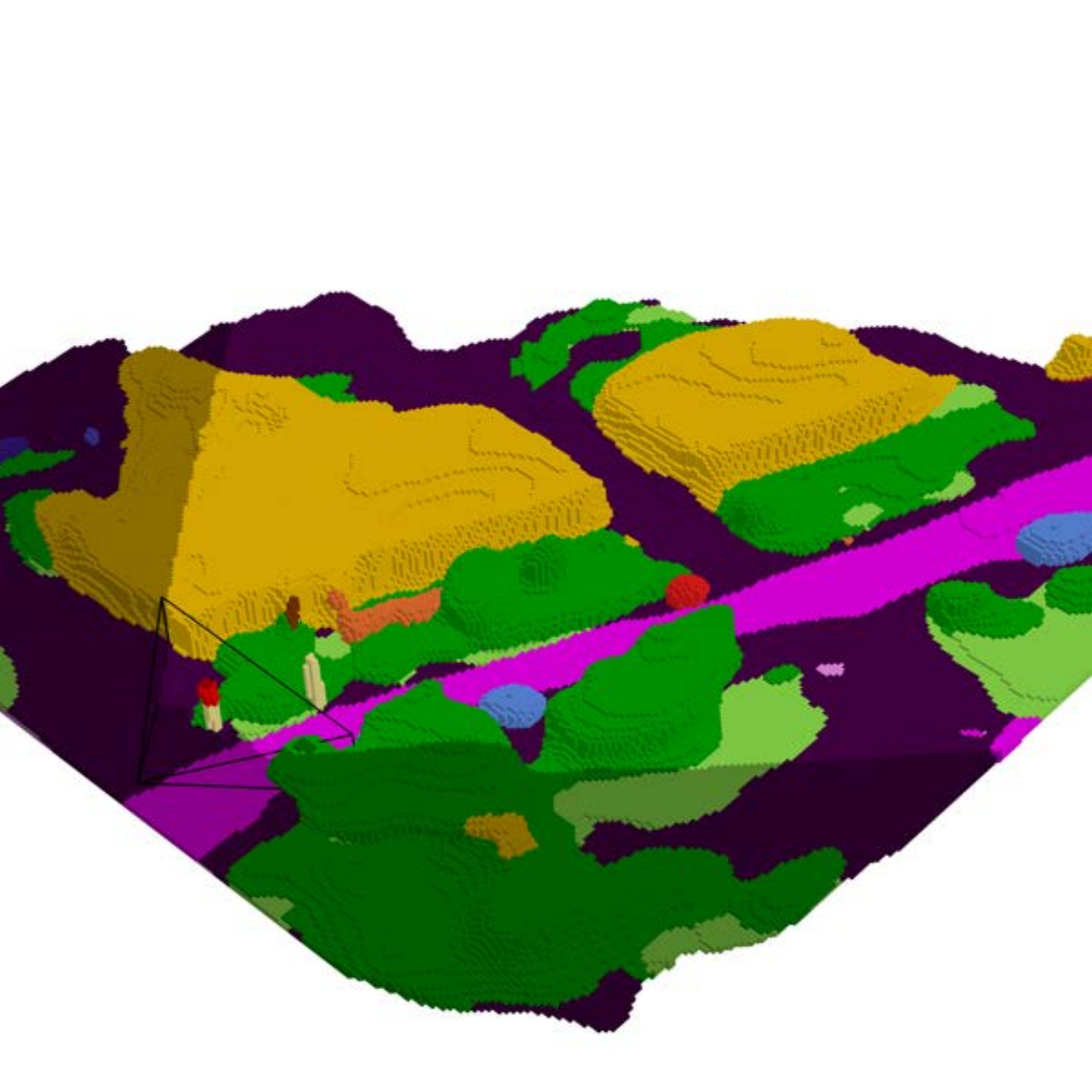} & 
		\includegraphics[width=.9\linewidth]{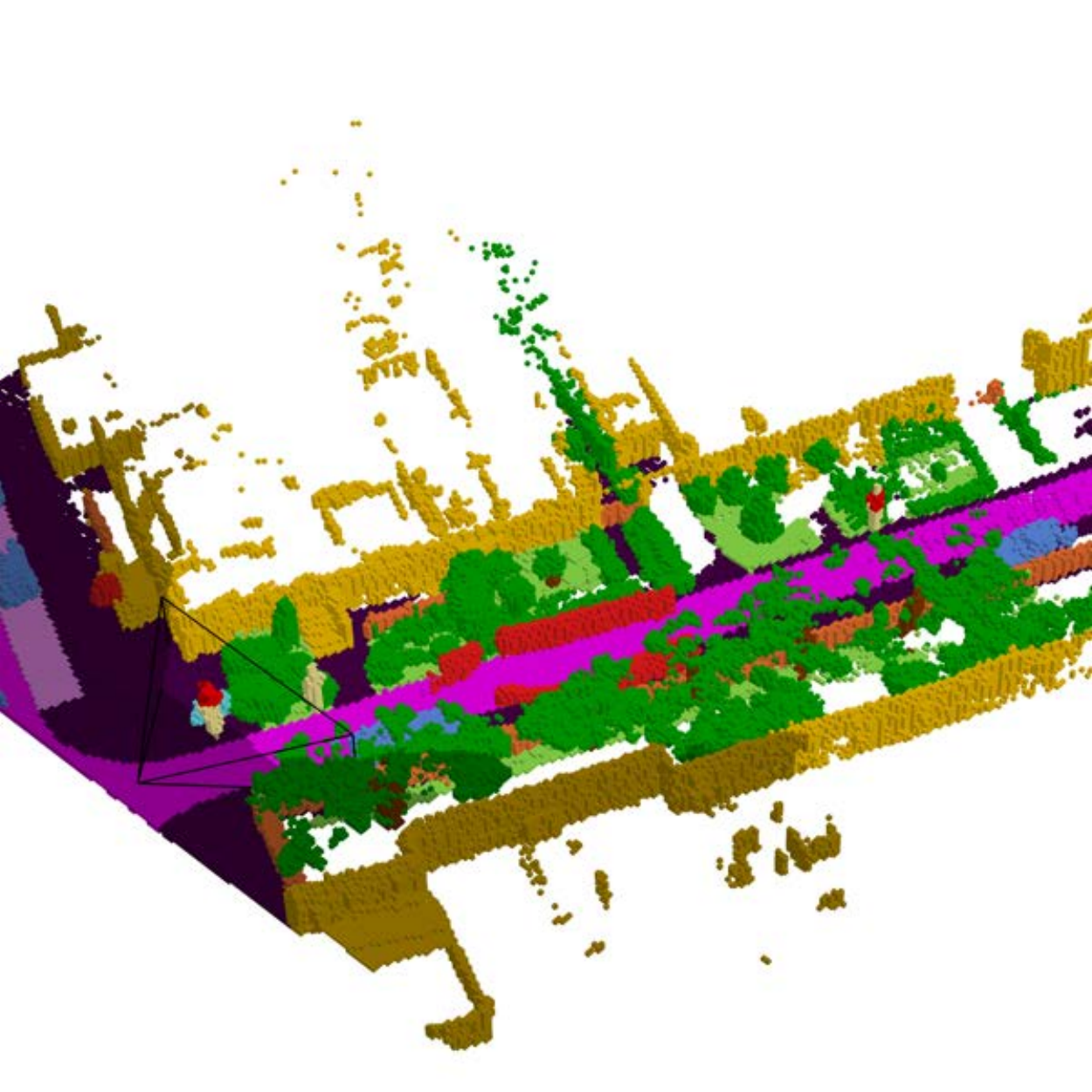} 
		\\[-0.1em]
		\includegraphics[width=.9\linewidth]{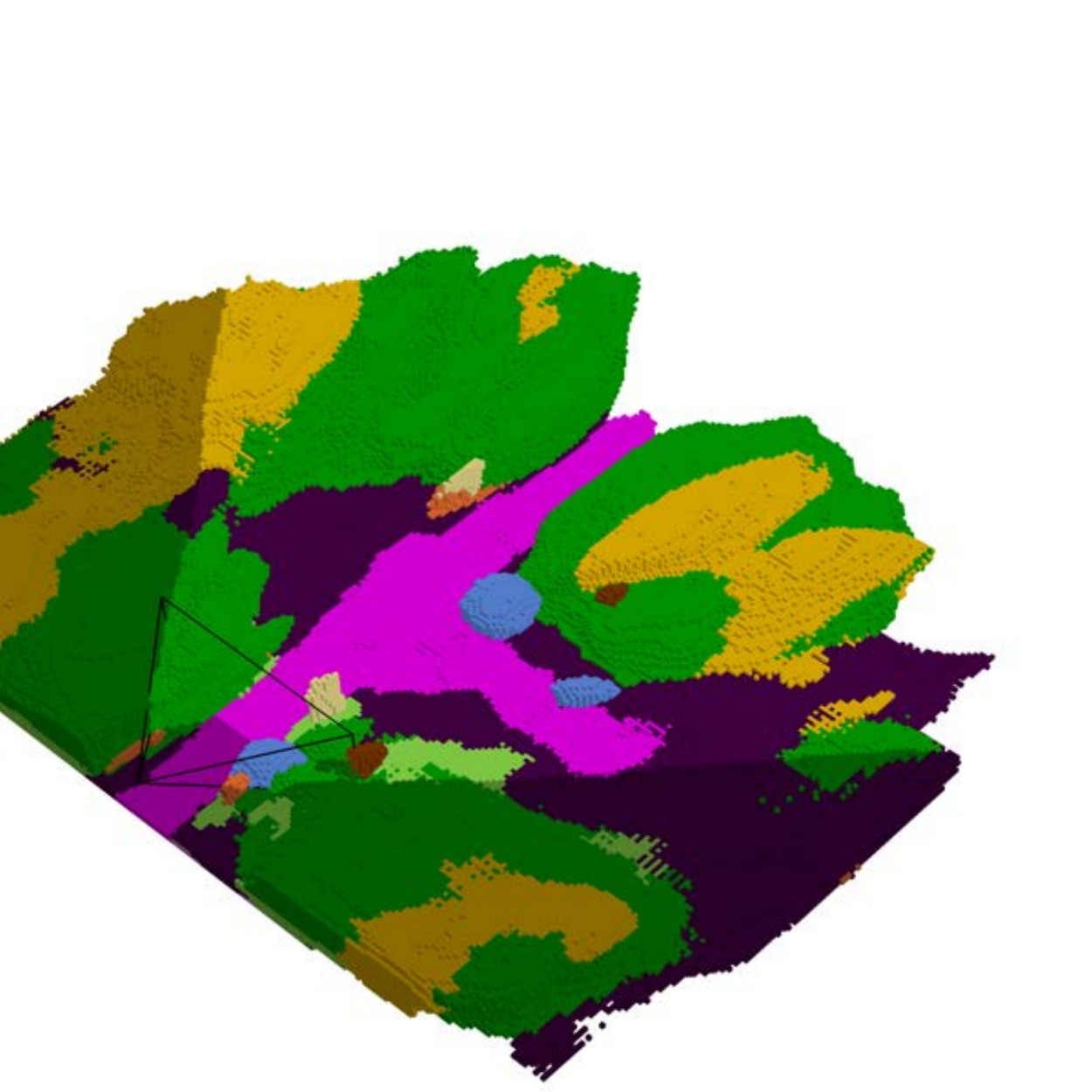} &  
		\includegraphics[width=.9\linewidth]{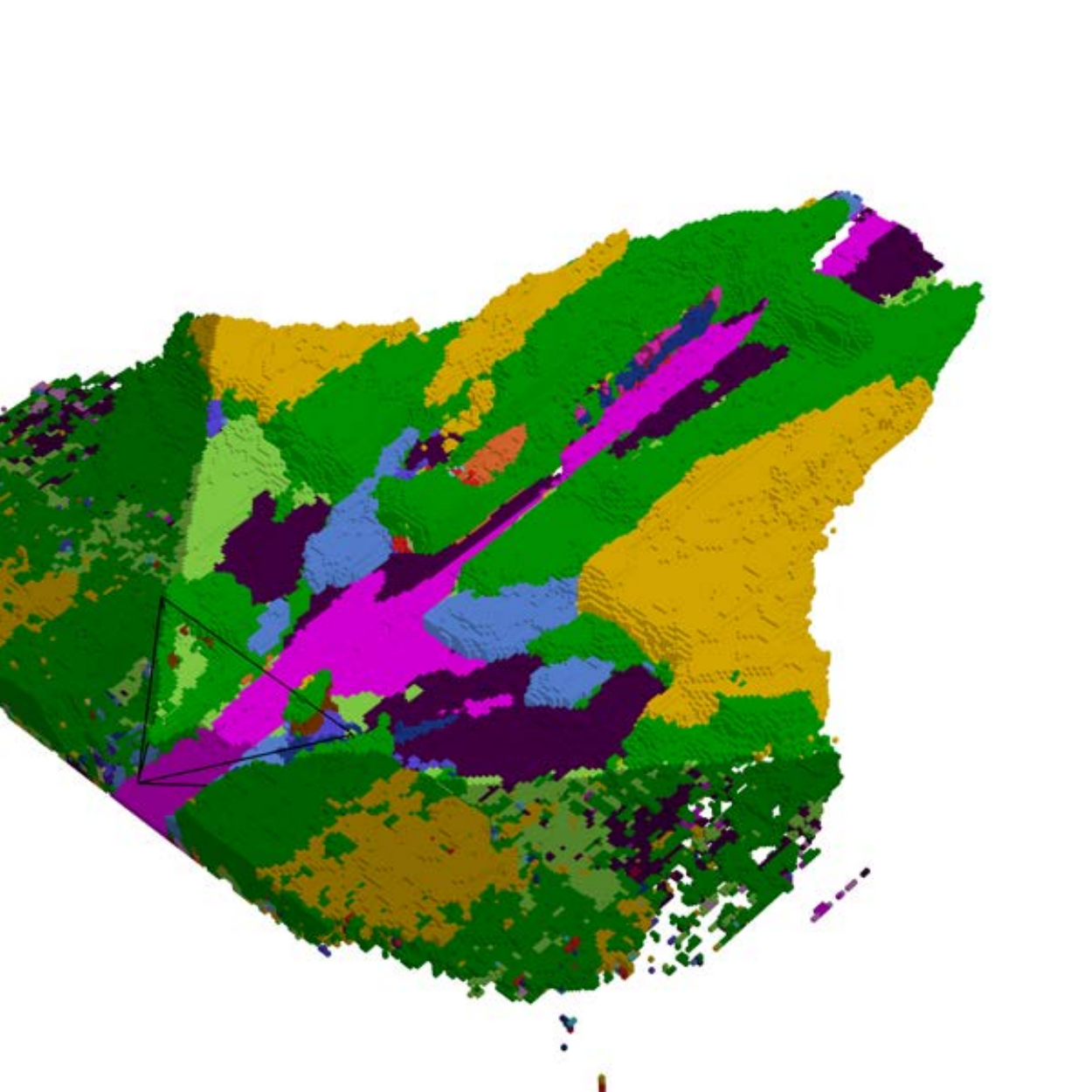} & 
		\includegraphics[width=.9\linewidth]{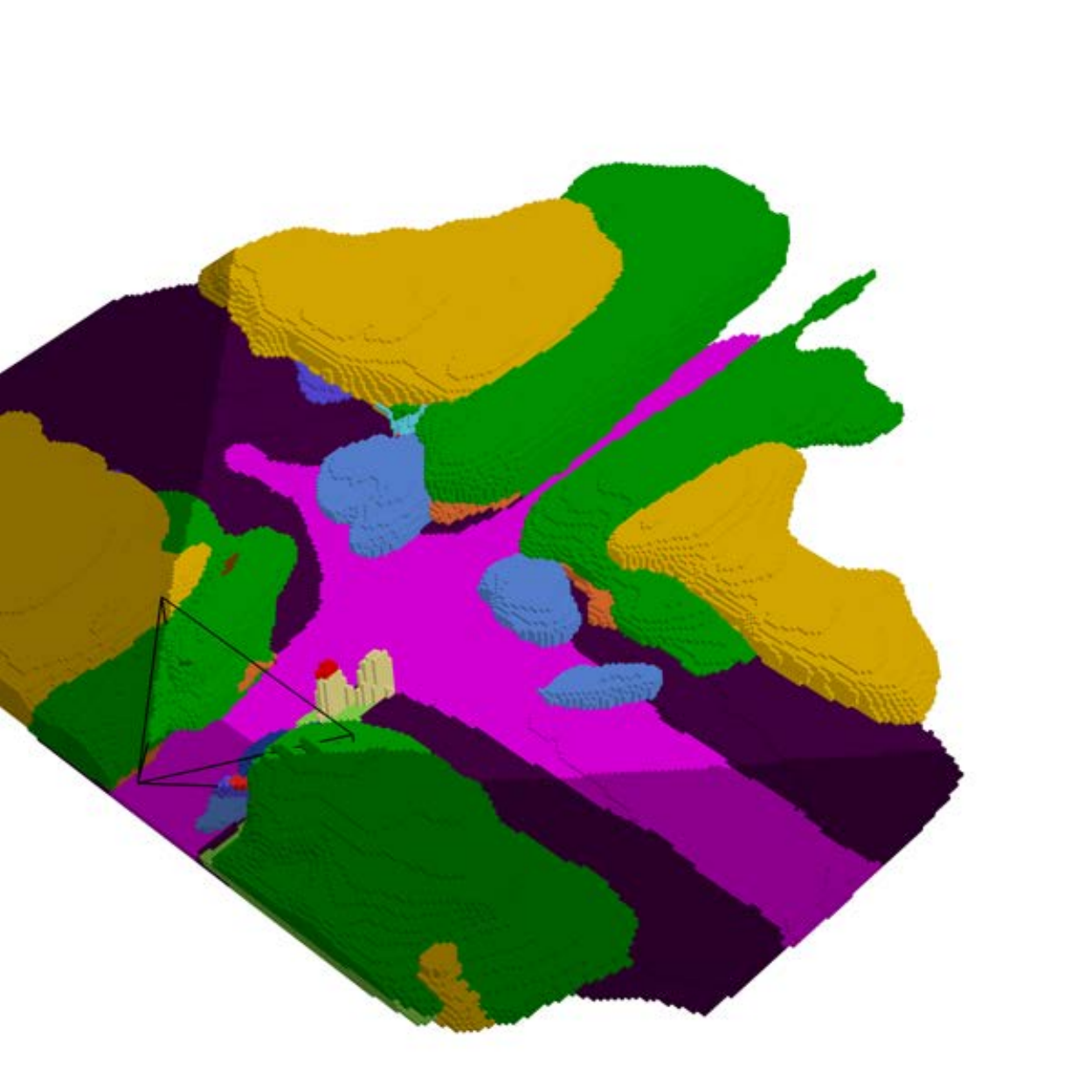} & 
		\includegraphics[width=.9\linewidth]{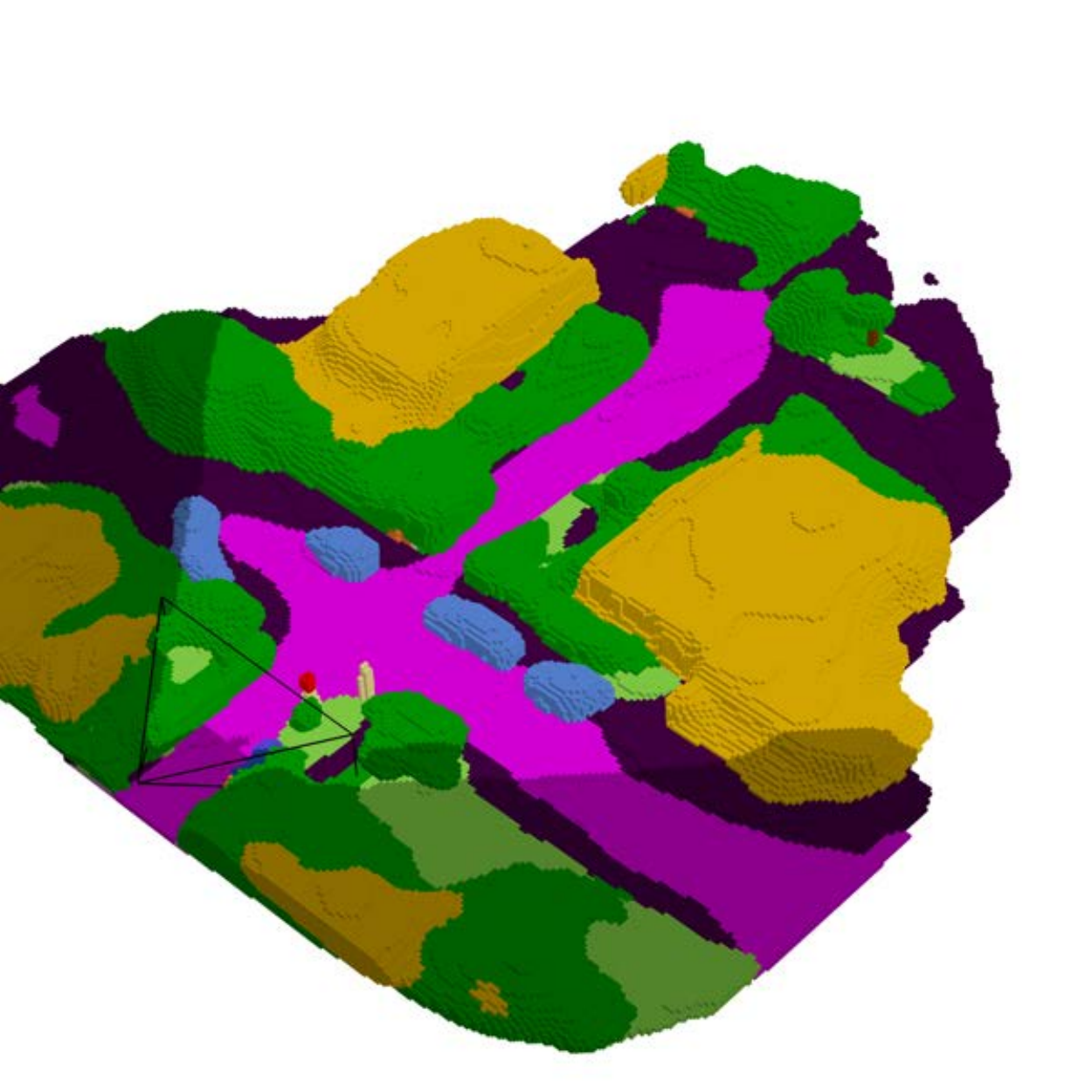} & 
		\includegraphics[width=.9\linewidth]{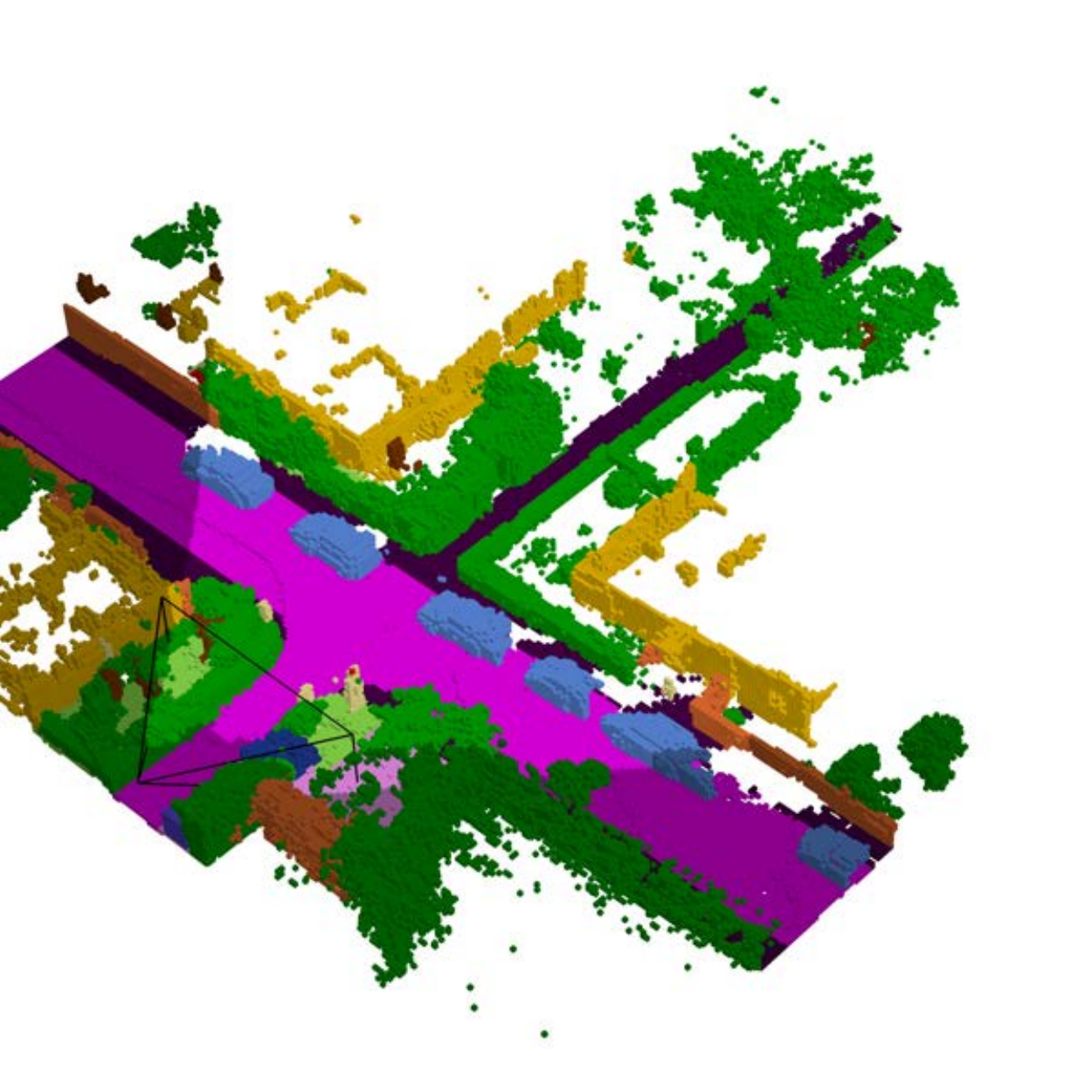} 
		\\[-0.1em]
		\includegraphics[width=.9\linewidth]{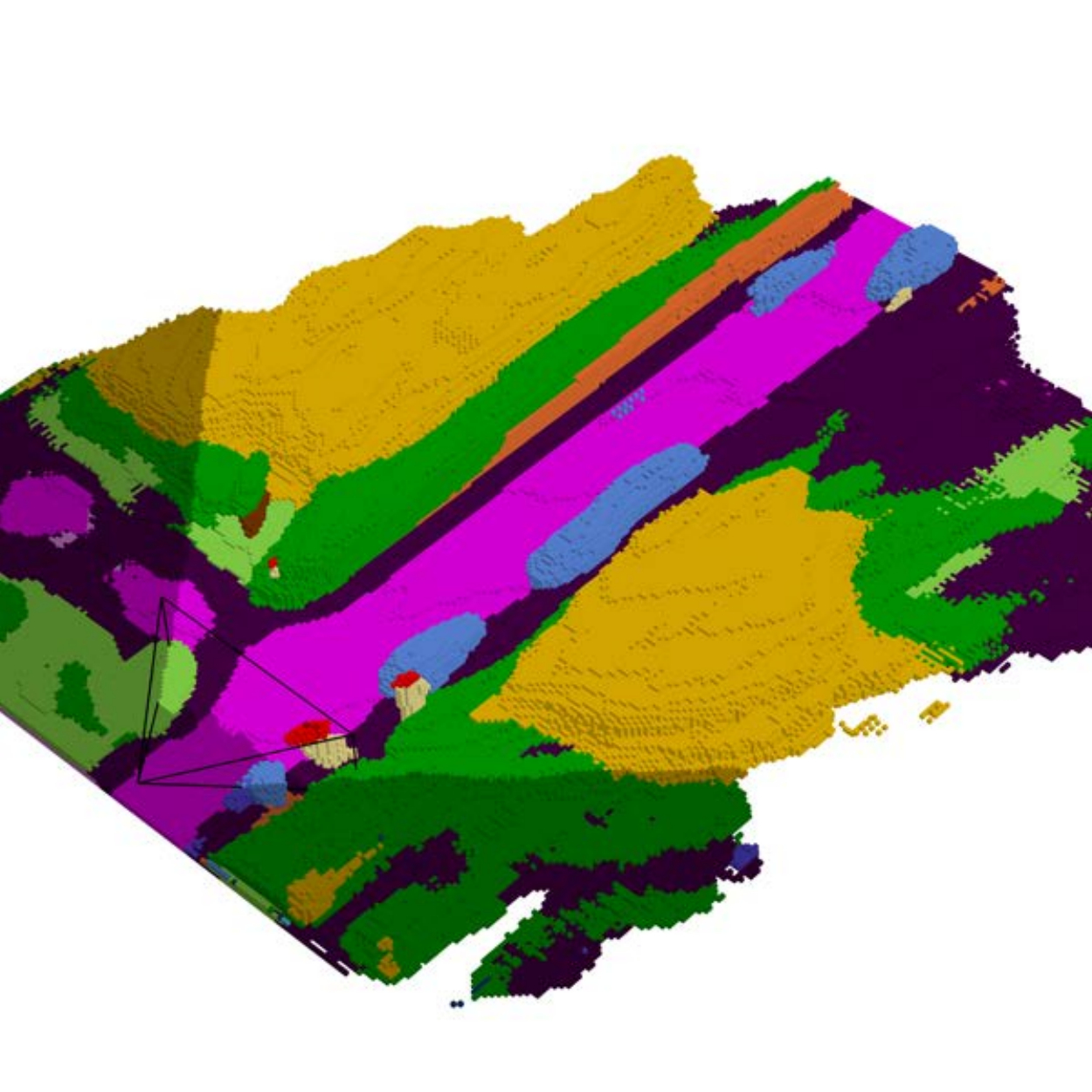} &  
		\includegraphics[width=.9\linewidth]{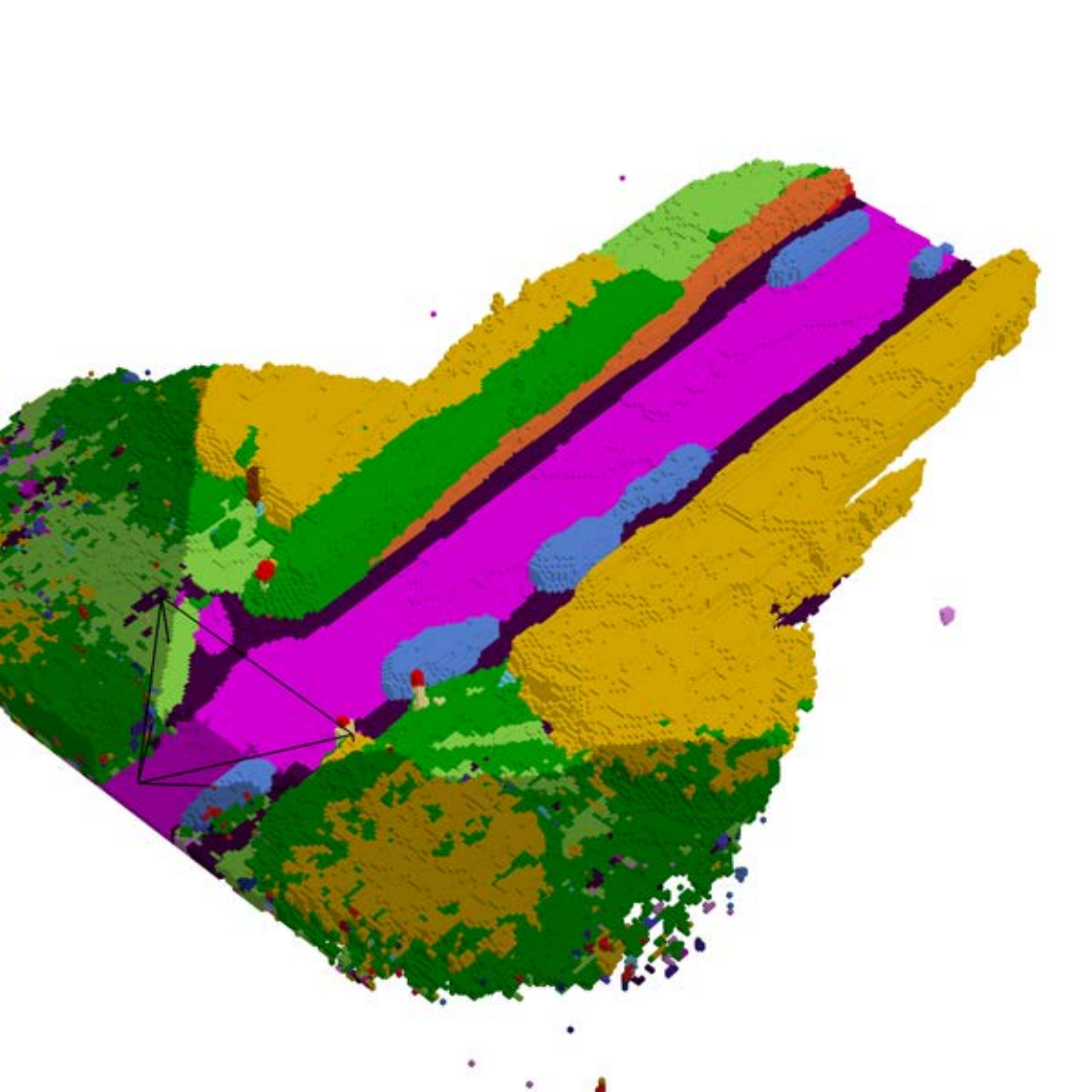} & 
		\includegraphics[width=.9\linewidth]{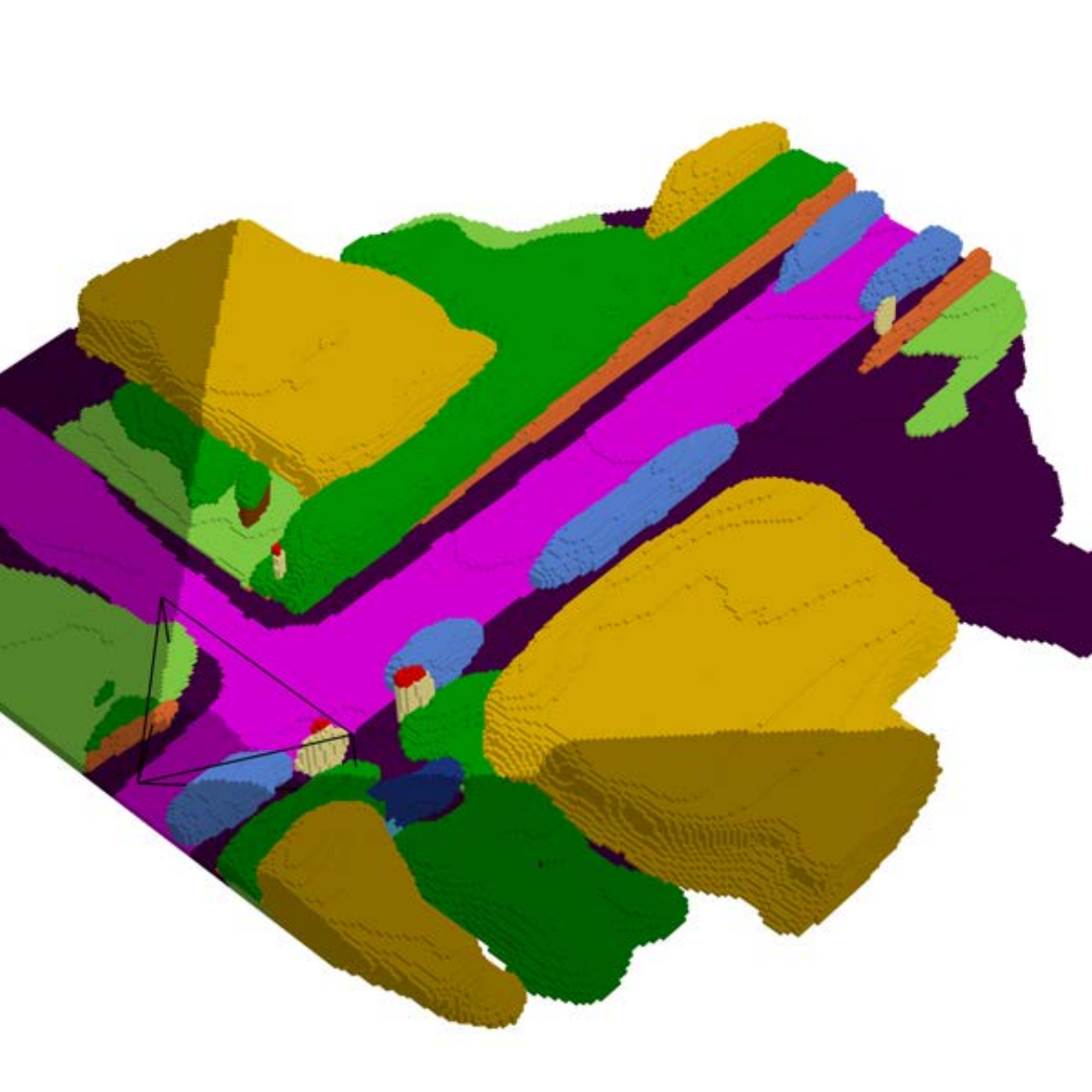} & 
		\includegraphics[width=.9\linewidth]{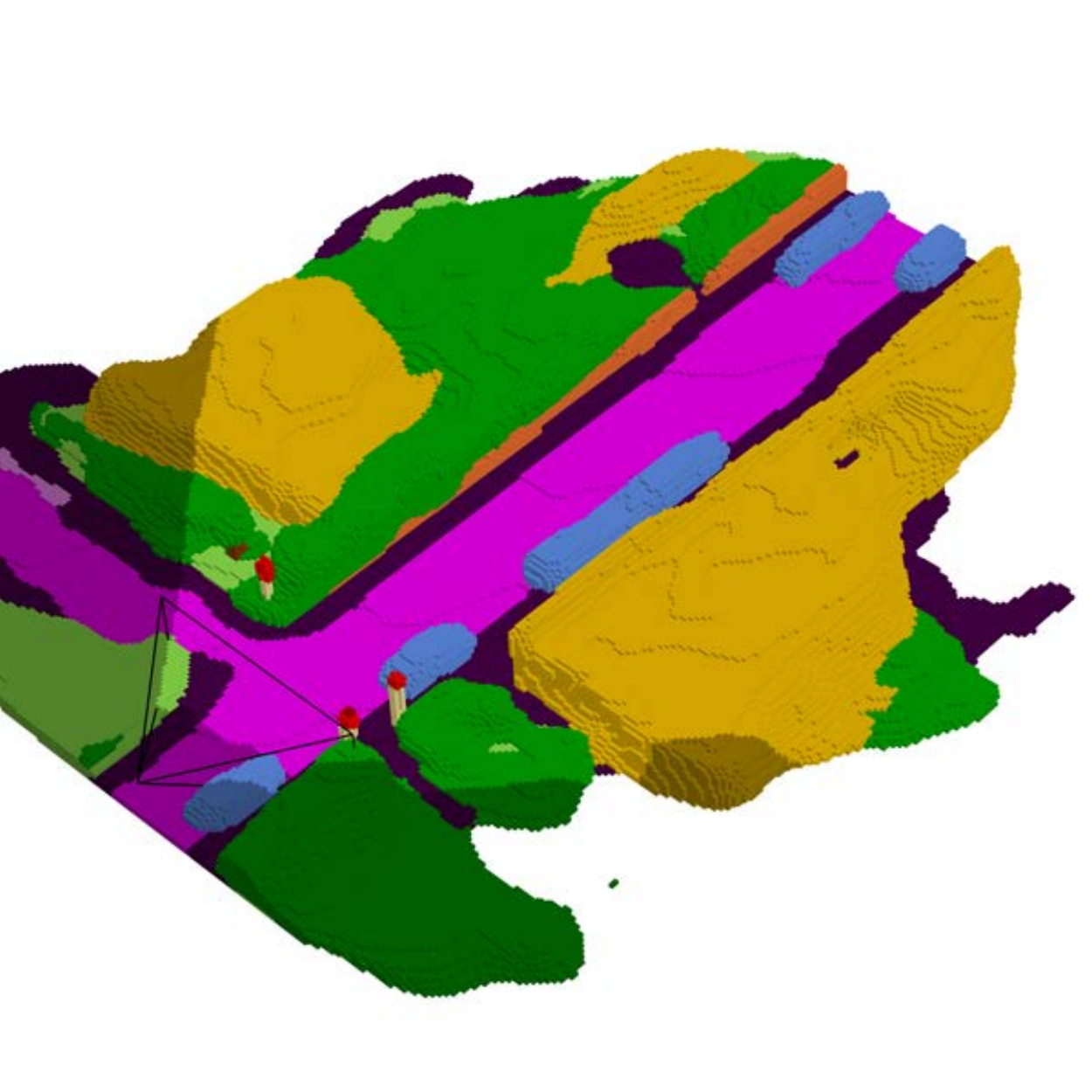} & 
		\includegraphics[width=.9\linewidth]{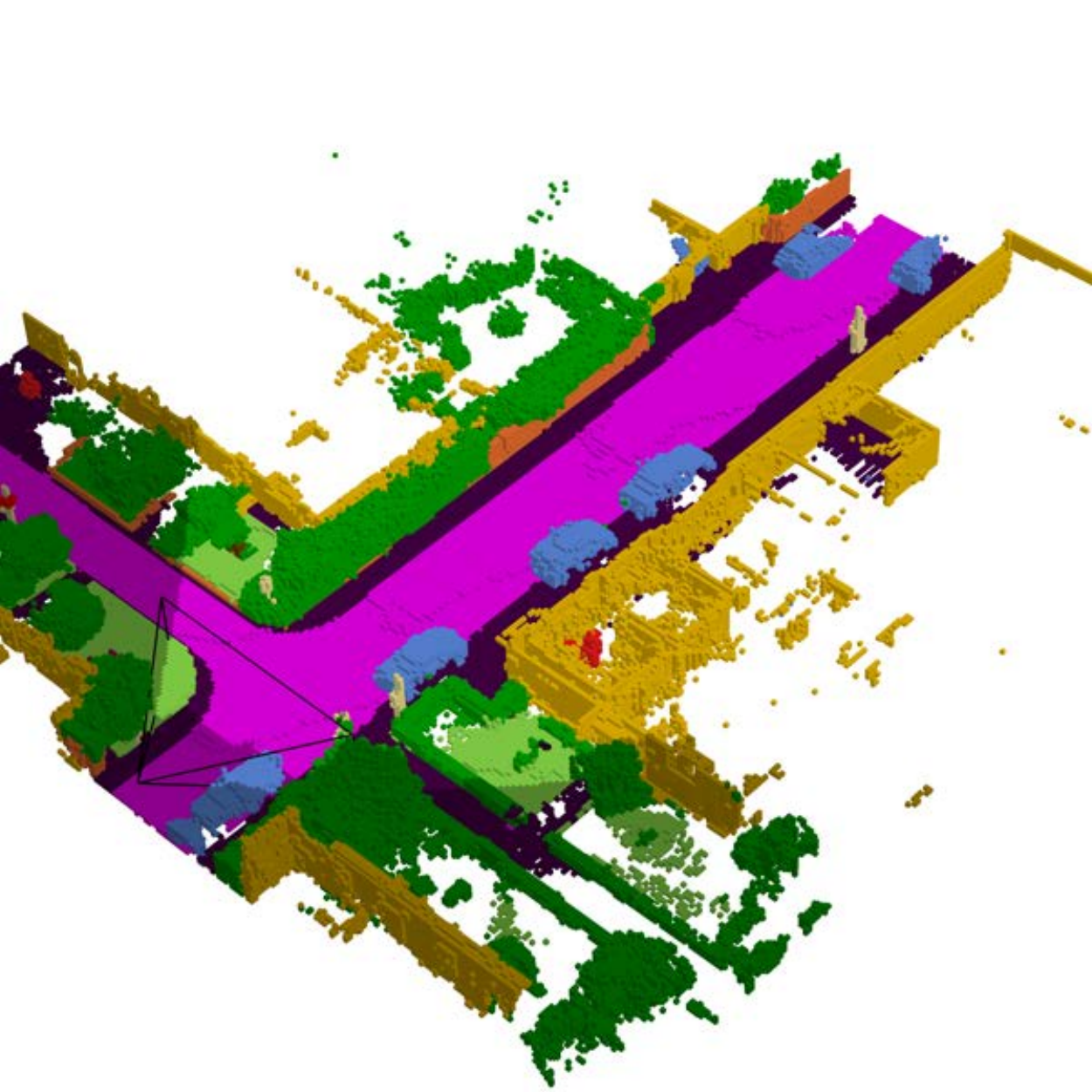} 
		\\[-0.1em]
		\includegraphics[width=.9\linewidth]{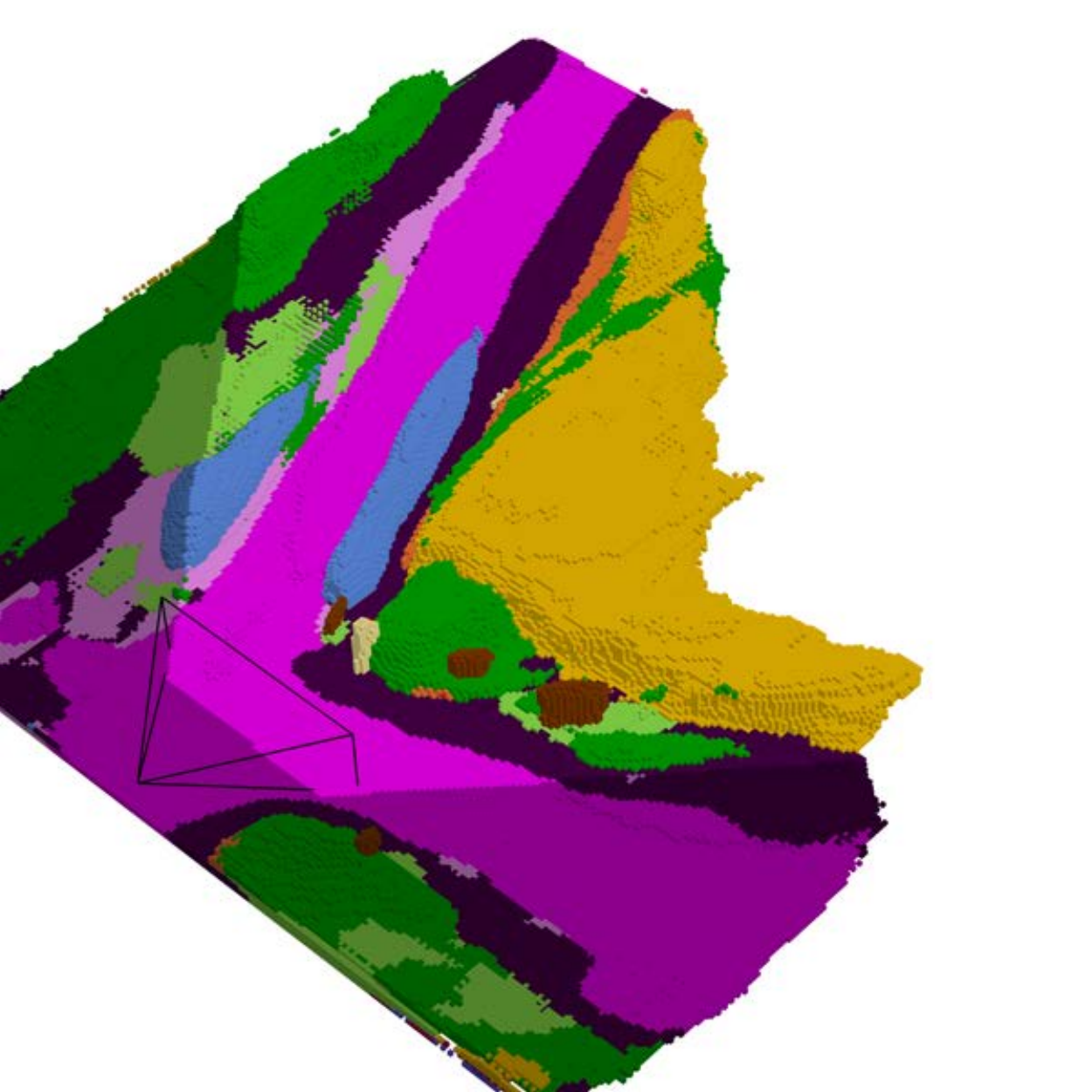} &  
		\includegraphics[width=.9\linewidth]{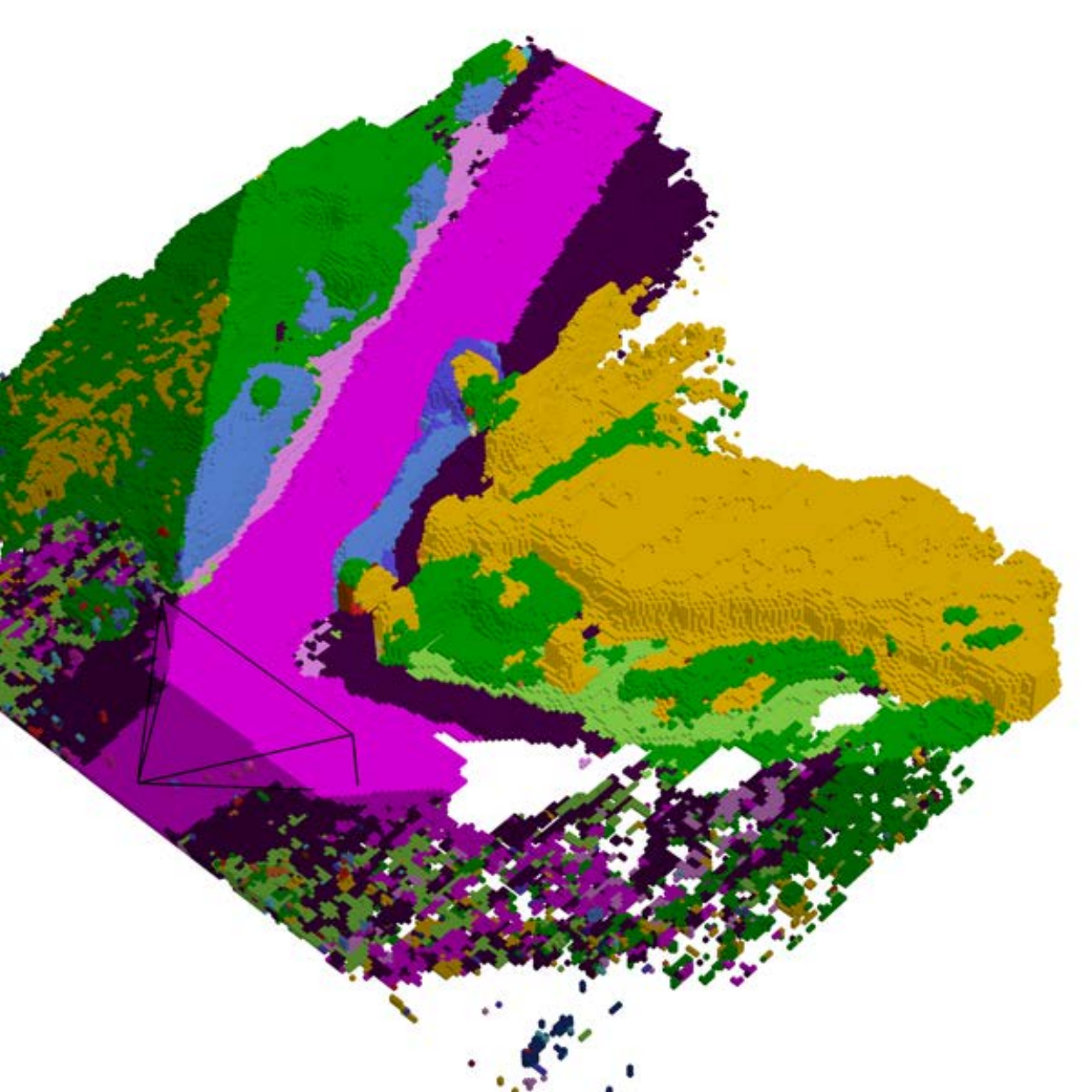} & 
		\includegraphics[width=.9\linewidth]{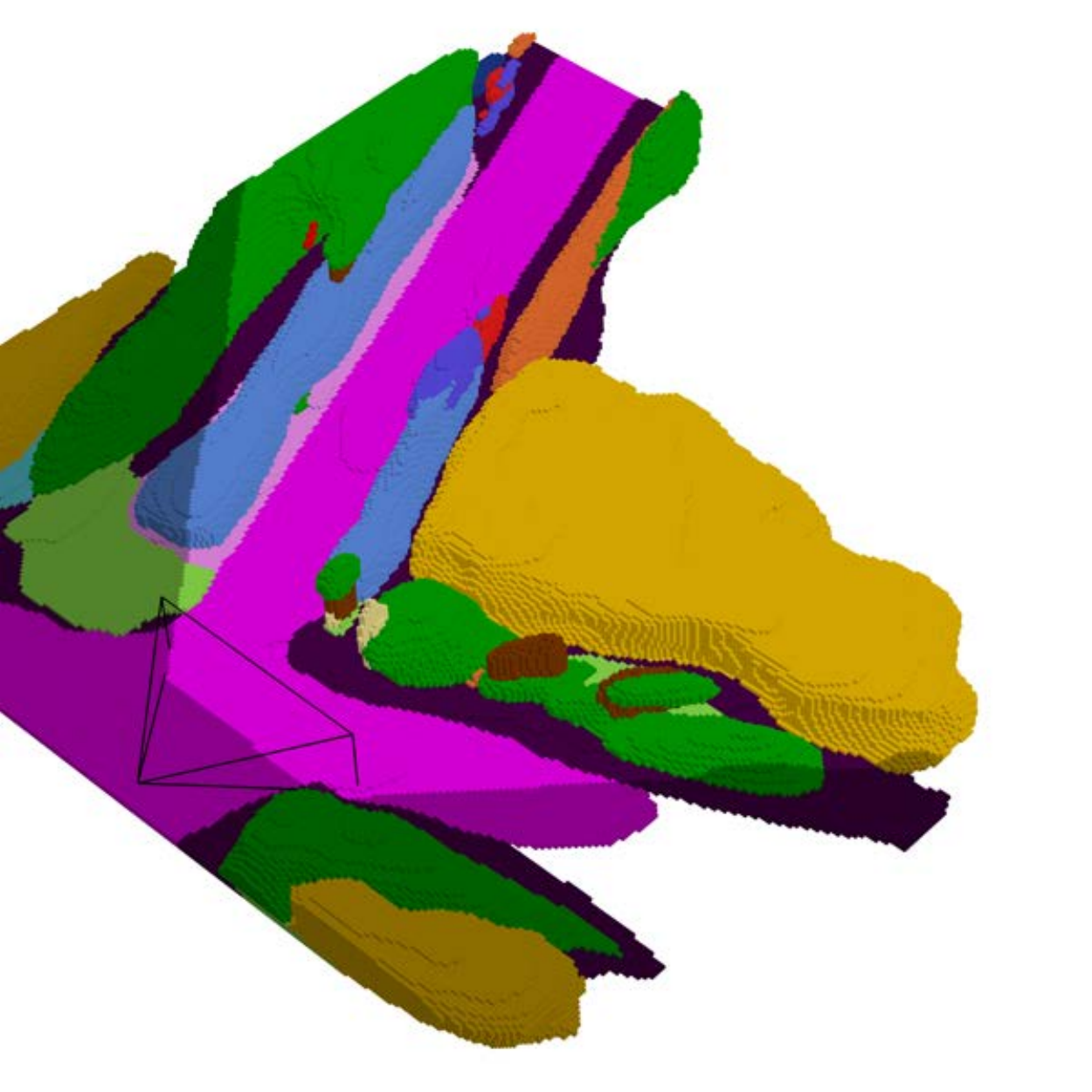} & 
		\includegraphics[width=.9\linewidth]{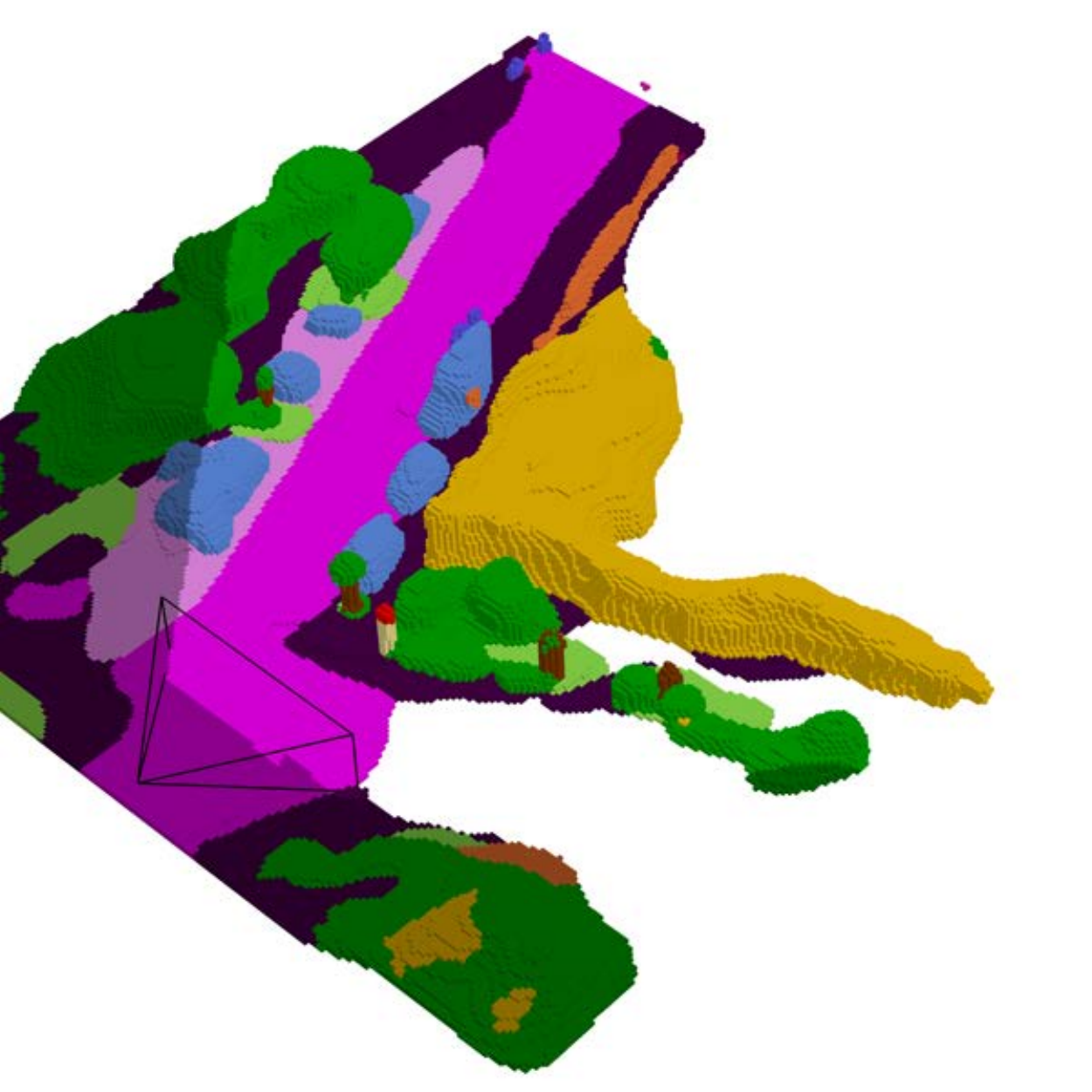} & 
		\includegraphics[width=.9\linewidth]{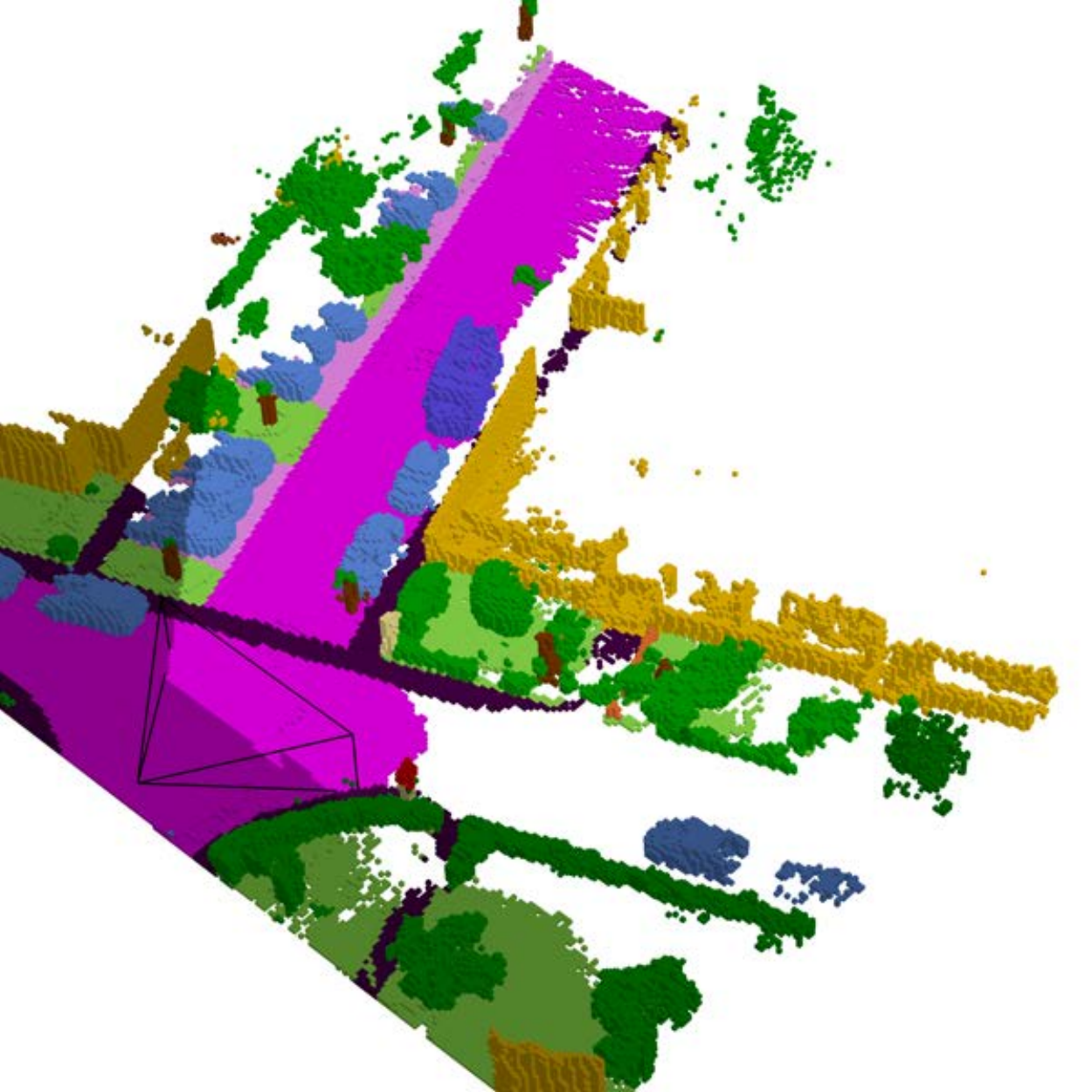} 
		\\[-0.1em]
		\includegraphics[width=.9\linewidth]{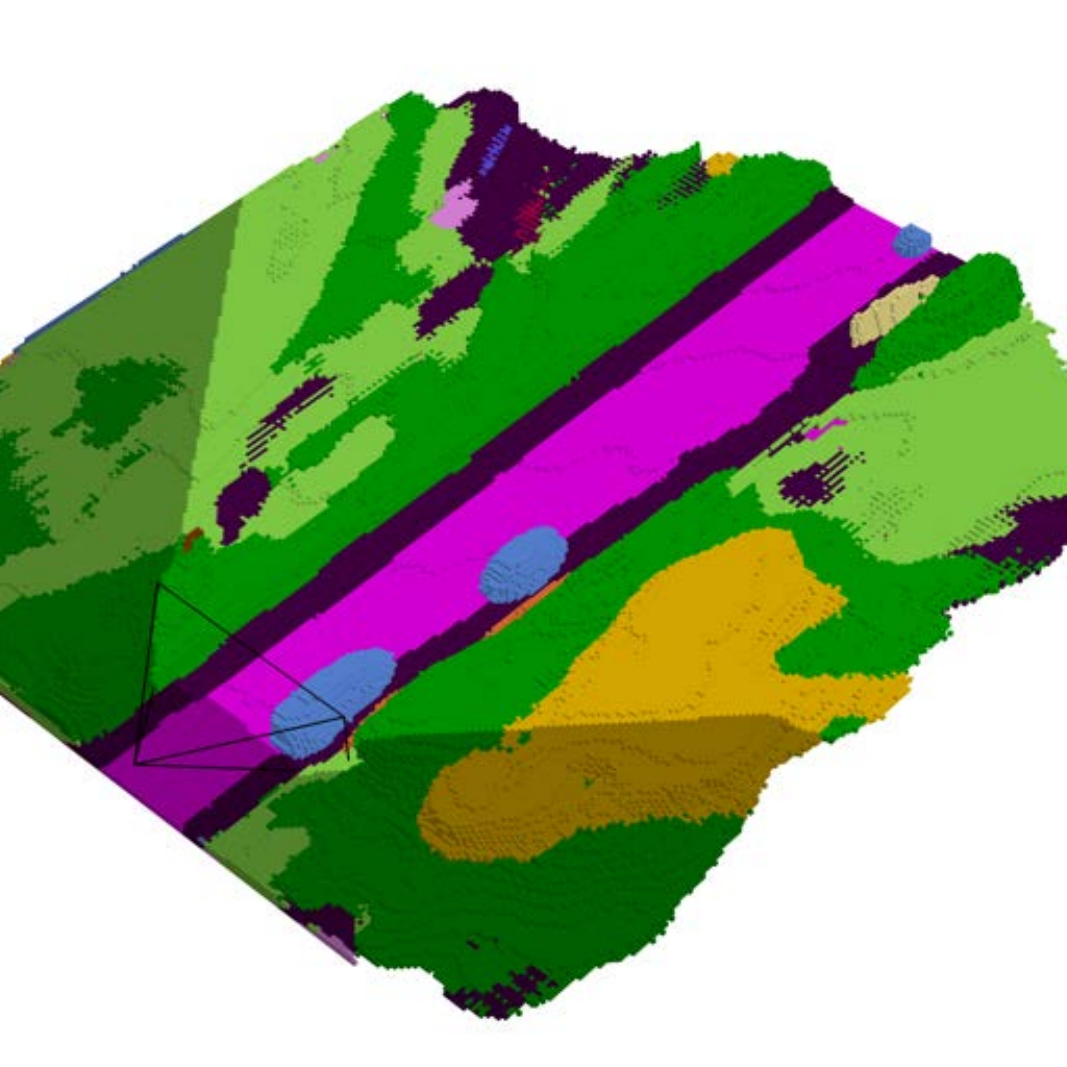} &  
		\includegraphics[width=.9\linewidth]{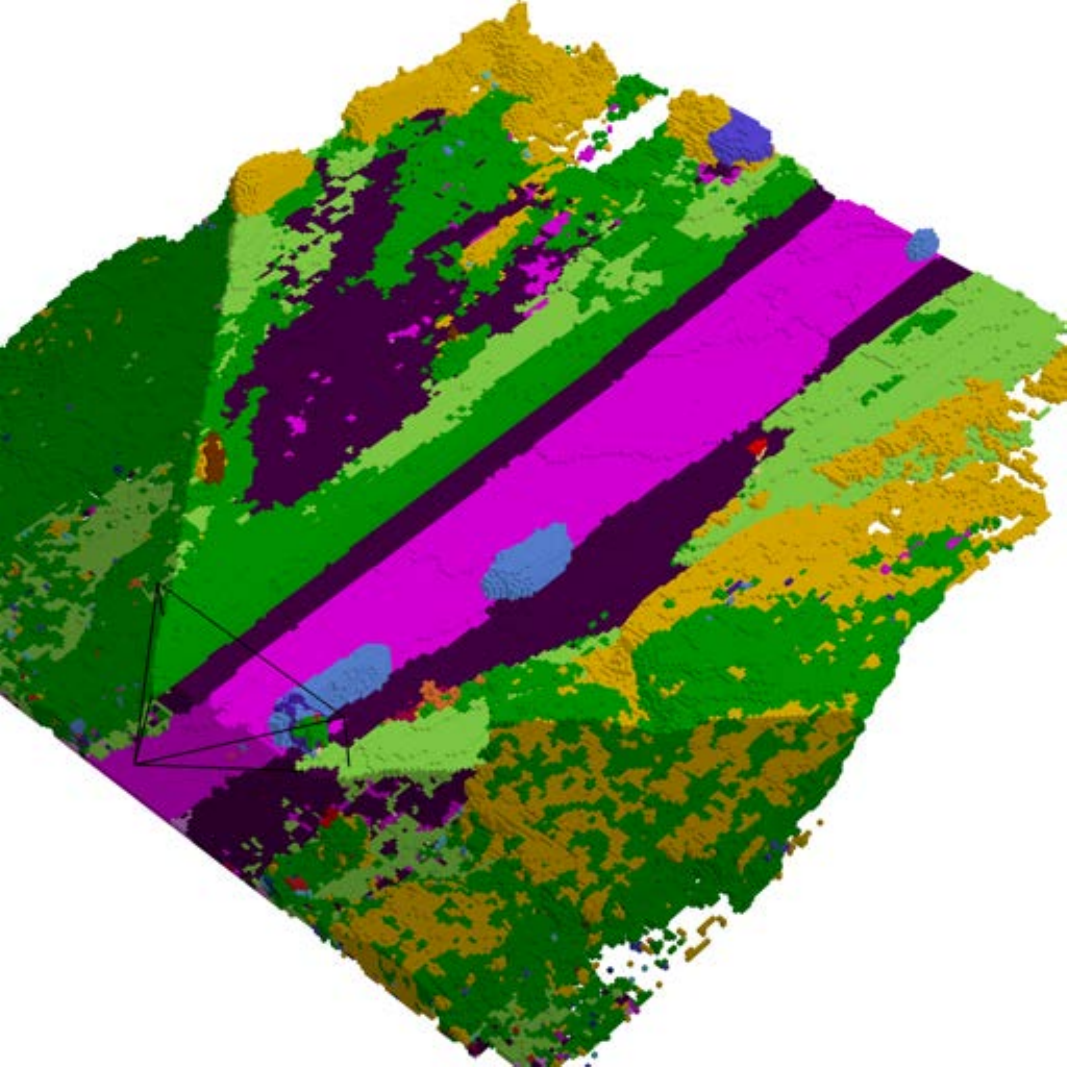} & 
		\includegraphics[width=.9\linewidth]{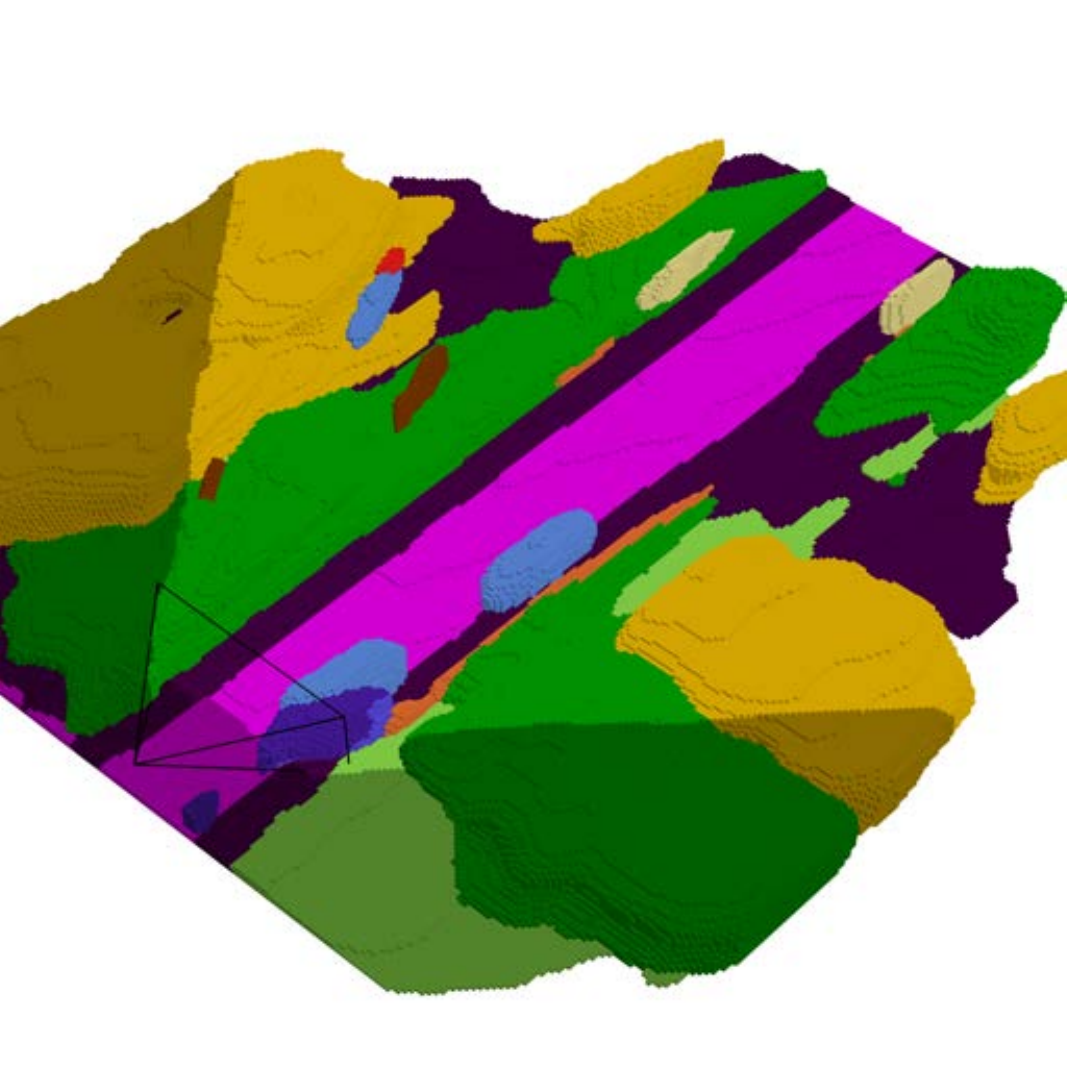} & 
		\includegraphics[width=.9\linewidth]{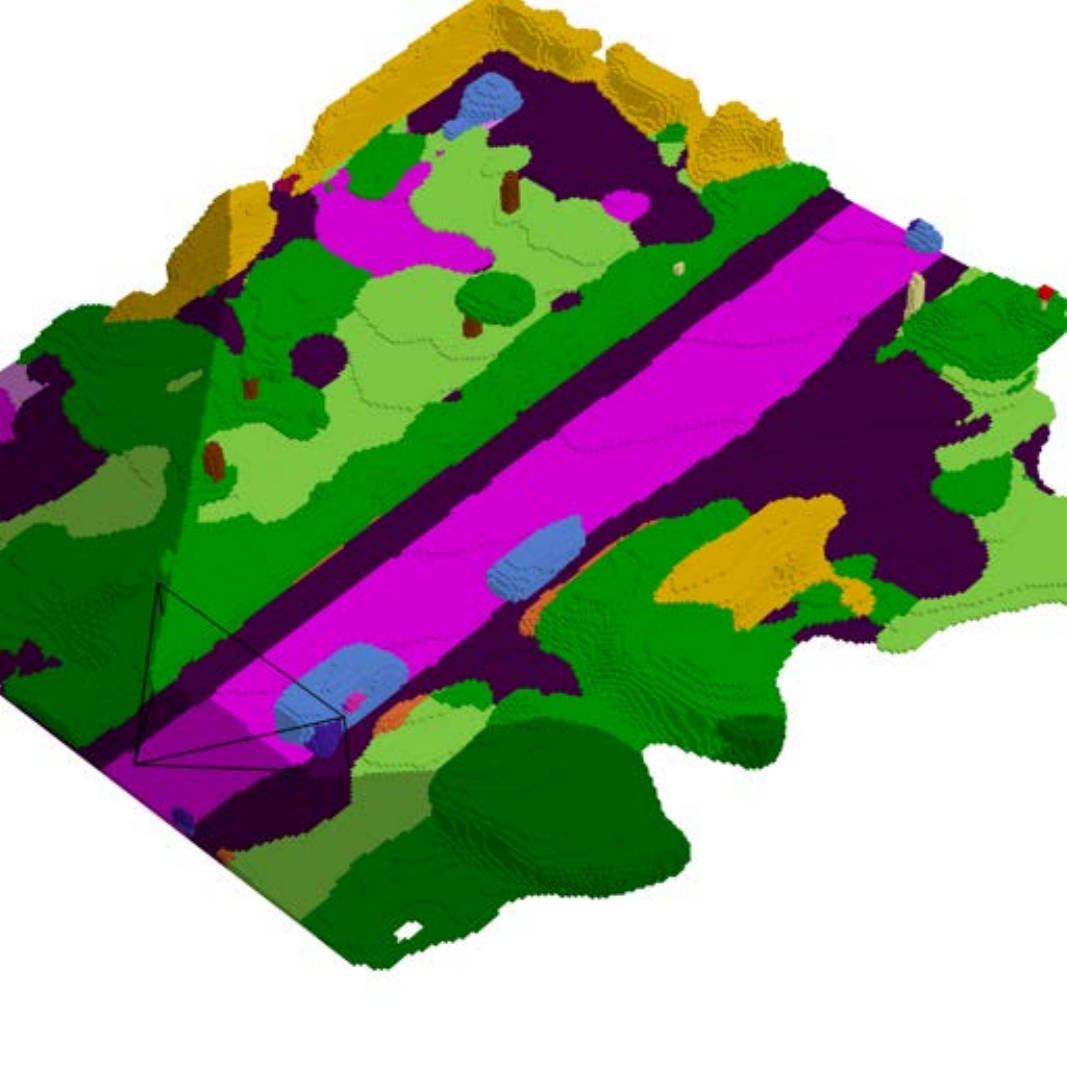} & 
		\includegraphics[width=.9\linewidth]{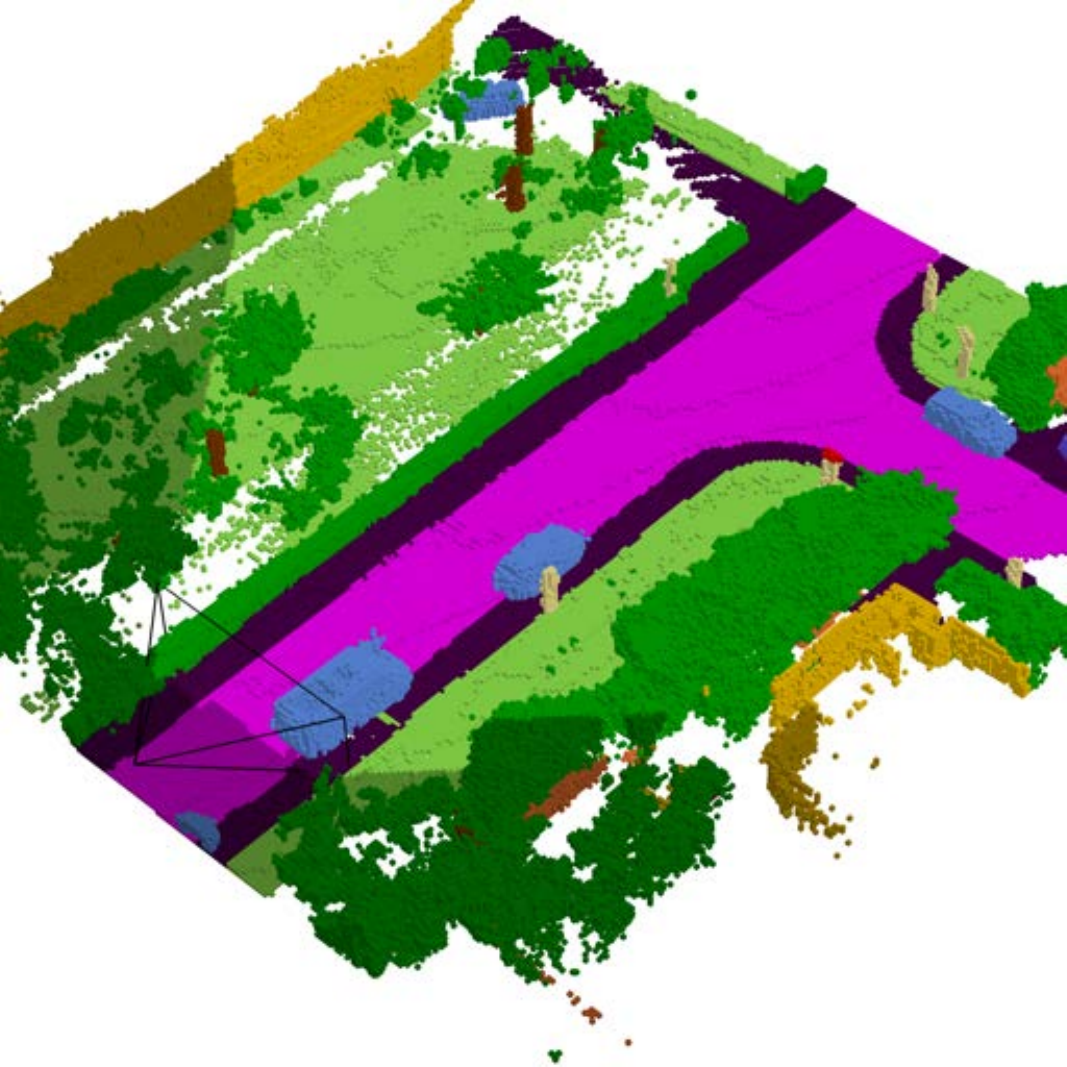} 
		\\[-0.1em]
		\includegraphics[width=.9\linewidth]{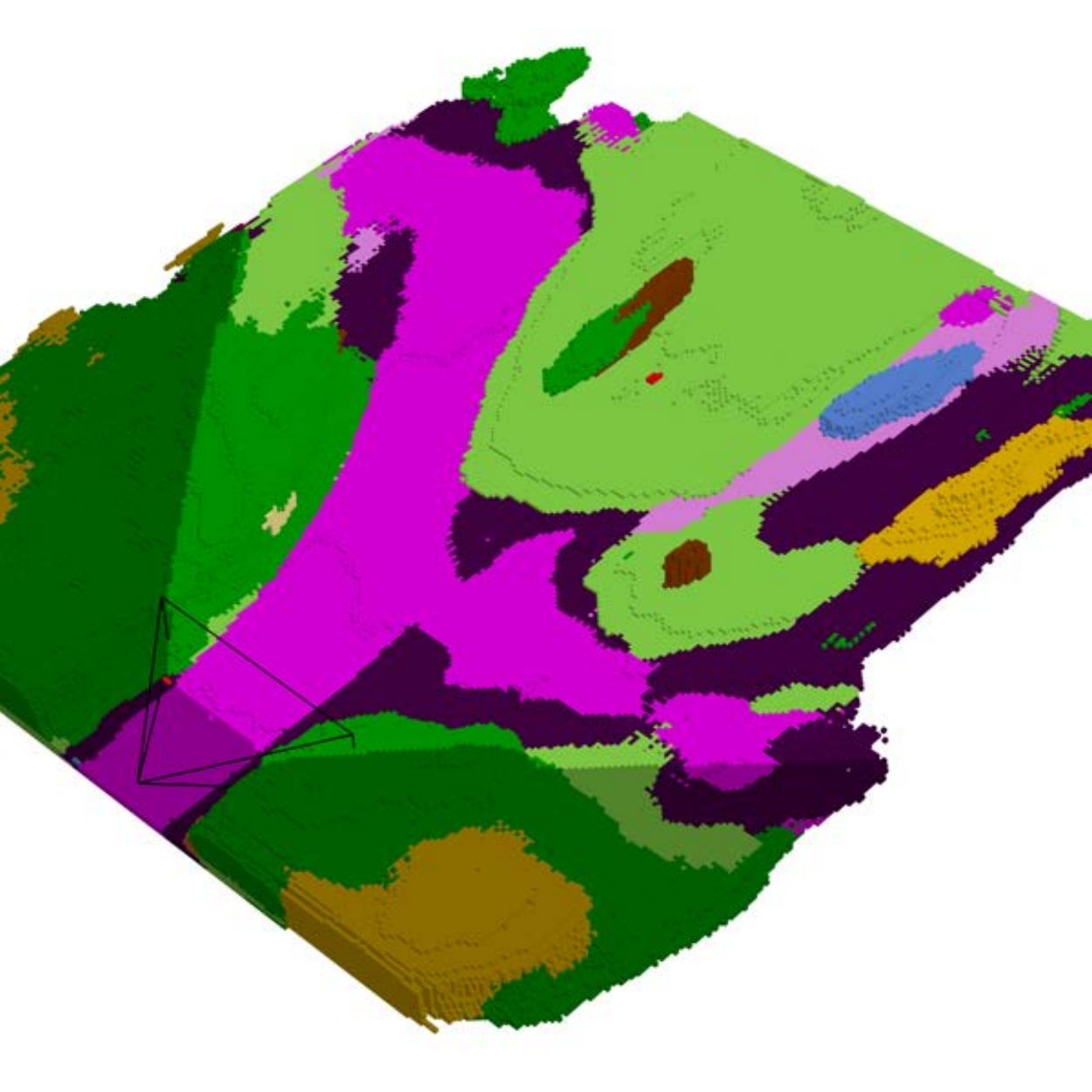} &  
		\includegraphics[width=.9\linewidth]{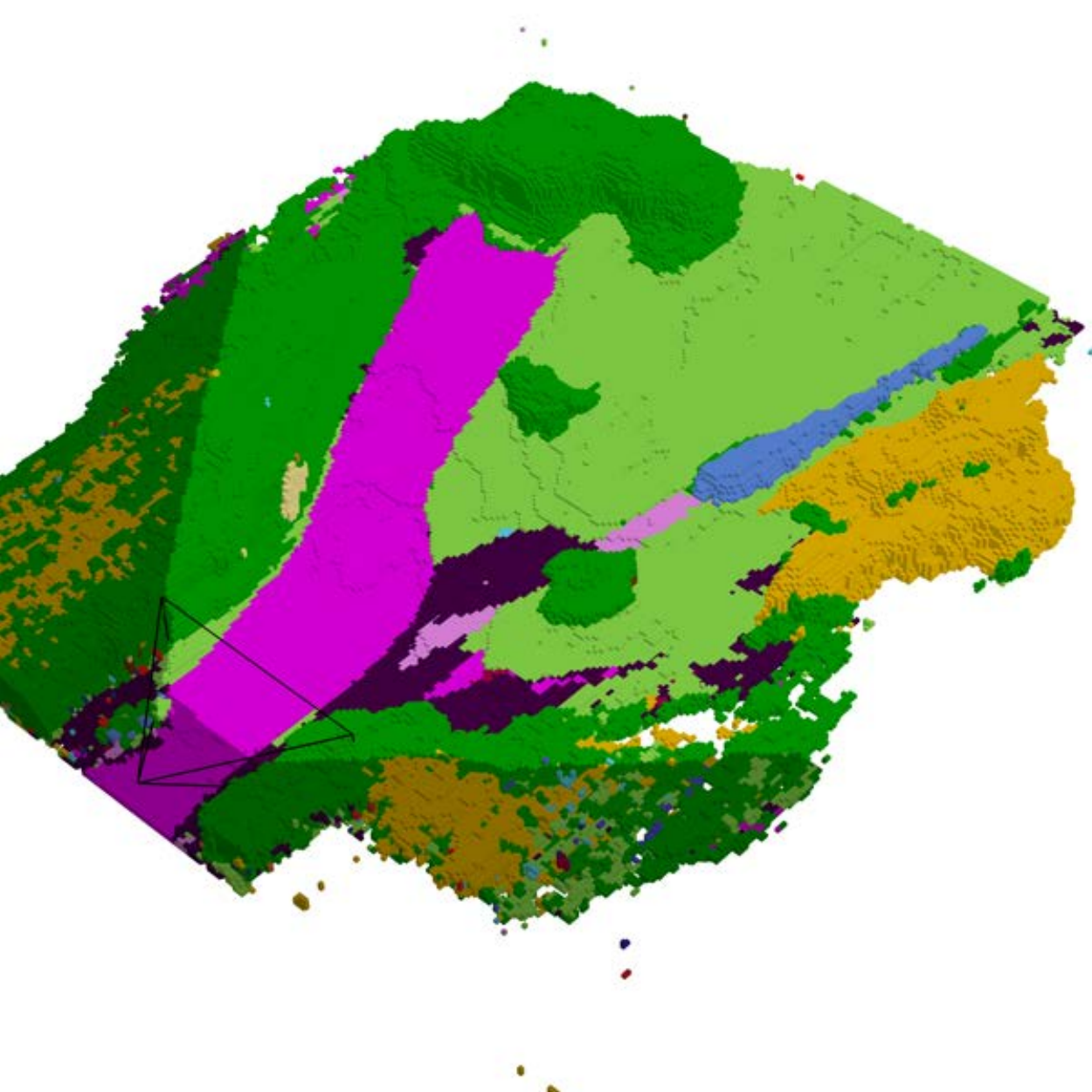} & 
		\includegraphics[width=.9\linewidth]{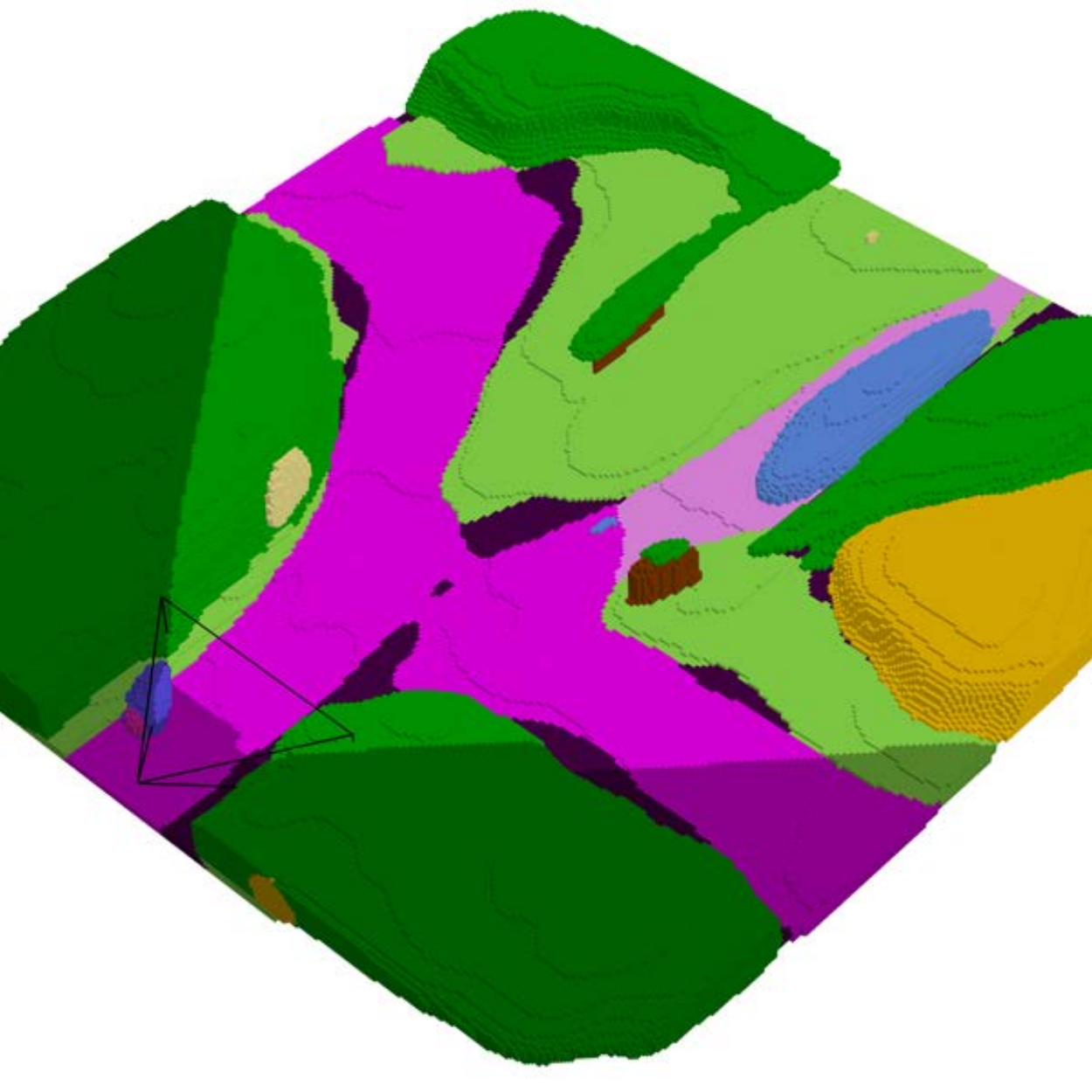} & 
		\includegraphics[width=.9\linewidth]{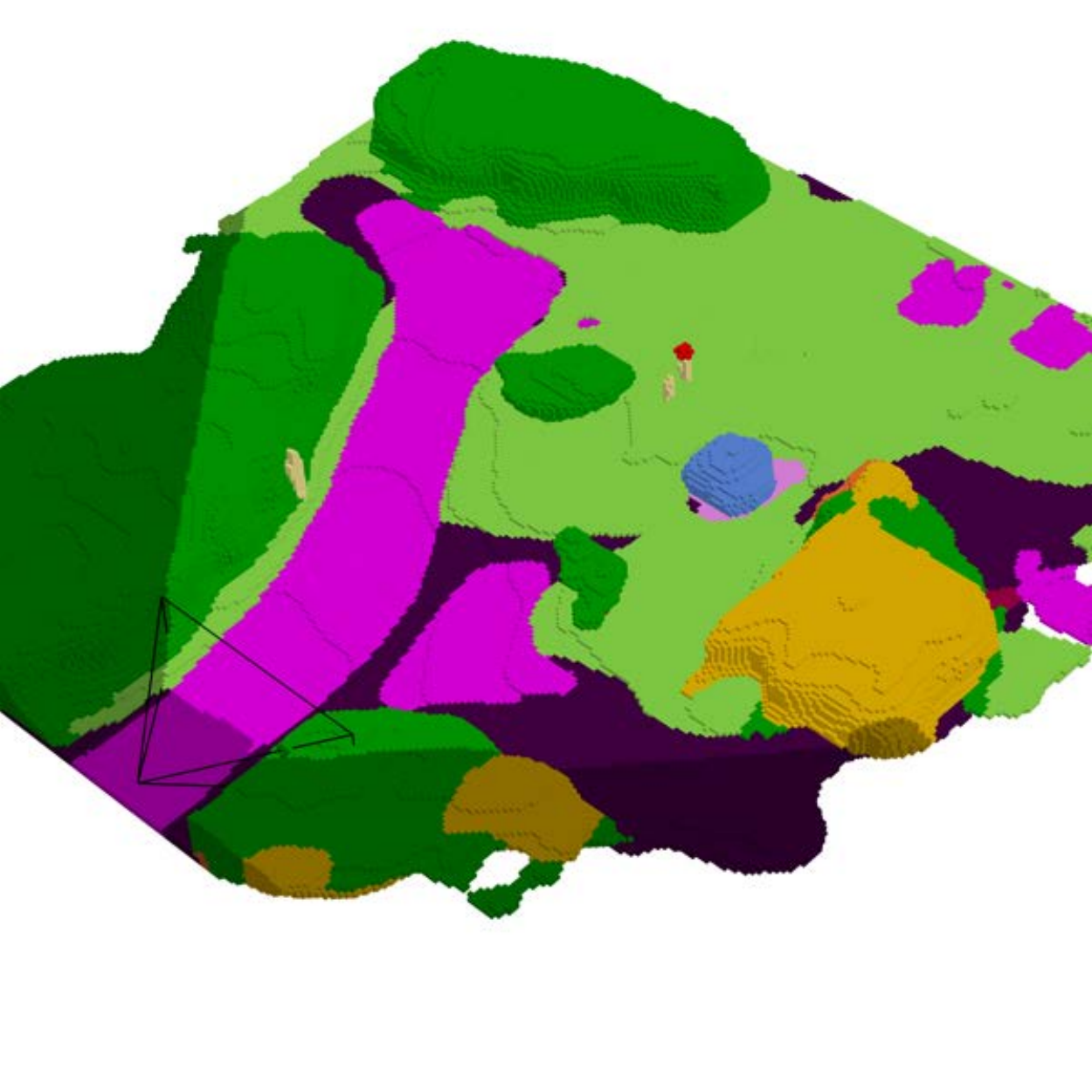} & 
		\includegraphics[width=.9\linewidth]{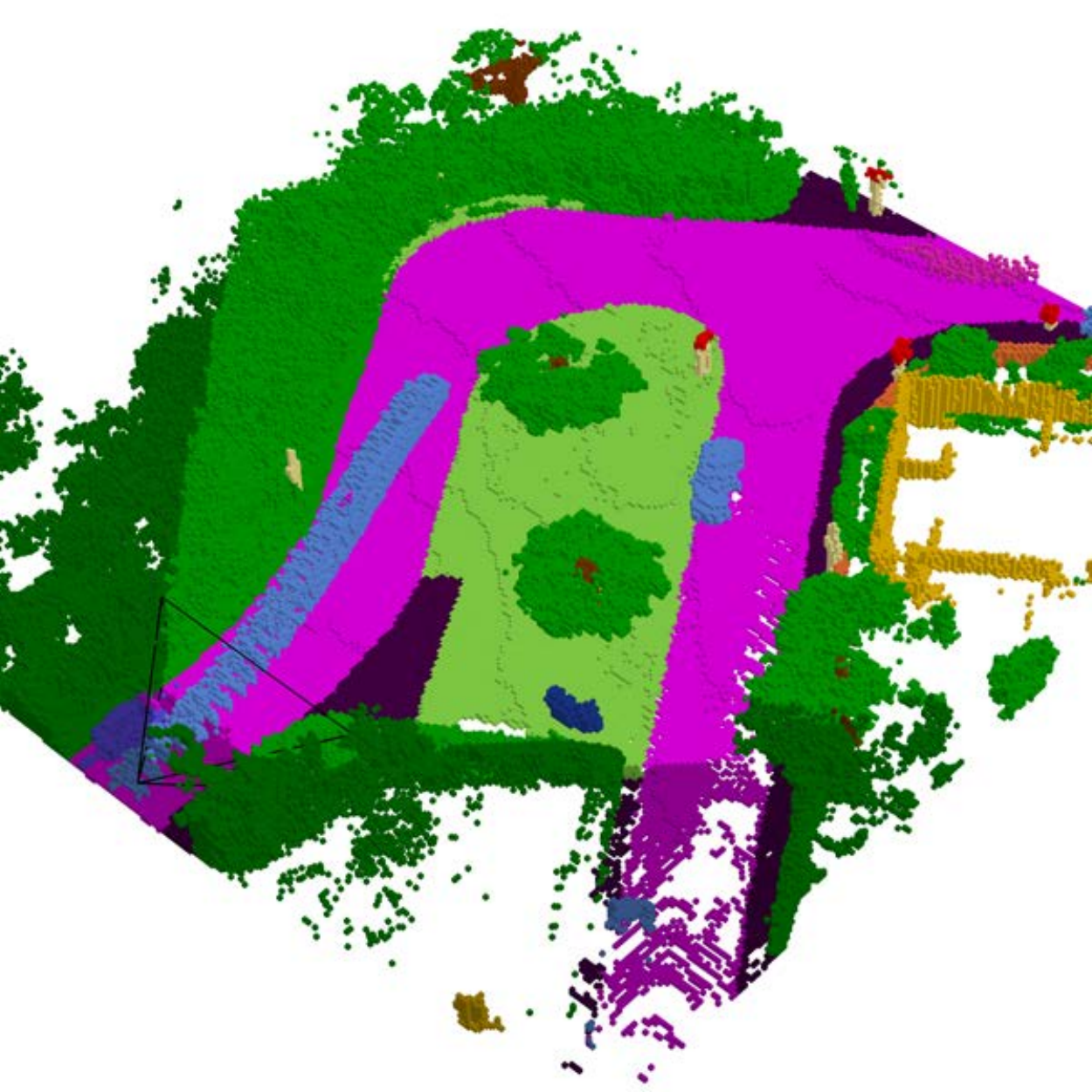}  
		\\[-0.1em]
		(a) MonoScene~\cite{MonoScene} & (b) VoxFormer~\cite{VoxFormer} & (c) OccFormer~\cite{OccFormer}& (d) CGFormer (ours) & (e) Ground Truth
	\end{tabular}
	\caption{More qualitative comparison results on the SemanticKITTI~\cite{SemanticKITTI} validation set.}
	\label{fig:AdditionQualitative2}
	\vspace{-4mm}
\end{figure*} 